\documentclass[a4paper, 12pt, twoside]{book}
\pdfoutput=1
\usepackage[utf8]{inputenc}
\usepackage{amsmath}
\usepackage{amsfonts}
\usepackage{amssymb}
\usepackage{graphicx}
\usepackage{subfigure}
\usepackage[english]{babel}
\usepackage{float}
\usepackage{chngcntr}
\usepackage{url}
\usepackage{fancyhdr}
\usepackage{gensymb}

\title{\textbf{Tesi}\\}
\author{Valentina Gregori}
\date{data della discussione}

\begin{document}

\begin{titlepage}
 \begin{center}
    \includegraphics[width=0.5\textwidth]{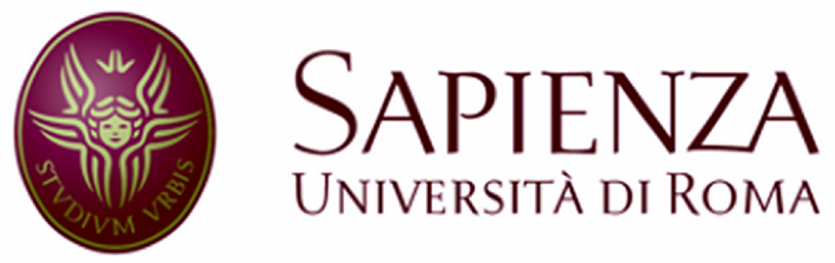}\\
     \vspace{4em}
      {\Large \textsc{Università degli studi di Roma\\ ``La Sapienza''}}\\
       \vspace{1em}
      {\Large \textsc{Facoltà di Scienze Matematiche Fisiche e Naturali}}\\
       \vspace{1em}
      {\small \textsc{Corso di Laurea in Fisica}}\\
       \vspace{3em}
      {\normalsize Dissertazione di Laurea Magistrale}\\
       \vspace{7em}
      {\LARGE \textbf{Leveraging over priors for boosting control of prosthetic hands}}\\
 \end{center}

\vskip 1.5 cm
  \begin{center}
    \begin{tabular}{l c c c c c c c c r}
      \textbf{Candidata} & & & & & & & & & \textbf{Relatore Interno} \\[0.2cm]
      \large{Valentina Gregori} & & & & & & & & & \large{Prof. Giovanni B. Bachelet}\\[0.3cm]
      \textbf{} & & & & & & & & & \textbf{Relatrice Esterna} \\[0.2cm]
      \large{} & & & & & & & & & \large{Prof.ssa Barbara Caputo}\\[0.3cm]
      \footnotesize{Matricola 1387986}  & & & & & & & & &  \\[0.3cm]
        & & & & & & & & &          \\
    \end{tabular}
    
    \normalsize Anno Accademico 2015/2016
\end{center}

\end{titlepage}

\newpage

\thispagestyle{empty}
\vspace*{\stretch{1}}
\begin{flushright} 
A Simona che,\\
nonostante la lontananza,\\
è sempre vicina.\\
\vspace{1cm}
A mamma e papà,\\
senza i quali tutto questo\\
non sarebbe iniziato.\\

\end{flushright}
\vspace{\stretch{2}}
\clearpage

\newpage
\tableofcontents
\newpage

\pagestyle{plain}
\pagestyle{fancy}

\fancyhead{}
\fancyhead[LE]{\leftmark}
\fancyhead[RO]{\rightmark}

\raggedbottom

\chapter*{Abstract}
\addcontentsline{toc}{chapter}{Abstract}%

The Electromyography (EMG) signal is the electrical activity produced by cells of skeletal muscles in order to provide a movement. The non-invasive prosthetic hand works with several electrodes, placed on the stump of an amputee, that record this signal. In order to favour the control of prosthesis, the EMG signal is analysed with algorithms based on machine learning theory to decide the movement that the subject is going to do.\\ 
In order to obtain a significant control of the prosthesis and avoid mismatch between desired and performed movements, a long training period is needed when we use the traditional algorithm of machine learning (i.e. Support Vector Machines). An actual challenge in this field concerns the reduction of the time necessary for an amputee to learn how to use the prosthesis.\\
Recently, several algorithms that exploit a form of prior knowledge have been proposed. In general, we refer to prior knowledge as a past experience available in the form of models. In our case an amputee, that attempts to perform some movements with the prosthesis, could use experience from different subjects that are already able to perform those movements.\\
The aim of this work is to verify, with a computational investigation, if for an amputee this kind of previous experience is useful in order to reduce the training time and boost the prosthetic control. Furthermore, we want to understand if and how the final results change when the previous knowledge of intact or amputated subjects is used for a new amputee.\\
Our experiments indicate that: (1) the use of experience, from other subjects already trained to perform a task, makes us able to reduce the training time of about an order of magnitude; (2) it seems that an amputee that tries to learn to use the prosthesis doesn't reach different results when he/she exploits previous experience of amputees or intact.\\

\clearpage{\pagestyle{empty}\cleardoublepage}
\chapter{Introduction} \label{cha:Intro}

\begin{figure} [H] 
\centering 
\includegraphics[scale=0.5]{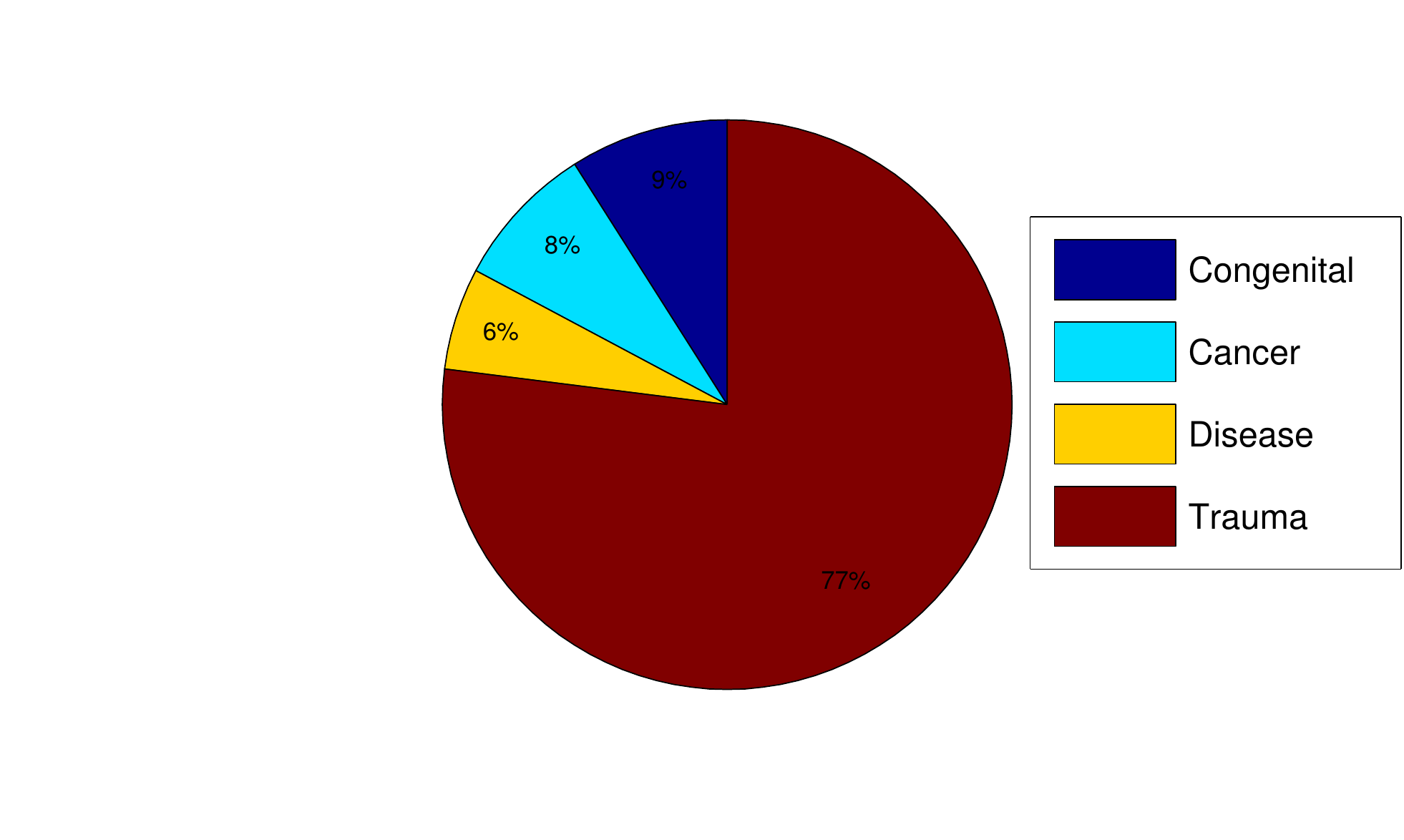}
\caption{ISHN statistic referred to upper limb amputees.}
\label{fig:Pie}
\end{figure}

The problem of amputations is very actual. According to data from the National Center for Health Statistics, there are 50,000 new amputations every year in the USA. The statistics of COPC (Center for Orthotic $\&$ Prosthetic Care) show that the level of amputation concerns the upper limb in 86 $\%$ of cases and lower limb in 14 $\%$. Among upper limb amputees, the trans-radial ones make up the 60 $\%$ of total wrist and hand amputations.\\
Starting from this statistics we focus our attention on upper limb trans-radial amputees. Reasons for amputations include cardiovascular disease, traumatic accidents, infection, cancer, nerve injury and congenital anomalies (in according to ISHN statistics). In the Figure \ref{fig:Pie} statistic for upper limb amputees is shown.

\begin{figure} [H]
\centering
\includegraphics[scale=0.6]{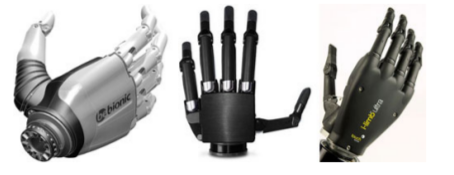}
\caption{Example of hand prosthesis. Source: \cite{Tom_hand_pros}.}
\label{fig:ProsthHand}
\end{figure}

Nowadays there exists two types of prosthetic hands: invasive and non-invasive. The first ones provide for direct installation in the arm of the amputee with the surgery. The second ones can be put on during day and removed in the night or when an amputee wants. In this work we focus on myoelectric prosthetic hands that belong to the second type of prosthesis. These are very advanced from the hardware point of view thanks to the exploitation of several portable sensors used to gather the electrical signal from the stump of an amputee. From a software view of point, the prosthesis is controlled with several algorithms that analyse the input signal and provide the best output movement. Examples of prosthesis are shown in Figure \ref{fig:ProsthHand}.\\ 

Nowadays robotics has reached great heights and there exist robotic hands designed to perform any desired task or a predetermined sequence of these. Also the prosthesis, built for the amputees, are very advanced: these have five fingers and can potentially perform all the possible configurations, like a human hand.\\ 
Despite the rapid progress that robotics has had in last years, for an amputee the control of non-invasive robotic prosthetic hand is still far (\cite{NinaData_presentation1}). The real challenge in the world of robotic hands is to achieve the total control of prosthesis. In fact, in most cases the tasks that an amputee can perform with this are limited to opening and closing. The control of prosthetic hand is far from simulation of natural movements and the training process to make amputees familiar with them is still long and sometime painful (\cite{NinaData_presentation1} and references therein).\\
The open challenge in the world of bio-robotics is to try to reduce the time required for learning of use of prosthesis and to make this control as natural as possible.\\

\section{Contributions of this thesis}
This thesis tackles the control of prosthesis with an approach based on machine learning, as previous works in this field suggest (\cite{Tom_hand_pros}, \cite{Novi_Nina1}, \cite{Orabona_DAforEMG}). Our purpose is to increase the level of dexterity in the use of prosthesis, in order to favour a natural form of control.\\ 
The prosthesis to which we refer in this work are composed by several electrodes that gather the Electromyography (EMG) signal, i.e. the electrical manifestation of muscular activation. With machine learning techniques, we analyse this signal in order to decide the most probable movement that the subject wants to perform.\\ 
Currently, machine learning algorithms have success in this field, and not only, for their generality and for the possibility to enlarge the traditional theory. One of the most famous definition of machine learning is provided by Tom M. Mitchell and can be summarized as follow: a computer program is said to learn if its performance at a given task is improving with experience.\\ 
The given definition is enough intuitive because it is not very far from the traditional human learning and skill. Human beings don't learn new things in isolation but with the help of what it is already known: i.e. their previous experience (\cite{Tom_thesis}).\\ 
This intuition can be extended also for the control of prosthesis: for an amputee could be simpler to learn a movement using experience from other subjects that are already able to perform this. If this statement is true, which previous experience to exploit could play a key role and could change the final learning result. Thus, the questions that this thesis aims to answer are essentially two: 

\begin{itemize}
\item Is it possible to speed up the process of control of prosthesis with the help of prior experience? 
\item What does it change when we use for an amputee the experience from other amputees, as opposed to that from intact subjects?
\end{itemize}

The contribution of this work is to answer to these queries following an experimental approach. In order to achieve this goal we performed a thorough series of simulations exploiting data from a public database. Our results show that the help of previous experiences boost the control of prosthetic hands and speed up the learning process. For an amputee it is not achieved a different result when experience from amputated or intact subjects is exploited.\\

\section{Outline}
The content of this thesis is organized as follows:\\

In Chapter \ref{cha:Few_land} we give some landmarks useful to have an introduction of the problems tackled in the following and a general vision of the entire work. We begin with an overview of different kind of prosthesis and the general problems and challenges related to their usage and control. In the end of the chapter we enumerate all the contributions given by previous works in this field about data, analysis and algorithms.\\

In Chapter \ref{cha:Data} we describe the experiment NinaPro that gathered kinematic, dynamic and electromyographic data from intact and amputated subjects and built the database that we used in our experiments. In the first part of the chapter we describe in detail the acquisition protocol, used devices and the data collection. In the second part we describe the path generally followed in this field for processing and representation of raw data acquired.\\ 

Chapter \ref{cha:Algorith} focuses on the learning algorithms used in this work to solve the problem of recognition and classification of different hand movements. First we present an introduction about the traditional and innovative theories on which these methods are based. In the second part we introduce the used algorithms and their implementation.\\

In Chapter \ref{cha:Results} we present our experiments and results obtained. We have worked with three different settings using data described in Chapter \ref{cha:Data} and algorithms introduced in Chapter \ref{cha:Algorith}. First we present each experimental protocol and the results obtained in the three experiments done. We conclude with an overall discussion and comparison of our findings.\\

The thesis ends with a conclusive summary and possible direction of research and future works.

\clearpage{\pagestyle{empty}\cleardoublepage}
\chapter{A few landmarks} \label{cha:Few_land}
This chapter gives a general vision of the entire works: it aims to focus the problems linked to prosthesis, the best path to tackle them and the related works useful in this task.\\ 

Section \ref{sec:Pros} focuses on different prosthesis currently used by amputees. After a general introduction, we list the problems that an amputee tackles when he/she learns how to use a prosthetic hand.\\
Section \ref{sec:Challeng} is about challenges and questions that this work aims to solve. We explain the general path taken and, briefly, the results obtained.\\
In section \ref{sec:exp_frame} we report a general vision and a summary of all the work developed.\\
In section \ref{sec:Prev_work} we list the principal works that have dealt with related problems.\\

\section{Prosthesis and control} \label{sec:Pros}

As said in the introduction, our study is referred to upper limb amputated, that are the majority in the community of amputees, and to non-invasive prosthesis.\\
The most common non-invasive prosthesis are: cosmetics, body-powered and myoelectrics (Figure \ref{fig:Prosthesis}).\\

\begin{figure} [H]
\centering
\includegraphics[scale=0.6]{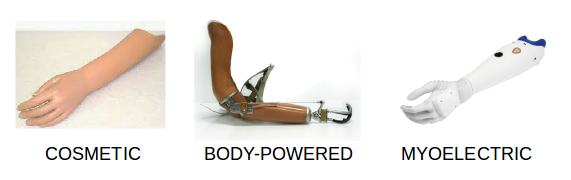}
\caption{Different types of non-invasive prosthetic hands.}
\end{figure}\label{fig:Prosthesis}

The cosmetic ones are only aesthetic hands, they are comfortable but any type of grasping or movement is forbidden.\\ 
The body-powered are mechanical prosthesis, they work by using cables to link the movement of the body to the prosthesis and to control it. When an amputee moves the body in a certain way he/she drives the cables, consequently the prosthetic hand is opened, closed, or bended. Obviously only few movements can be performed.\\
The myoelectric prosthesis use several electrodes placed in contact with the skin of the stump. These kind of electrodes are called electromyography surface and they detect electrical activity produced by skeletal muscles when a movement is performed. The recorded electromyographic signal (EMG) varies in: $\sim 10 \mu V \div 10 mV$ (\cite{NinaData_presentation1} and references therein). The electrodes record the muscular activity, the signal is analysed in order to decide the intentional movement and, theoretically, infinite positions can be performed.\\

In this work we study the control of non-invasive myoelectric prosthesis. They could potentially improve the quality of life of an amputee, but the control system is difficult. The open challenge in the world of myoelectric prosthesis is to try to reduce the training time, i.e. the time required for learning how to use them. Nowadays, this is still a very long process, often with great mismatch between desired and performed movements. Moreover this is generally perceived as very tiring and sometimes painful by the users. These reasons make the use of myoelectric prosthesis still limited in practice. Often the amputees stop using this kind of prosthetic hands and replace them with a cosmetic ones (\cite{NinaData_presentation1}).\\

\section{Path in learning problems} \label{sec:Challeng}
The control of robotic prosthetic hands using non-invasive techniques, like myoelectrical surface, is still a challenge nowadays. In bio-robotics and rehabilitation community it is clear that the success of myoelectric prosthesis is linked to the creation of an accurate control system to make them easy to use by the patient (see \cite{Tom_hand_pros} and references therein).\\
The path generally chosen to tackle the problem is machine learning (\cite{Tom_hand_pros}, \cite{Novi_Nina1}, \cite{Orabona_DAforEMG} and references therein). It makes it possible to analyse the electromyography signal with modern statistical techniques like support vector machines, neural networks and linear discriminant, in order to guess the movement that the subject wants to perform (\cite{Tom_hand_pros}). Generally with these techniques a subject can choose among a finite number of hand postures. Thus, the final configurations are not all the possible ones but only a selection of these.\\
In a machine learning problem a learner system tries to perform a task with the help of previous experience. In our work the learner is the subject that wants to learn to perform several movements with a robotic hand. Evidently, the choice of experience influences the whole problem.\\ 
Most of the existing machine learning methods build a new learning model directly over the data from the learner itself. In our case, starting from the EMG signals collected from the user, it is built a function that, for any future unknown EMG signal, chooses the most probable hand posture associated. This method gives good results only if a large amount of training samples of the subject is available.\\
%
%
%
Most recent algorithms suggest to exploit a source of knowledge external to the subject studied: i.e. experience from other subjects. If this previous experience consists in robust statistical models built in the past, these can be reused when a new patient trains the prosthesis.\\ 
Obviously a form of adaptation is needed when we use experience from a subject to boost the learning of a new patient because the two domains could differ. The EMG signal recorded from different subjects can vary due to characteristics of the forearm (like size and shape), personal characteristics of subjects (like gender, age, use of the arm), electrode displacement and muscular fatigue (\cite{Castellini_EMG}).\\
By using state of the art adaptive learning algorithms, we want to understand if knowledge from other subjects can boost the control of prosthesis and reduce the learning time for a new amputee. Previous experience that an amputee exploits to achieve the goal of a natural use of prosthesis could come from amputees or from intact subjects. Our task is to study how the final result might change in the two cases.\\

%
%
%
%

\section{Experimental framework} \label{sec:exp_frame}
\begin{figure} [H] 
\centering
\includegraphics [scale=0.56]{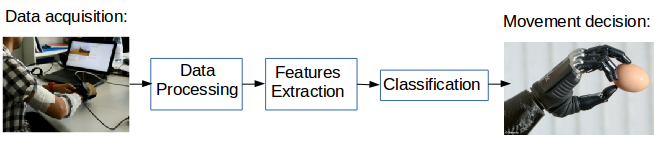} 
\caption{Steps in a complete experiment.}
\label{fig:MovProcesses}
\end{figure}

We clarify now all the passages to perform an experiment that aims at the control of non-invasive myoelectric prosthesis.\\
We consider intact and amputated subjects from which we gather electromyographic data. These data are collected by using, as sensors, surface electrodes placed on the last part of the arm for an amputee, or on the forearm for an intact subject. These electrodes detect the electromyographic signal generated by muscular contractions, where each signal corresponds to a different movement.\\ 
The aim of prosthetic control is to establish the best output movement for each given input signal. In order to achieve this goal the input data are processed to remove a component of noise. In a second step, we extract useful information from the data, in order to determine the discriminant characteristics of signal that we want to classify. This process is called features extraction.\\
Data appropriately processed are used as input of different recognition algorithms. These algorithms solve in different ways a classification problem: given an input signal their aim is to find the right output movement (i.e. the right class) between the ones proposed. In this kind of problem is not possible to select other classes than the initial ones, thus the final hand movements are not all those that are the possible. Each algorithm works by exploiting differently the previous experience coming from source subjects.\\
All the described steps are reported schematically in Figure \ref{fig:MovProcesses}.\\

In this work we exploited data acquired during the experiment \textit{NinaPro} (2011-2013) and online available. After the choice of data with which to work, we were dedicated to their processing and representations, following what the literature proposes. To solve the final classification problem we used the adaptive learning algorithms that constitute the state of the art in the field of machine learning.\\

\section{Previous works} \label{sec:Prev_work}

\paragraph{Database and EMG signal.} Data used in this work have been collected in the experiment \textit{NinaPro}, where 10 or 12 sEMG electrodes (it depends on the chosen configuration) gather the electical signal from the arm of the users. In \cite{NinaData_presentation1}, \cite{NinaData_presentation2}, \cite{NinaData_presentation3} all the details about used devices, the acquisition protocol and any information on database are present.\\
NinaPro represents an important contribution in this field because it is, in according to our knowledge, the only existing public database that gathers data of a consistent number of subjects and postures. As \cite{NinaData_presentation1} and references therein underline, previous studies usually include too few subjects and too few tasks: the maximum is represented by 11 intact subjects and 6 amputees with a maximum of 12 tasks. The NinaPro database collects data and clinical information of 78 subjects, 67 intact and 11 trans-radial amputees, that perform 53 or 50  postures (it depends on the chosen experiment). This large amount of data makes results obtained with this database statistically relevant.\\

Data acquired need several preprocessing steps in order to make them available for a recognition process or for movements classification. The preprocessing of EMG data and their final representation, called features extraction, could have a profound impact on the final performance of the algorithm chosen to solve the classification problem of EMG signal.\\
In \cite{Zecca_Features} the interested reader can find all the details about feature extraction methods used in order to analyse sEMG signals. The methods described in that work can take into account amplitude or spectral properties of the signal. The first ones are the algorithms that work in the time domain: Mean Absolute Value, Variance, Waveform Length or Cepstral Coefficients. The second ones are the algorithms that operate in the frequency domain: Frequency Ratio or Mean Frequency. Considering time and frequency domain at once we obtain the time-frequency domain features: Short Time Fourier Transform, Wavelet Transform, or Wavelet Packet Transform. These are richer of information about the signal but the computational cost increases.\\
Authors of \cite{Kuzb_FeaturesExtrac27} studied and compared different methods in order to understand which are the ones that give the best ratio between the final performance and computational cost. Their results show that Mean Absolute Value and Waveform Length, despite their simplicity, can reach similar performance to the computationally more demanding marginal Discrete Wavelet Transform.\\

\paragraph{Domain Adaptation.} It is a field associated with machine learning used to overcome the distribution mismatch between different domains. In general, in order to solve a new target problem with few labelled data, we can exploit solid models from sources, built with a large amount of data. The adaptive learning methods transfer useful information from the source domain to the target domain, although these are different, when the task to solve is the same. With this technique we avoid the collection of new samples from our target and the building of a little robust model based only on the few data gathered.\\

How to exploit the source prior knowledge and how to adapt it to the new target model depends on the algorithm used. Over the last years different directions have been proposed on how to tackle the problem. Each one suggests a different adaptive method between sources and target.\\
There are methods that aim to approach directly data from target and sources using some mathematical tricks. In \cite{2SW_MDA} a process based on two stage of weighting for each source sample is used in order to overcome the distribution mismatch. The authors of \cite{GFK} take into account the fact that source and target live in different space; in order to overcome this problem it is exploited a function that builds a path from source to target in a dimensional reduced space.\\
Another possible solution is to leverage over source models exploiting their parameters. The basic idea is that the new parameters of the target models must be close to the source ones and that they must be found solving an optimization problem. The first algorithm proposed in this direction is presented in \cite{Orabona_DAforEMG}, in this case only a single source model is exploited (the best one). In \cite{Tom_hand_pros} this approach is enlarged by exploiting different linear combinations of source models.\\
In the last method proposed in domain adaptation the sources are considered as experts. The judgement given by them for each target sample consists in extra features. We can distinguish three different levels in this approach, depending on the processing of the extra features. A middle level method based on this theory is proposed in \cite{Orabona_MKL_1} and \cite{Orabona_MKL_2}. An high level method is instead exploited in \cite{Novi_Cue}.\\
All these different approaches evaluate autonomously the importance of each previous knowledge and decide individually from where and how much to transfer.\\
  
The traditional fields in which domain adaptation is applied are: language processing for speech recognition (\cite{DA_SpeackRec1}) and computer vision for image classification (\cite{DA_CompVis_Saenko}, \cite{Novi_Cue}).\\

\paragraph{Application of domain adaptation to biological signal.} The problem that we tackle in the prosthetic control is a domain adaptation problem due to the differences of EMG signal distribution among different subjects. In fact, the direct use of sources data for the solution of a target problem could result poorly in performance due to the many differences between the two domains. EMG signal from sources and target can differ for the user's age, gender, height, weight, dominant hand, different exercise of the arm muscle and placement of electrodes (\cite{Castellini_EMG}). Thus an adaptation process is necessary.\\ 

This kind of studies and application are very recent. One of the first work, according to our knowledge, proposed in this environment is \cite{Orabona_DAforEMG}. This proposes a method that chooses automatically the best source that the target problem can exploit. It uses directly its parameters in order to build the new target model. This algorithm is generally called \textit{Best-Adapt}.\\ 
Previous method is resumed and revisited in \cite{Tom_hand_pros}, where two adaptive algorithms able to exploit many prior knowledge models are proposed and compared with the previous one. The two methods are called \textit{Multi-Adapt} and \textit{Multi-perclass-Adap}. The first exploits a weighted combination of sources, the second assigns a different weight to each class of each source. In \cite{Tom_hand_pros} these algorithms are tested on sEMG data from 10 and 20 intact subjects with 6 postures. The algorithm that gives the best result in performance is \textit{Multi-perclass-Adap}.\\
In \cite{Novi_Nina1} several existing algorithms are tested. The adaptive methods that are compared come from \cite{2SW_MDA}, \cite{GFK}, \cite{Orabona_MKL_1} and \cite{Tom_hand_pros} (see previous paragraph). The experiments of this work involve 10 intact subjects with 9 postures and 27 intact subjects with 12, 17, 23 and 52 postures. The results show that the method proposed by authors of \cite{2SW_MDA} need a running time much longer than the others. Generally the method that achieve the best performance is the one proposed by authors of \cite{Orabona_MKL_1}.\\
In both cases (\cite{Tom_hand_pros},\cite{Novi_Nina1}) experiments showed that for intact subjects the postures recognition and classification can be improved and boosted by the use of prior knowledge of other intact subjects.

\clearpage{\pagestyle{empty}\cleardoublepage}
\chapter{Data description and representation} \label{cha:Data}
Research in the area of hand prosthetics suffered from a number of problems. First, previous studies in this field are based on few subjects with few hand postures: according to our knowledge, up to 11 intact subjects and 6 amputees that perform a maximum of 12 movements (see \cite{NinaData_presentation1} and references therein). It makes it hard to obtain statistically relevant results. Second, it is necessary to establish a data analysis method accepted in order to compare final results.\\
Currently, the best public database is NinaPro (Non Invasive Adaptive Prosthetics). Its data have been collected during the experiment NinaPro (2011-2013) and, according to our knowledge, represents the state of the art among public database in this field. In this work, we decided to use data coming from NinaPro for our experiments in order to overcome the first problem described above.\\
Regarding data processing, we decide to use the algorithms of features extraction proposed in literature that, currently, represent the state of the art in the field of analysis of raw data.\\

In section \ref{sec:Data_acq} it is introduced the NinaPro database. We describe the acquisition set-up with particular attention to the experimental protocol and used devices to collect data from subjects. In the last part of this section we focus on the organization and structure of the dataset.\\
In section \ref{sec:Feat_ext} we explain all the procedures for data processing.\\
Data and all the reference papers are available in \url{http://ninapro.hevs.ch/}.\\
 
\section{Data description: the NinaPro database} \label{sec:Data_acq}
The NinaPro (Non Invasive Adaptive Prosthetics) database holds data and clinical information about 78 subjects: 67 intact and 11 trans-radial amputees. According to our knowledge, it is the only existing public database with a consistent number of subjects. Data have been acquired from 2011 to 2013.\\
All the information about data acquirement and database can be found in \cite{NinaData_presentation1},\cite{NinaData_presentation2} and \cite{NinaData_presentation3}.

\subsection{Acquisition setup}
\begin{figure} [H]
 \centering
 \includegraphics [scale = 0.5] {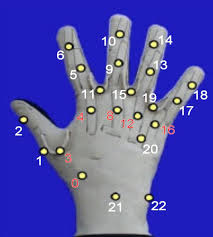}
 \caption{CyberGlove and placement of all the electrodes. Source: \cite{NinaData_presentation1}.}\label{fig:CyberGlove}
\end{figure}

The acquisition setup (\cite{NinaData_presentation1}) is composed by several devices for the recording of hand kinematics, dynamics and muscular activity. All the sensors are connected to the laptop to make the data recording possible.\\
The \textit{CyberGlove II} (CyberGlove Systems LLC, \url{www.cyberglovesystems.com}) is a motion capture data glove that takes information about hand kinematic using 22 sensors (see Figure \ref{fig:CyberGlove}). These sensors register hand and fingers motions and return 22 8-bit values of resistance. That resistance is proportional to the angles between pairs of hand joints of interest (for example metacarpum-phalanx, inter-finger, palm arch angles and so on). The average resolution is less than one degree and it depends on the size of the subject's hand.\\ 

\begin{figure} [H]
 \centering
 \includegraphics [scale = 0.7] {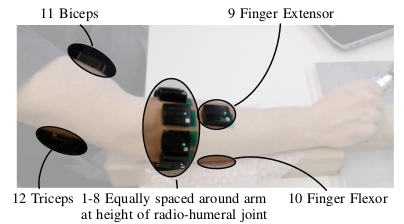}
 \caption{Placement of electrodes for sEMG signal: 8 electrodes equally spaced around the forearm, 2 electrodes for biceps and triceps brachii (they are present only in the second configuration), 2 electrodes for extensor/flexor digitorum superficialis. Source: \cite{NinaData_FeaturesExtrac40}. }\label{fig:sEMGElectrodes}
\end{figure}

A standard commercially available 2-axis \textit{IS40} inclinometer (Fritz Kübler GmbH, \url{www.kuebler.com}) is fixed to the subject's wrist to measure the wrist orientation. This device covers a range of 120\degree and it has a resolution of 0.15\degree.\\
Hand dynamics is measured by a strain gauge sensor: \textit{Finger-Force Linear Sensor} (FFLS). It records flexion and extension forces of all fingers and abduction and adduction of the thumb.\\
Muscular activity is measured using sEMG electrodes. In the first configuration of the experiment 10 \textit{MyoBock 13E200-50} electrodes (Otto Bock HealthCare GmbH, \url{www.ottobock.com}) were used. These electrodes are set to amplify the signal of about 14.000 times. They have also a shielding and filtering system in order to avoid low and high frequency interferences, for example from 50–60 \textit{Hz} power sources, mobile phones or security systems. sEMG signals are sampled at a rate of 100 \textit{Hz}. In the second configuration of the experiment 12 \textit{Trigno Wireless} electrodes (Delsys Inc, \url{www.delsys.com}) are used. sEMG signals are sampled at a rate of 2 \textit{kHz}. In both cases the equipment is fixed on the forearm using a hypoallergenic elastic latex–free band.\\
Figure \ref{fig:sEMGElectrodes} shows how electrodes are placed. In the first configuration, with 10 electrodes, eight of these are equally spaced around the forearm, at the height of the radio-humeral joint. Two are placed on the main activity spots of the flexor digitorum superficialis and of the extensor digitorum superficialis. In the second configuration two electrodes are added on the main activity spots of the biceps brachii and of the triceps brachii.

In the case of intact subjects, the recording of sEMG and kinematic data is from the same arm. In the case of amputees, sEMG signals are recorded from the missing limb while kinematic and dynamic data from the intact limb.\\

\subsection{Acquisition protocol} 
\begin{figure} [H]
 \centering
 \includegraphics [scale = 3] {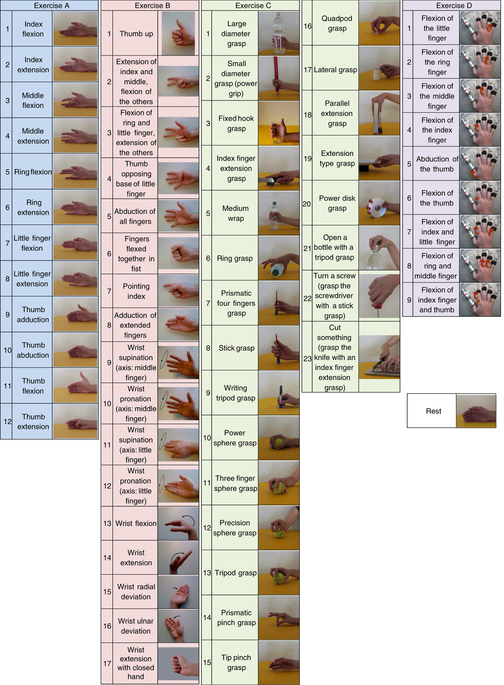}
 \caption{Exercise A (light blue): 12 basic movements of
the fingers; Exercise B (pink): 8 isometric and isotonic hand configurations and 9 basic movements of the wrist; Exercise C (green): 23 grasping and functional movements; Exercise D (purple): 9 force patterns; Rest position (white). Source: \cite{NinaData_presentation1}.}\label{fig:AllMovements}
\end{figure}

Before the beginning of data acquisition, each subject has a written and an oral explanation about the experiment and he/she has to sign an informed consent form. All the experiments are approved by the Ethics Commission of the Canton Valais (Switzerland).\\
As first step, several clinical data like age, gender, height, weight and laterality are annotated. For amputated subjects, information about date, type and reason of the amputation, remaining forearm percentage, use of prostheses and phantom limb sensation are collected. The remaining forearm percentage is the ratio between the length of the amputated forearm and the length of the intact forearm from the elbow to the wrist, rounded to the tens.\\
As second step, subjects are seated on a office chair. A laptop in front of the subjects shows with a movie the exercises that they have to perform. The intact subjects are asked to mimic movies on the screen with their dominant hand. Amputees are asked to mimic as naturally as possible the same movement with their missing limb. For amputees, generally, it is very difficult to reproduce an action with missing limb. Thus, they can simulate a task bilaterally or they can follow a visual stimulus, that could be the movie on the screen or the experimenter that performs the same movement.\\
Before starting to record data, there is a training phase. This consists of a mix of the future exercises, it is important to make the subject practical with the experiment.\\
All the performed movements are divided in four different group: A, B, C and D (see Figure \ref{fig:AllMovements}). The first group (Exercise A) is about 12 basic movements of fingers. The second (Exercise B) is about 8 hand configuration and 9 basic movements of the wrist. The third (Exercise C) is about 23 grasping movements and it involves everyday objects that are presented to the subject to mimic daily-life actions. The fourth (Exercise D) is about 9 force pattern, it consists in a press combinations of fingers with an increasing force.\\
The movements are chosen from the literature about taxonomy, robotics and rehabilitation.
Each movement is alternated by rest posture to avoid muscular fatigue. The expected execution time of each movement is about 5 \textit{s}, of rest posture is about 3 \textit{s}. The sequence of movements is not randomized in order to favour unconscious movements.\\

\subsection{Data collections} \label{sec:Ninapro_dataset}
The NinaPro database, acquired with the setup and procedure explained in the previous section, consists of three sub-datasets, according to the devices used and the subjects' characteristics.\\
The first database contains data from 27 intact subjects: 7 females and 20 males, 2 left handed and 25 right handed with 28 $\pm$ 3.4 years. The electrodes used for sEMG data acquisition are the 10 \textit{Otto Bock}. The subjects perform the exercise A,B and C described in the previous section. Each movement has been repeated 10 times and each repetition is interspaced by the rest posture.\\
The second database contains data obtained from 40 intact subjects: 11 females and 29 males, 5 left handed and 35 right handed with 29.9 $\pm$ 3.9 years. The electrodes used for sEMG data acquisition are the 12 \textit{Delsys}. The subjects perform the exercise B,C and D described in the previous section. Each movement has been repeated 6 times and each repetition is interspaced by the rest posture.\\
The third database contains data obtained from 11 trans-radial amputated subject: 11 male, 1 left handed and 10 right handed with 42.36 $\pm$ 11.96 years. The subjects perform the exercise B,C and D described in the previous section. Each movement has been repeated 6 times and each repetition is interspaced by the rest posture.\\
All the details about the datasets are reported in Table \ref{tab:Dataset}.

\begin{table}[H]
	
	\resizebox{15cm}{!}	{
	
	\begin{tabular} {|| l | c | c | c ||}
	
	\hline
	\hline
	  & Database 1 & Database 2 & Database 3\\ 		\hline
	\hline
	Intact Subjects & 27 & 40 & 0 \\    \hline
	Trans-radial Amputated Subjects & 0 & 0 & 11 \\   \hline
	sEMG Electrodes & 10 Otto Bock  & 12 Delsys & 12 Delsys\\ \hline
	Total Number of Postures (rest included) & 53 & 50 & 50 \\ \hline
	Number of Movement Repetitions & 10 & 6 & 6\\ \hline	
	\hline
	Exercise 1 & Exercise A & Exercise B & Exercise B\\ \hline
	Number of Movements & 12 & 17 & 17\\ \hline
	\hline
	Exercise 2 & Exercise B & Exercise C & Exercise C\\ \hline
	Number of Movements & 17 & 23 & 23\\ \hline	
	\hline
	Exercise 3 & Exercise C & Exercise D & Exercise D\\ \hline
	Number of Movements & 23 & 9 & 9\\ \hline		 	  
	\end{tabular}
	
	}
	
	\caption{Description of database 1, 2 and 3.}
	
\label{tab:Dataset}
\end{table}

Before the raw data could be used for classification, several steps are necessary. The first step is the filtering: the Delsys sEMG signals, that are not shielded against power line interferences, are cleaned from 50 Hz power-line interference using a Hampel filter (\cite{Kuzb_FeaturesExtrac27}). The second step is the synchronization: it is a linear interpolation or nearest-neighbour interpolation of data. The third step is the relabelling: it is a correction of movements label by maximizing the likelihood of a rest-movement-rest sequence. In fact the movements performed by the subjects may not perfectly match with the stimuli proposed, due to human reaction times and experimental conditions.\\
The raw data are not online but are available upon request.\\
For each subject and exercise one can find online (\url{http://ninapro.hevs.ch/}) a \textit{Matlab} file with the following data:

\begin{itemize}
\item subject: subject number;
\item exercise: exercise number;
\item emg (10 or 12 columns): sEMG signal of the 10 or 12 electrodes. Columns from 1 to 8 include the signal from 8 electrodes around the forearm, columns 9 and 10 include the signal of the muscle Flexor/Extensor Digitorum Superficialis, columns 11 and 12 include signal of the muscle Biceps/Triceps Brachii;
\item acc (36 columns): three-axes acceleration values of the 12 electrodes (only for database 1 and 2);
\item glove (22 columns): signal from the 22 sensors of the Cyberglove;
\item inclin (2 columns): signal from the 2 axes inclinometer (of the wrist);
\item stimulus (1 column): the original label of the movements repeated by the subject (before the relabelling phase);
\item restimulus (1 column): label of the movements a-posteriori (after the relabelling phase);
\item repetition (1 column): repetition of the stimulus;
\item rerepetition (1 column): repetition of restimulus;
\item force (6 columns): force recorded during the third exercise of database 2 and 3;
\item forcecal (2 $\times$ 6 values): minimal and maximal force values for each sensor;
\end{itemize}

The database 3 presents some exceptions. The amputated subjects 7 and 8 had only 10 electrodes instead of 12 due to insufficient space. The amputated subjects 1, 3 and 10 asked to interrupt the experiment before its end due to fatigue or pain. They performed respectively 39, 49 and 43 postures (including rest).\\

\section{Data representation: features extraction} \label{sec:Feat_ext}
Online sEMG data require some processes to make them available for classification. Thus, we want to extract from them useful information to achieve this goal. This process is called features extraction. It consists in several steps of preprocessing in order to clean up data from the component of noise and in choosing of the algorithm for data representation.\\
For a complete reading about this approach we refer to \cite{NinaData_FeaturesExtrac40} and \cite{Kuzb_FeaturesExtrac27}.\\
The described approach and processes are universally accepted in this field.\\
In this thesis we work with data of Exercise 1 (17 hand configuration and movements of the wrist) of the second and third database. Thus the reported values of parameters are referred to this analysis.

\subsection{Preprocessing}
\begin{figure} [H]
\centering
\includegraphics [scale=0.40]{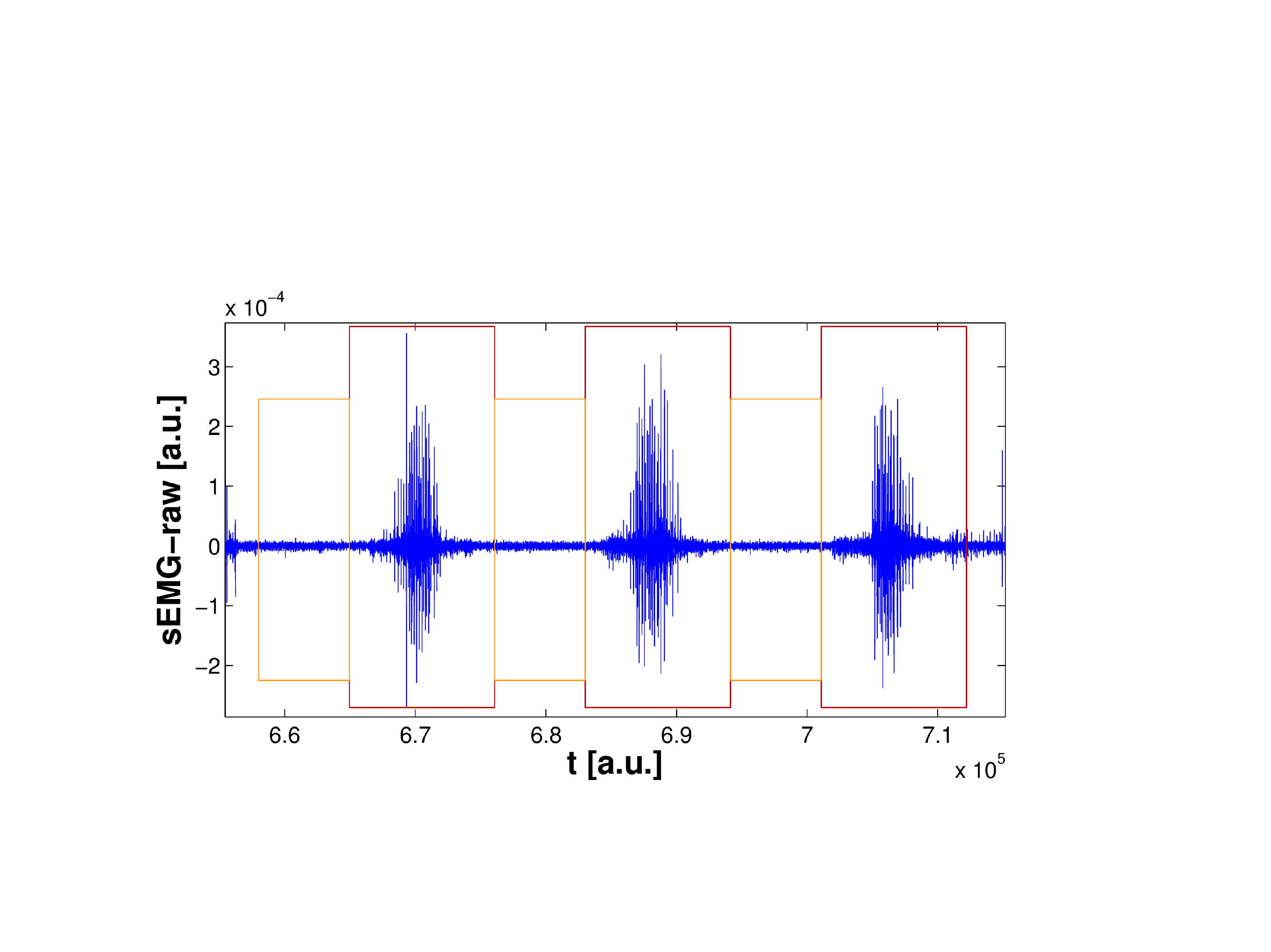}
\caption{Movements division process for a little part of total raw signal. Orange squares contain rest postures, red squares contain a movements.}\label{fig:MovementsDivision}
\end{figure}

The initial data consists of a single continuous signal (one for each electrode) that contains the information about the sequence that goes from the first movement to the last, interspaced by rest posture.\\
The first step consists of the division of a movement from another. Signals with different labels are separated. Each resulting matrix identifies different movements or rest, the columns of this matrix are the channels (10 for first database, 12 for second and third). In Figure \ref{fig:MovementsDivision} is shown the movements division process.\\

\begin{figure}  [H]
\centering
\subfigure
   {\includegraphics[scale=0.39]{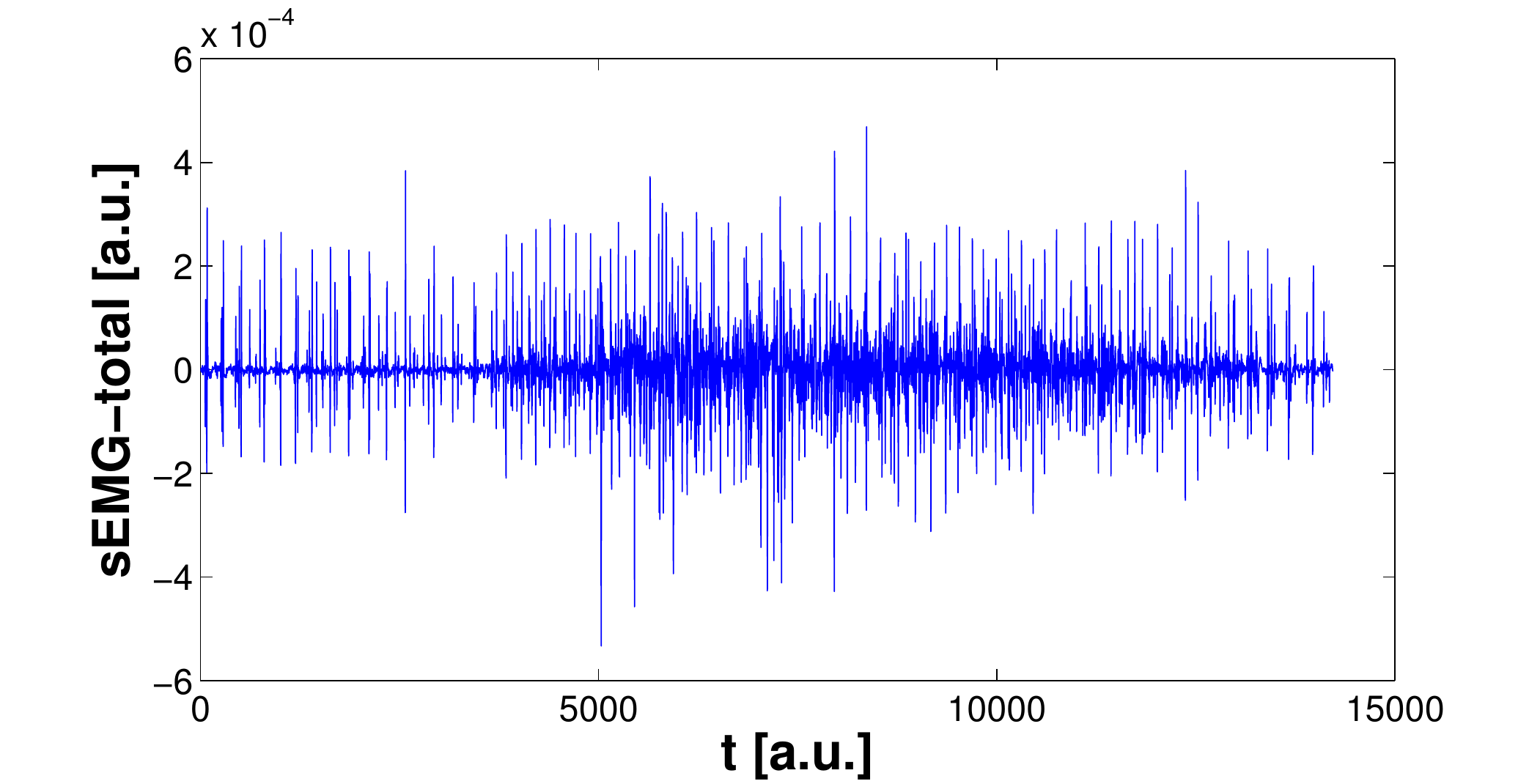}}
\hspace{5mm}
\subfigure
   {\includegraphics[scale=0.29]{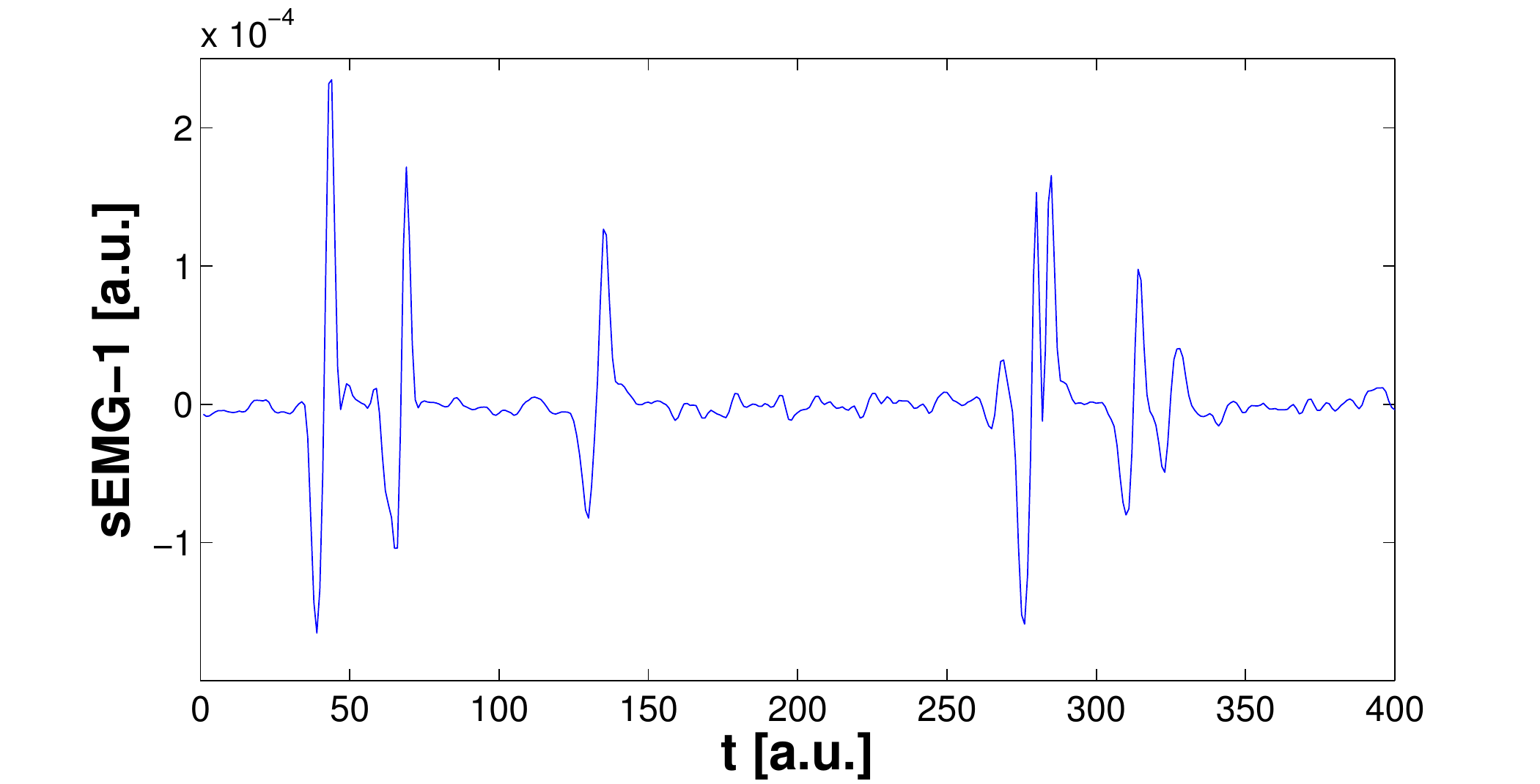}}
 \qquad
 \subfigure
   {\includegraphics[scale=0.29]{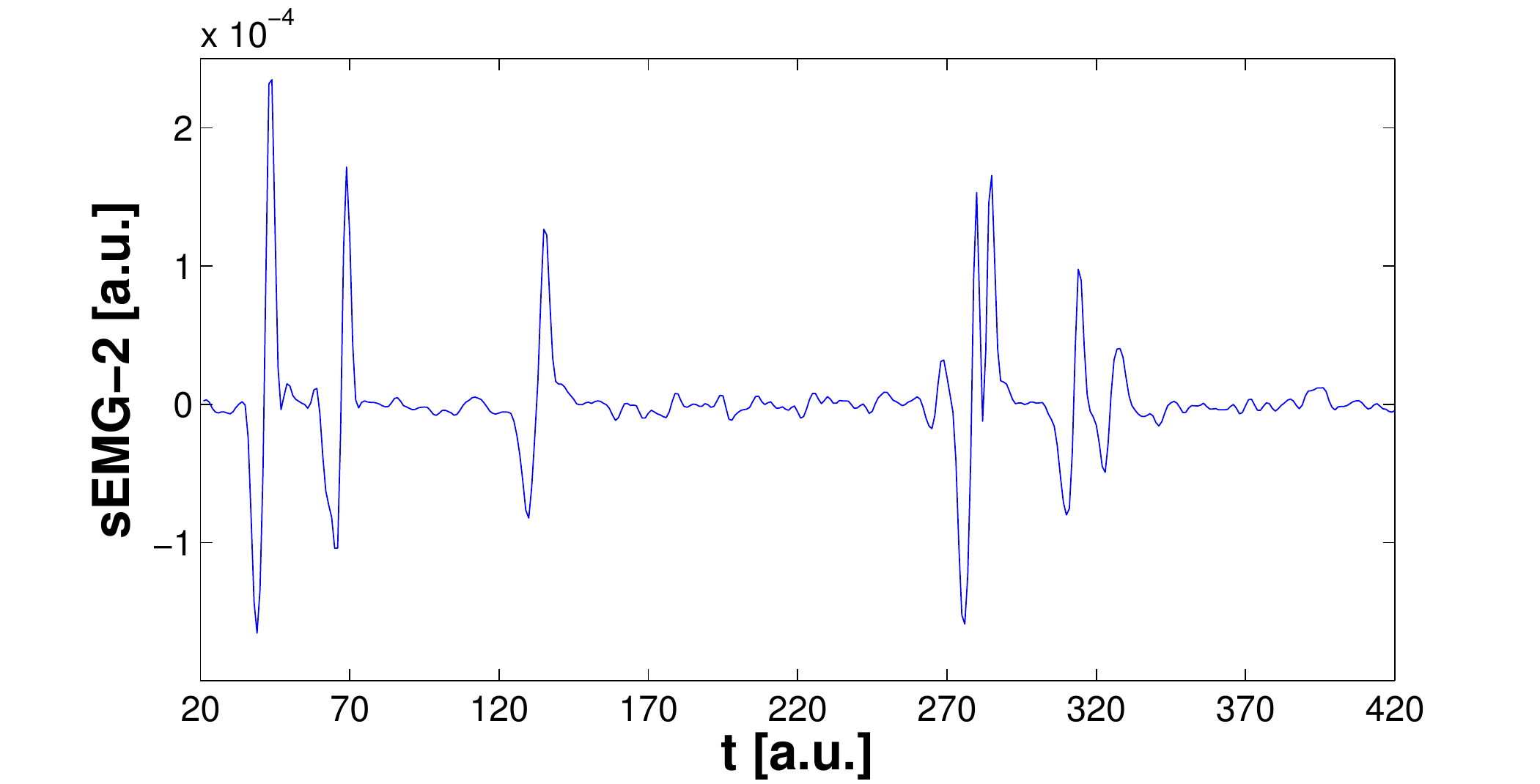}}
\hspace{5mm}
 \subfigure
   {\includegraphics[scale=0.29]{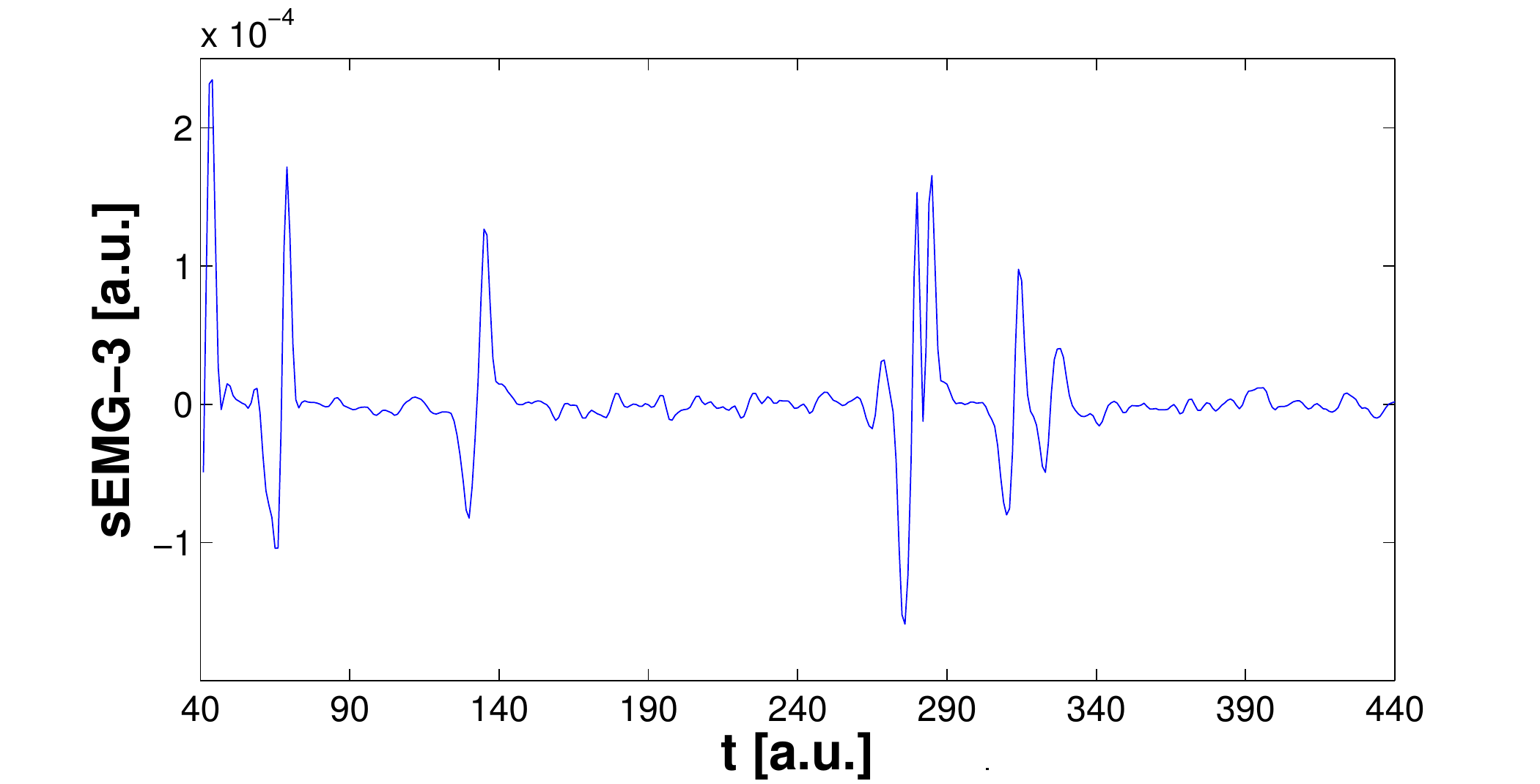}}
   \qquad
 \subfigure
   {\includegraphics[scale=0.29]{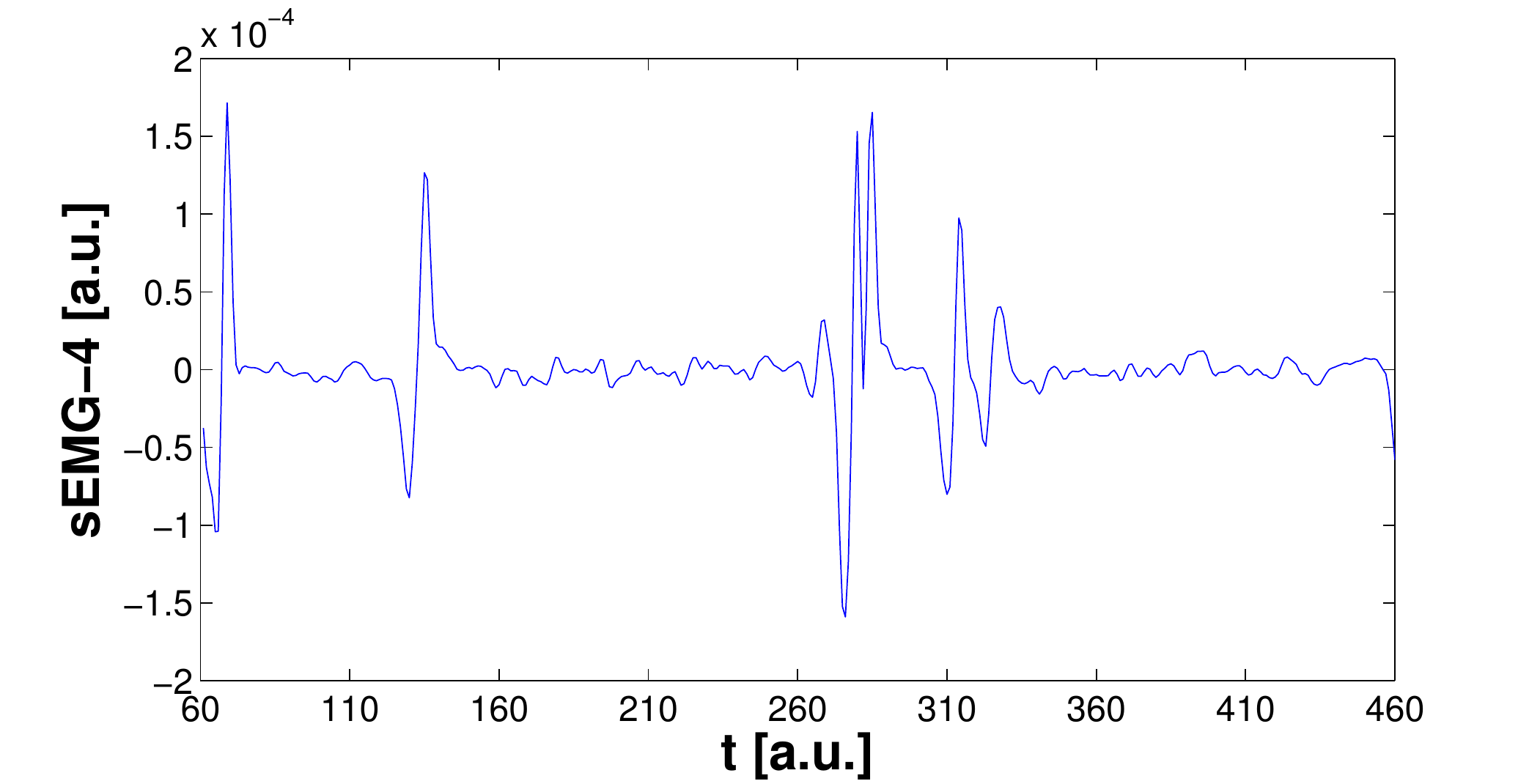}}
\caption{Windowing of signal from channel 1. Up: Total signal of a single movement. Down: Windowed signal after 1, 2, 3 and 4 steps.}\label{fig:Windowing}
\end{figure}

The second operation is called windowing. Each matrix from the previous point is divided in overlapping windows of fixed length. The windows length used in the literature is of 100 \textit{ms}, 200 \textit{ms} and 400 \textit{ms}, the shift considered is of 10 \textit{ms} (\cite{Kuzb_FeaturesExtrac27}). Thus, between windows there is an overlap of $N - 10 \textit{ms}$, where $N$ is the window length considered.\\
All the parameters are analysed in the literature. The experiments of \cite{Kuzb_FeaturesExtrac27} show that a window length of 200 \textit{ms} or 400 \textit{ms} results in higher accuracy. Indeed, if a window is long enough, the error of label on the edge of a movement can be reduced. In this work the used windows length is $N = 200 \textit{ms}$ (i.e. 400 samples) and the increment of the sliding window is 10 \textit{ms} (i.e. 20 samples).\\
An example of windowing is shown in Figure \ref{fig:Windowing}.\\

As said before, each movement is repeated a fixed number of times. The third passage consists of the split of the signal in test and training part, depending on number of repetitions considered.\\ 
For the second and third database the repetitions $\lbrace$ 1, 3, 4, 6 $\rbrace$ are used as training, the others ($\lbrace$ 2, 5 $\rbrace$) as test.\\
At this point the training part can be sub-sampled to reduce the amount of data and to achieve a computationally feasible problem. The training set is reduced by keeping every 10th sample.\\

\subsection{Feature extraction}
The last step is the choice of the algorithm for the data representation.\\ 
There are some experimental evidence that we can consider in sEMG processing operation. First, there is a quasi-linear relation between Root Mean Square (RMS) amplitude of sEMG signal and force exerted by a muscle (\cite{Kuzb_FeaturesExtrac27}, \cite{TimeDomFE}). Second, sEMG spectral characteristics might be related to conduction velocity of muscle fibers (\cite{Kuzb_FeaturesExtrac27}, \cite{FreqDomFE}). If a time domain algorithm is used the first aspect is privileged. If a frequency domain algorithm is used the second aspect is favoured.\\
The most important algorithms are reported in Table \ref{table:FeatureExt}. We refer to $\hat{x}$ as a feature computed from a signal $x$ of length $T$, where its elements are indexed as $x_{t}$.

\begin{table}[H]
	
	\resizebox{13,7cm}{!}	{
	
	\begin{tabular} {|| p{7cm} | p{5cm} | c ||}
	
	\hline
	\hline
	\textbf{Feature}  & \textbf{Definition (per channel)} & \textbf{Dimension} \\ 		\hline
	\hline
	Mean Absolute Value (MAV) & \(\displaystyle \hat{x} = \frac{1}{T} \sum\limits_{t=1}^T \mid x_{t} \mid \) & C  \\    \hline
	Variance (Var) & \(\displaystyle \hat{x} = \frac{1}{T} \sum\limits_{t=1}^T (x_{t} - \bar{x} )^{2} \) & C  \\   \hline
	Waveform Length (WL) & \(\displaystyle \hat{x} =  \sum\limits_{t=1}^T \mid x_{t} - x_{t+1} \mid \) & C  \\ \hline
	sEMG Histogram  & \(\displaystyle \hat{x}_{1:B} = hist(x_{1:t},B) \) & CB \\ \hline
	Cepstral Coefficients & \(\displaystyle \hat{x}_{k} = \mathcal{F^{-1}}(log \mid \mathcal{F}(x_{1:t}) \mid )_{k} \) & CT  \\ \hline	
	Short-Time Fourier Transform (STFT) & \(\displaystyle \hat{x}_{k,t} = \sum\limits_{m=0}^{R-1} x_{m-1} g_{m} e^{-\textit{i} \tiny\frac{2 \pi}{M} k_{m}}\) & CMT  \\ \hline
	marginal Discrete Wavelet Transform (mDWT) & \(\displaystyle \hat{x}_{t} = \sum\limits_{\tau=0}^{T/2^{l}-1} \mid \sum\limits_{t=1}^{T} x_{t} \psi_{l,\tau}(t) \mid \psi_{l,\tau}(t) = 2^{-\frac{m}{2}} \psi(2^{-l}t-\tau) \) & CL \\ \hline
	 	  
	\end{tabular}
	
	}
	
	\caption{Algorithm for features extraction. C is the number of channels (10 or 12 depending on database). B is the number of bins in the histogram of sEMG Histogram method. For STFT, M is the number of windows in which one divides each part of the signal already divided in windowing step. These are indexed with $k$ and computed over blocks obtained by a sliding window function $g$ of length R. For mDWT, one uses $\psi_{l,t}$ to denote the mother wavelet with translation $l$. Source: \cite{Kuzb_FeaturesExtrac27}.}
	
\label{table:FeatureExt}
\end{table}

\begin{figure}  [H]
\centering
\subfigure
   {\includegraphics[scale=0.4]{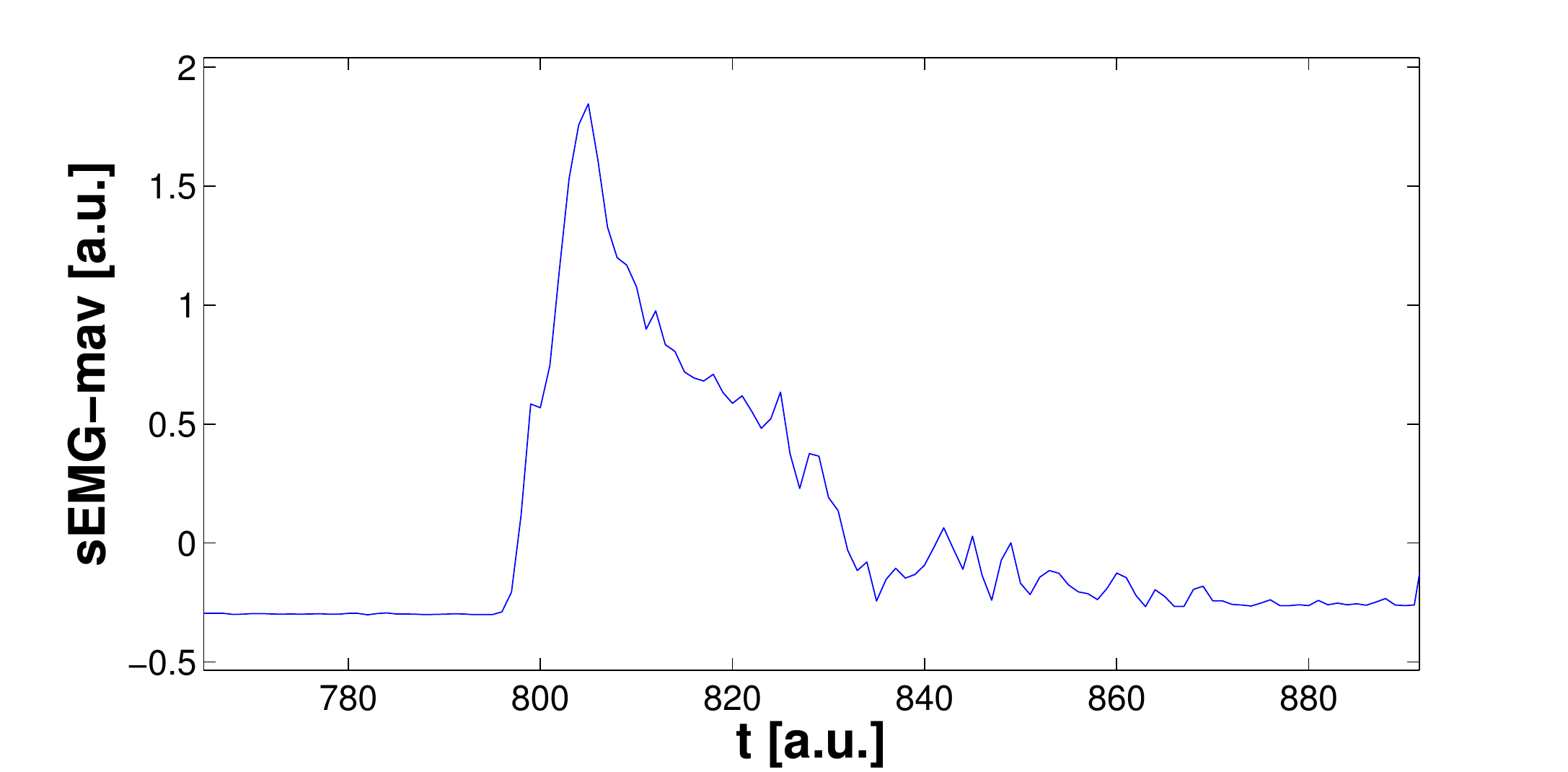}}
 \hspace{5mm}
 \subfigure
   {\includegraphics[scale=0.4]{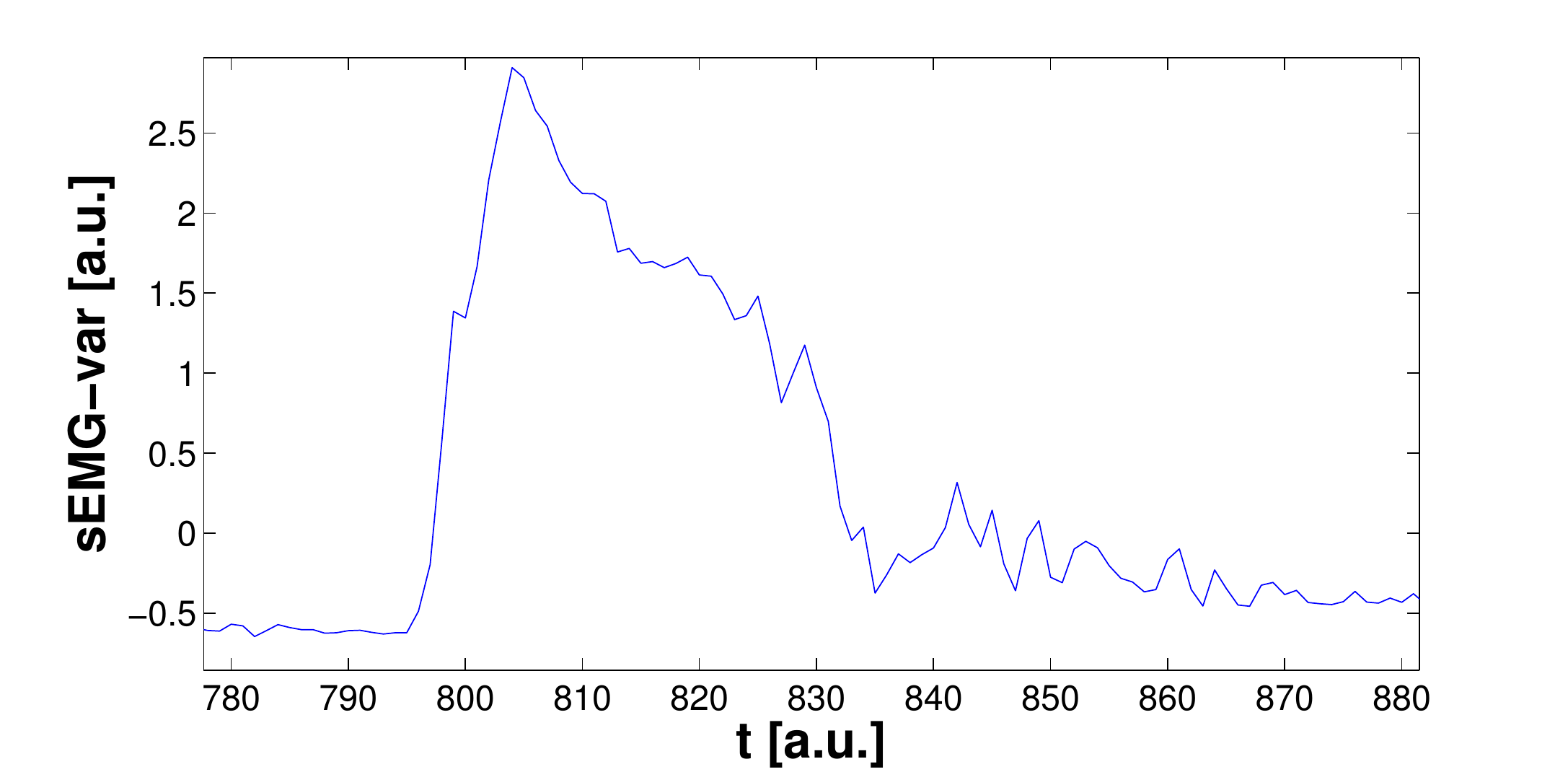}}
\hspace{5mm}
 \subfigure
   {\includegraphics[scale=0.4]{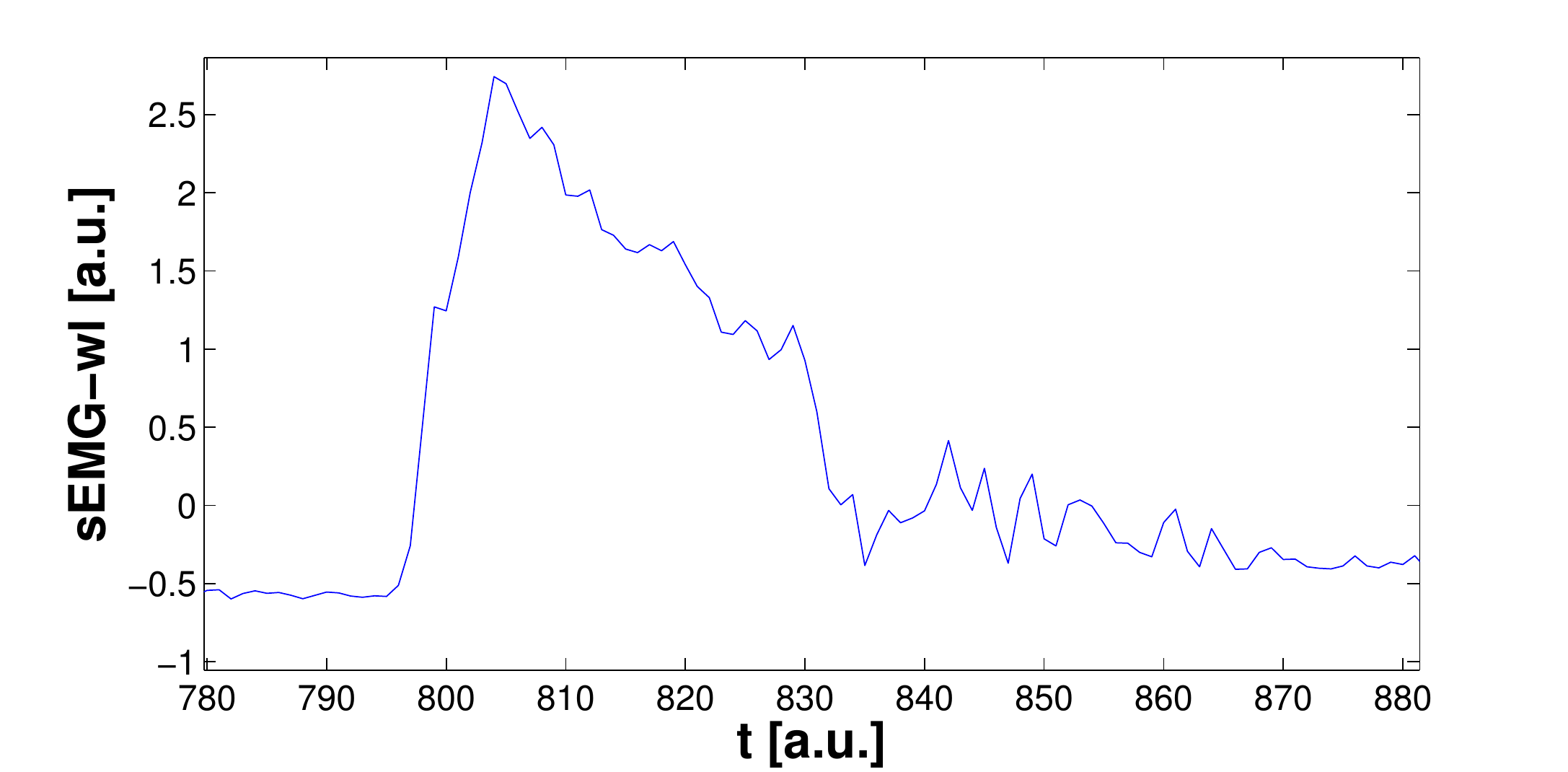}}
\caption{Representation of movement "Extension of index and middle, flexion of the others" from channel 1.}\label{fig:FeaturesExample}
\end{figure}

The authors of \cite{Kuzb_FeaturesExtrac27} compared the previous different algorithms for features extraction.\\ 
From these studies emerge that, in the time domain, Mean Absolute Value and Waveform Length have similar performance to the computationally more demanding marginal Discrete Wavelet Transform. In the frequency domain Short Time Fourier Transform is more robust than simple Fourier Transform when dealing with non-stationary signal.\\ 
In this thesis we work in time domain with the mean between Mean Absolute Value, Variance and Waveform Length to be as independent as possible from the method. In Figure \ref{fig:FeaturesExample} is shown the representation of a movement after features extraction.

\begin{figure} [H]
\centering 
\subfigure
   {\includegraphics[scale=0.4]{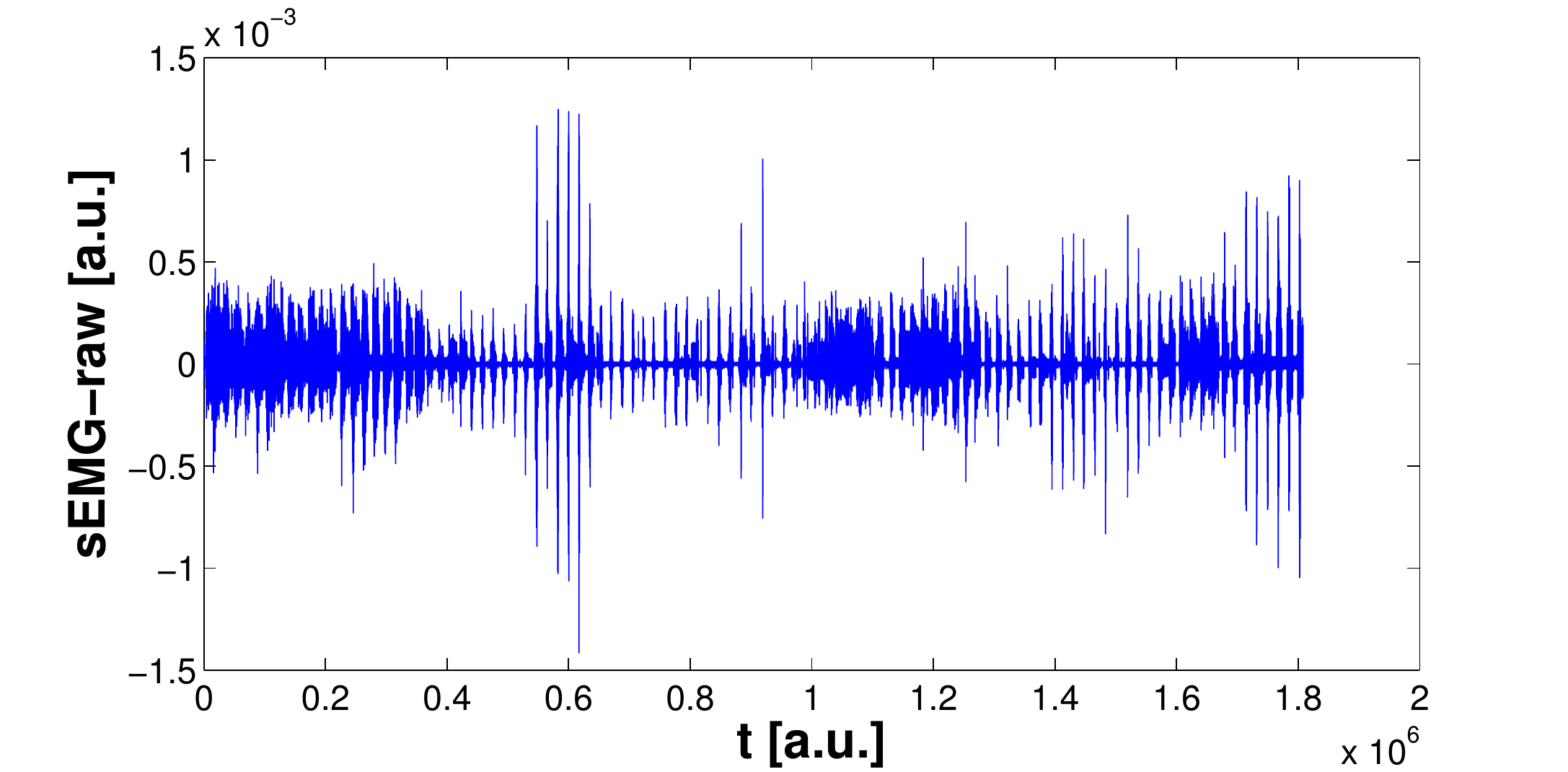}}
 \hspace{5mm}
 \subfigure
   {\includegraphics[scale=0.4]{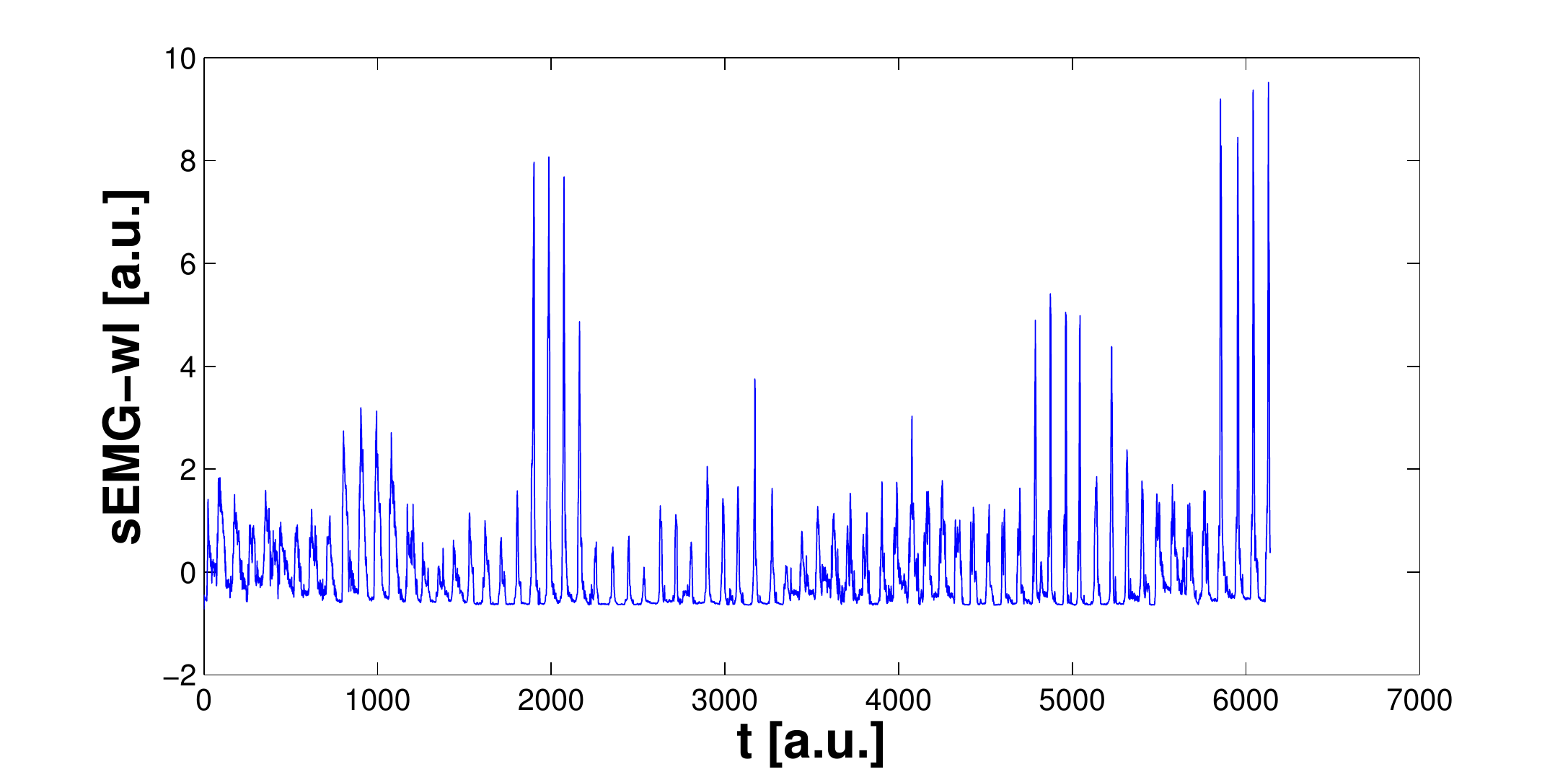}}
   \caption{On top: raw data. On bottom: signal after processing and features extraction.}\label{fig:SignalBefAftProcess}
\end{figure}

In Figure \ref{fig:SignalBefAftProcess} we show the total signal before and after processing and feature extraction. 

At the end of the feature extraction step, we have a set of vectors with dimension equal to the number of channels. The output of each vector is one of the possible movements inside the three exercises with the adding of rest. The number of training vectors and test vectors for each subject are respectively of the order of $3 \cdot 10^{3}$ and $2 \cdot 10^{4}$.\\

\clearpage{\pagestyle{empty}\cleardoublepage}
\chapter{From theory to learning algorithms} \label{cha:Algorith}

After extracting features, we have a set of data to classify. In this chapter we present the state of art learning algorithms that solve classification problems and the theory on which they are based.\\

The section \ref{sec:Mat_frame} is an introduction about the mathematical framework and background required to understand the rest of the work.\\
In section \ref{sec:Sup_learn_SD} we focus on the traditional form of supervised learning, where we exploit only the information from the target domain. It is the supervised learning on a single domain.\\
In section \ref{sec:SVM} we introduce Support Vector Machines (SVMs). It is one of the most used and theoretically well motivated methods in machine learning scenario. It represents the state of the art of supervised learning on a single domain.\\
In section \ref{sec:Sup_learn_DD} we introduce a new form of supervised learning, where we exploit information from target domain and source domains. It is the supervised learning on different domains.\\
Section \ref{sec:DA} begins with a general introduction about different proposed directions to tackle the problem of domain adaptation. In the last part of the section we introduce the state of the art algorithms for supervised learning on different domains. We explain the theory on which each algorithm is based and the main structure of the code with the help of flow charts. All the mentioned algorithms are implemented with $Matlab^{\circledR}$.\\

\section{Mathematical framework and background} \label{sec:Mat_frame}
In this section we present mathematical tools necessary to tackle the rest of the work. In the following we indicate column vectors with small bold letters, e.g. $\textbf{a} = [a_{1},a_{2},...,a_{N}]$ and matrices with capital bold letters, e.g. $\textbf{A} \in \mathbb{R^{M \times N}} $. With $A_{i,j}$ and $\textbf{A}_{i}$ we denote respectively the element $(i,j)$ and the \textit{i}-th column of matrix $\textbf{A}$.\\
In the rest of the work we refer to $\textbf{x}_{i} \in \mathcal{X}$ as an input vector of a learning algorithm and to $y_{i} \in \mathcal{Y}$ as its associated output. $\mathcal{X} \subseteq \mathbb{R^{d}}$ and $\mathcal{Y} = \mathbb{R}$ are respectively the input space and the output space.\\ 
We denote with $D = \lbrace\textbf{x}_{i},{y}_{i}\rbrace_{i=1}^{N} $ a set of data that come from an unknown probability distribution. The goal of a learning algorithm is to find a function that, for any future input vector $\textbf{x}_{i}$, can determine the best corresponding output $y_{i}$.\\
In a classification problem, like the one analysed in this work, $y_{i}$ can assume values from a finite set that are the possible output classes. If the possible outputs are only two we have a binary problem, if these are more we have a multiclass problem. In this thesis we have the second type of problem and classes are equal to the possible hand postures. In the following we refer to $G$ as the total number of classes, or possible output label.\\

\paragraph{Loss function.} In general we can define a loss function as a function that associates a real number to a given event in order to represent some “cost" associated with the event. In a classification problem it is the function that represents the price paid for a misclassification. For a binary problem let us refer to $y_{i}$ and $\tilde{y}_{i}$ as, respectively, the true and the predicted label of a vector $\textbf{x}_{i}$. The loss function can be defined as:

\begin{equation}
\label{eq:LossBinaryProb}
l(\tilde{y}_{i},y_{i}) = \text{max} \lbrace 0,(1- \tilde{y}_{i} y_{i})^{p} \rbrace,
\end{equation}
\\where different choices of parameter $p$ lead to different penalties for misclassified vectors.\\
In a multiclass problem, let us define with $\tilde{\textbf{Y}}_{i}$ the column of the confidences for each class referred to a given vector $\textbf{x}_{i}$. The loss function is modified as follow:

\begin{equation}
\label{eq:LossMulticlass}
l(\textbf{Y}_{i},\tilde{\textbf{Y}}_{i}) = \text{max} \lbrace 0, 1-\tilde{\textbf{Y}}_{y_{i},i} + \underset{g \neq y_{i}}{\text{max}} \lbrace \tilde{\textbf{Y}}_{g,i} \rbrace \rbrace.
\end{equation}
\\The loss is zero if the confidence value for the correct class is at least greater than one with respect to the other confidence values. Otherwise, the loss is linearly proportional to the differences between the confidence values of real class and the maximum confidence referred to another class.\\
The goal of a learning problem is to minimize this risk.\\

\section{Supervised learning on a single domain} \label{sec:Sup_learn_SD}
We refer to training samples and test samples as a set of vectors, or items, for which we know the output classification, i.e. the label value.\\ 
In a supervised learning problem the first ones are used in the training phase to build a classification model. The second ones are exploited in the test phase to evaluate the performance of a learning algorithm from the agreement between the predicted labels and their real values. Thus, the learner receives a set of labelled examples as training data and makes predictions for all unseen points.\\
In the following we refer to $N$ as the total number of training vectors.\\
We define the target problem as the new classification problem that we aim to solve. In our case it is the new subject that learns to perform some hand postures.\\
In the simplest case, in order to build a classification model for the new target subject, we can exploit only the training data of the target himself. Thus, data used to train the model and data used to test it come from the same domain, i.e. the target domain. Taking this into account, it is reasonable to assume that the data present the same distribution.\\
We refer to this process as a supervised learning on a single domain: in fact in order to build the model, we use available labelled data (i.e. supervised process) from only the target subject (i.e. single domain).
In section \ref{sec:SVM} we present the Support Vector Machines (SVMs) method. It represents the state of art of supervised learning on a single domain for classification problem.\\

\section{Support Vector Machines} \label{sec:SVM}
In this section we introduce one of the most effective classification algorithms in machine learning: Support Vector Machines (SVMs). The goal of SVMs is to produce a model, based on the training data, able to predict the output of a given test vector.\\
For a complete reading about topics, one can refer to \cite{Mohri}, \cite{Shalev-Shwartz} and \cite{Burger_SVM_Kernel}.\\

\subsection{Linear SVMs} \label{sec:L-SVM}
\begin{figure} [H]
 \centering
 \includegraphics [scale = 0.55] {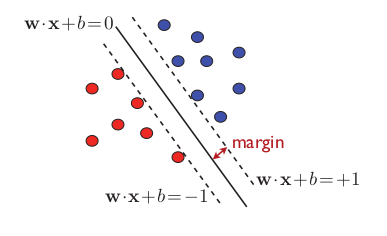}
 \caption{Separating hyperplane and margins. Source: \cite{Mohri}.}\label{fig:SeparableCase}
\end{figure}

We start with the introduction of linear SVMs for the solution of a binary problem. We refer to the two types of vectors as positive and negative ones, i.e. $ y_{i} \in \mathcal{Y} = \lbrace 1;-1 \rbrace $.\\ 
The aim of a classification problem is to find an hyperplane which separates positive from negative vectors. For a point that lies on this hyperplane the following equation holds: 

\begin {equation}
\label{eq:hyperp}
\textbf{w} \cdot \textbf{x} + b = 0,
\end {equation}
\\where \textbf{w} identifies the hyperplane (it is the direction vector of the plane) and $\frac{\mid b \mid}{\parallel \textbf{w} \parallel} $ is the normal distance from the hyperplane to the origin.\\
Usually there are many hyperplanes that a learning algorithm can choose to solve a problem. The SVMs algorithm returns the hyperplane with the maximum margin \textit{m}. The margin is the maximum distance between the separating hyperplane and the closest points of each sets.\\
All these concepts are reported in Figure \ref{fig:SeparableCase}.

If we define marginal hyperplanes as $ \textbf{w} \cdot \textbf{x} + b = \pm 1 $, the following inequalities hold :

\begin {equation}
\label{eq:hyperp1}
\begin{aligned}
\textbf{w} \cdot \textbf{x}_{i} + b \geq 1&&  &\text{     for  $y_{i} = 1$}, \\
\textbf{w} \cdot \textbf{x}_{i} + b \leq -1&&  &\text{     for  $y_{i} = -1$}. \\
\end{aligned}
\end {equation}
\\The vectors that lie on marginal hyperplanes are called support vectors.
The equations \ref{eq:hyperp1} can be combined in order to obtain a unique inequality:

\begin {equation}
\label{eq:hyperp1_const}
y_{i} (\textbf{w} \cdot \textbf{x}_{i} + b) - 1 \geq 0 .
\end {equation}
\\Using equations of marginal hyperplanes we can find that $m = \frac{1}{\parallel \textbf{w} \parallel}$, thus it is clear that searching the hyperplane that maximizes the margin is equivalent to minimize $\parallel \textbf{w} \parallel$ with constrain (\ref{eq:hyperp1_const}).

\begin{figure} [H]
 \centering
 \includegraphics [scale = 0.55] {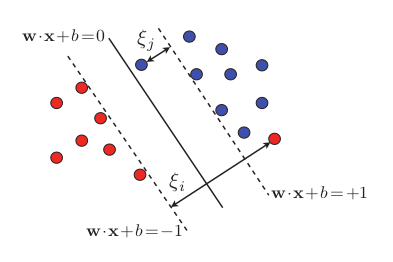}
 \caption{Separating hyperplane and margins in a non-separable classification problem. Source: \cite{Mohri}.} \label{fig:NonSeparableCase}
\end{figure}

In most of classification problems, data present a component of noise, an example is shown in Figure \ref{fig:NonSeparableCase}.

To take into account the possible noise in the data, we can introduce the slack variables $\xi_{i}$. This variable measures the distance by which vector $\textbf{x}_{i}$ violates the desired inequality: $y_{i} (\textbf{w} \cdot \textbf{x}_{i} + b) \geq 1$.\\ 
At this point we can define the minimization problem to solve as:

\begin {equation}
\label{eq:objfunc_const_primalNS}
\begin{aligned}
&\underset{\textbf{w},b,\boldsymbol{\xi}}{\text{min}} \dfrac{1}{2} \parallel \textbf{w} \parallel^{2} + \dfrac{C}{p} \sum\limits_{i=1}^{N} \xi_{i}^{p}\\
&\text{subject to:}\quad y_{i}(\textbf{w} \cdot \textbf{x}_{i} + b) \geq 1 - \xi_{i} \quad \wedge \quad \xi_{i} \geq 0, \quad \forall \textit{i} \in [1,N].\\
\end{aligned}\end {equation}
\\This can be formulated as a Lagrangian problem as follows:

\begin {equation}
\label{eq:primal_lagNS}
L (\textbf{w},b,\boldsymbol{\alpha},\boldsymbol{\xi},\boldsymbol{\beta}) = \dfrac{1}{2} \parallel \textbf{w} \parallel^{2} + \dfrac{C}{p}\sum\limits_{i=1}^N \xi_{i}^{p} - \sum\limits_{i=1}^N \alpha_{i} [y_{i} (\textbf{x}_{i} \cdot \textbf{w}_{i} + b) - 1 + \xi_{i}] - \sum\limits_{i=1}^N \beta_{i} \xi_{i}.
\end {equation}
\\First and second terms of sum are the objective function that we want to minimize. The others are the constraints multiplied by different Lagrangian multipliers $\boldsymbol{\alpha}$ and $\boldsymbol{\beta}$. In this optimization problem, we minimize the amount of slack variables and $\parallel \textbf{w} \parallel$ . Generally, these two requests are conflicting. The parameter that sets this trade-off is C. The parameter $p$ describes different penalties for misclassified vectors as we explain in the end of this paragraph.\\
Until objective function and constrains are convex and differentiable, the KKT conditions can be applied at the optimum: 

\begin{align} 
  \label{eq:KKTcondNS1}
  \boldsymbol{\bigtriangledown}_{\textbf{w}}L_{p} &=0 \> \Rightarrow \> \textbf{w} = \sum\limits_{i=1}^N \alpha_{i} y_{i} \textbf{x}_{i} ,\\
  \label{eq:KKTcondNS2}
  \boldsymbol{\bigtriangledown}_{b}L_{p} &=0 \> \Rightarrow \>  \sum\limits_{i=1}^N \alpha_{i} y_{i} = 0 ,\\
  \label{eq:KKTcondNS3}
  \boldsymbol{\bigtriangledown}_{\xi_{i}}L_{p} &=0 \> \Rightarrow \> \alpha_{i}+\beta_{i} = C \xi_{i}^{p-1} ,\\
  \label{eq:KKTcondNS4}
  \forall \textit{i},\quad \alpha_{i}[y_{i}(\textbf{w} \cdot \textbf{x}_{i} + b)-1 + \xi_{i}] &=0 \> \Rightarrow \> \alpha_{i} = 0 \vee y_{i}(\textbf{w} \cdot \textbf{x}_{i} + b) = 1 - \xi_{i} ,\\
\label{eq:KKTcondNS5}  
  \forall \textit{i}, \quad \beta_{i}\xi_{i} &= 0 \> \Rightarrow \> \beta_{i} = 0 \vee \xi_{i} = 0.
\end{align} 
\\The first equation shows that $\textbf{w}$ is a liner combination of training vectors. The forth equation indicates that the vectors that really appear in that combination are only the support vectors. Indeed for other vectors the Lagrangian multiplier is zero. In this case we have two different types of support vectors. The first ones are vectors that lie on margin hyperplane (they have $\xi_{i} = 0$). The second ones are called outlier and are the misclassified vectors for which $\xi_{i} \neq 0$.\\
The task of determine a classifier is equivalent to find the parameters $\textbf{w}$ and $b$ of the model. In the test phase, given a new vector $\textbf{x}$ with unknown label, the output hypothesis is:

\begin {equation}
\label{eq:TestNS}
h(\textbf{x}) = \text{sgn}(\textbf{w} \cdot \textbf{x} + b) = \text{sgn}(\sum\limits_{i=1}^N \alpha_{i} y_{i}(\textbf{x}_{i} \cdot \textbf{x}) + b ).
\end {equation}
\\The second equality can be simply obtained using equation (\ref{eq:KKTcondNS1}).\\
The extension of the theory to the multiclass case, i.e. $y_{i} \in \mathcal{Y} = \lbrace 1,2,...,G \rbrace$, is straightforward. We solve different optimization problems, one for each class $g=[1,2,...,G]$. Thus, during the training phase, a pair $(\textbf{w}_{g},b_{g})$ is found for each class. The considered class takes the label $y = 1$ and the others $y = -1$ (one-vs-all approach), in this way we solve $G$ different binary problems.\\ 
In the test phase, given a vector $\textbf{x}$, we choose the class $g$ that gives the maximum output according to: 

\begin {equation}
\label{eq:TestNS-KMultiClass}
h(\textbf{x}) = \underset{g \in G}{\text{argmax}} (\textbf{w}_{g} \textbf{x} + b_{g}) =\underset{g \in G}{\text{argmax}} \left( \sum\limits_{i=1}^N \alpha_{i,g}y_{i} \textbf{x}_{i} \textbf{x} + b_{g} \right).
\end {equation}

\begin{figure} [H]
 \centering
 \includegraphics [scale = 0.5] {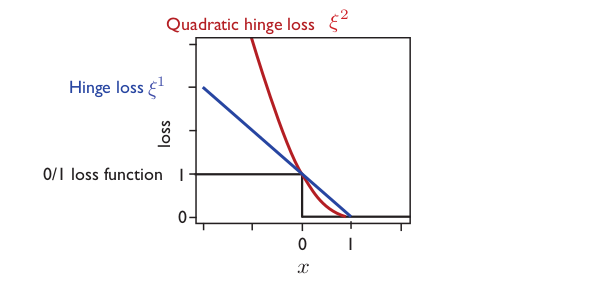}
 \caption{The hinge loss and the quadratic hinge loss compared to the zero-one loss. Source: \cite{Mohri}.}\label{fig:LossFunctions}
\end{figure}

\paragraph{Loss Function.} As anticipated in section \ref{sec:Mat_frame} the loss function is the penalty associated to a misclassified vector.\\
For a binary problem the most common loss functions are the hinge loss and the quadratic hinge loss. These are respectively associated to $p = 1$ and $p = 2$ in equation (\ref{eq:LossBinaryProb}). As shown Figure \ref{fig:LossFunctions}, penalties are different depending on the value of $p$ chosen.\\ 

We have no loss if the prediction falls in the right part of the hypersurface (i.e. distance between point and hyperplane is greater than 1). When the prediction falls into the margin we have a loss $0 \leq l \leq 1$. If the loss is greater than one the prediction is in the wrong part of the hypersurface.\\

\subsection{Non-linear SVMs} \label{sec:NL-SVM}
\begin{figure} [H]
\centering
 \includegraphics [scale = 6] {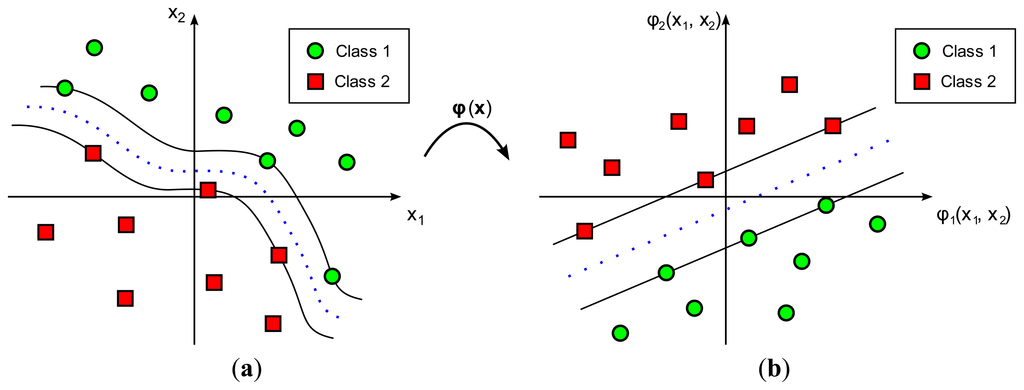}
 \caption{Kernel application to a non-linear SVM classifier.}\label{fig:FromNonLinearToLinearHyperplane}
\end{figure}

Until now, a linear decision boundary is used to solve a classification problem. However there are techniques that extend SVM algorithms in order to define non-linear decision boundaries. Kernel methods are an example of these.\\
A way to define a non-linear decision boundary is to use a non-linear map function defined as: $\Phi(\textbf{x}): \mathcal{X} \longmapsto \mathcal{H}$. $\mathcal{X} \subseteq \mathbb{R^{d}}$ is the input space of training vectors and $\mathcal{H}$ is a high-dimensional
(possibly infinite-dimensional) Hilbert space called feature space. In other words, if the training vectors are not linearly separable in the input space we can replace each of them with $\Phi(\textbf{x}_{i})$ to make the problem solvable by changing the metric. An example is shown in Figure \ref{fig:FromNonLinearToLinearHyperplane}.\\

However, determining the hyperplane solution requires multiple computations of inner products in high-dimensional spaces: $\langle \Phi(\textbf{x}) \cdot \Phi(\textbf{x}) \rangle$ instead of the simple product between training vectors. A solution to this problem is to use kernel methods.\\ 
A kernel over $\mathcal{H}$ is a function $K: \mathcal{X} \times \mathcal{X} \longmapsto \mathcal{R}$. The idea is to define a kernel K such that:

\begin {equation}
\label{eq:K}
\forall \textbf{x},\textbf{x}^{'} \in \mathcal{X}, K(\textbf{x},\textbf{x}^{'}) = \langle \Phi(\textbf{x}) \cdot \Phi(\textbf{x}^{'}) \rangle.
\end {equation}
\\An inner product is a measure of the similarity between two vectors, thus $K$ can be interpreted as a similarity measure between elements of the input space $\mathcal{X}$. $K$ is often significantly less complex to compute than $\Phi(\textbf{x})$ and its inner product. But there is another bigger advantage: under precise conditions (\textit{Mercer's conditions}) we can work directly with $K$ without knowing $\Phi(\textbf{x})$.\\
Now given a vector $\textbf{x}$ with unknown label, the output hypothesis is:

\begin {equation}
\label{eq:TestNS-K}
h(\textbf{x}) = \text{sgn}(\sum\limits_{i=1}^N \alpha_{i} y_{i} K(\textbf{x}_{i}, \textbf{x}) + b ).
\end {equation}
\\One of the most used kernel is the Gaussian one, or Radial Basis Function (RBF):

\begin {equation}
\label{eq:GausKer}
K(\textbf{x}^{'}, \textbf{x}) = e^{-\dfrac{\parallel \textbf{x}^{'} - \textbf{x} \parallel^{2} }{2 \sigma^{2}}} = e^{- \gamma \parallel \textbf{x}^{'} - \textbf{x} \parallel^{2}}.
\end {equation}
\\It is also the kernel chosen in the computer's implementation of algorithms used in this work. For completeness, we underline that there are many kinds of kernel like: linear, polynomial and sigmoid.

\paragraph{Least Square Support Vector Machines (LS-SVM).} The LS-SVM (\cite{LS-SVM}) problem is a new formulation of equation (\ref{eq:objfunc_const_primalNS}) in order to avoid the high computational burden of the original SVM problem. In this formulation we replace the inequality constraints with equality. The quadratic hinge loss ($p = 2$) and the kernel method are used.\\
The optimization problem is written as follow: 

\begin {equation}
\label{eq:objfunc_const_dual}
\begin{aligned}
&\underset{\textbf{w},b}{\text{min}} \dfrac{1}{2} \parallel \textbf{w} \parallel^{2} + \dfrac{C}{2} \sum\limits_{i=1}^N \xi_{i}^{2}\\
&\text{subject to:}\quad  y_{i} = \textbf{w}\phi(\textbf{x}_{i})+b+\xi_{i}, \quad \forall \textit{i} \in [1,N],\\
\end{aligned}\end {equation}
\\The solution of the reformulated Lagrangian problem is characterized by a linear system which takes a similar form as the linear system of standard SVM (see equations from \ref{eq:KKTcondNS1} to \ref{eq:KKTcondNS5}).\\
Let us indicate with $\textbf{I}$ the identity matrix and with $\textbf{K}$ the kernel matrix defined as:

\begin{equation}
\label{eq:K_matrix}
    \textbf{K} = \begin{bmatrix}
        K(\textbf{x}_{1},\textbf{x}_{1}) & K(\textbf{x}_{1},\textbf{x}_{2}) & \dots\\
          & \ddots & \\
        \dots &   & K(\textbf{x}_{N},\textbf{x}_{N}) \\
     \end{bmatrix}
\end{equation}
\\The solution of the problem is given by:

\begin{equation}
\label{eq:NoTrSol1}
    \begin{bmatrix}
        \boldsymbol{\alpha} \\
        b \\
    \end{bmatrix}   
    =  \textbf{P} \begin{bmatrix}
        \textbf{y} \\
        0 \\
    \end{bmatrix},
\end{equation}
\\where $\textbf{P} = \begin{bmatrix} \textbf{K}+\frac{1}{c}\textbf{I} & \textbf{I} \\ \textbf{I}^{T} & 0 \\ \end{bmatrix}^{-1} $ and \textbf{y} is the vector of outputs.\\
The LS-SVM solution that we obtain now presents a difference with respect to the original SVMs. Sparsity is lost and all data contribute to the model.

\section{Supervised learning on a different domain} \label{sec:Sup_learn_DD}
Up to now we have dealt with the traditional learning. One of the main assumptions on which it is based is that the training data, used to learn the target model, and the test data, used to test this model, come from the same distribution. However in many real problems this assumption is not true: this happens because data can be dependent on dynamic factors like time, acquisition devices and space (\cite{Tom_thesis}).\\ 
We tackle a similar problem when we want to learn a model with few available labelled data. To avoid the creation of a not very solid model we can exploit also data coming from different domains (for example from different data acquisitions or databases, gathered with different devices and so on). For example, we can have a lot of labelled data on a source problem and the need to solve the same problem for a different target domain with few labelled data. If source and target present a distribution mismatch the source data can't be used directly to solve the target problem, hence a form of adaptation is needed.\\


Let us define the concept of source as a known classification model built with a great number of training vectors. In a domain adaptation problem we aim to solve a new target problem exploiting information from a source domain. The classification problem of target and source is the same. In particular we have the same output label set, $Y^{s} = Y^{t}$. However, the target domain $D^{t}$ and the source domain $D^{s}$ are different in terms of marginal data distribution, $P^{s}(X) \neq P^{t}(X)$. Thus, the conditional probability distribution $P(y| \textbf{x})$ might be slightly different for source and target (\cite{Tom_thesis}).\\
When we attempt to leverage over existing source knowledge to solve a target problem, we are combining information from different domains: the target domain and all the available sources domains. We can consider the source as a type of prior knowledge that, after an appropriate adaptation, can be used to learn something for new target problem.\\

\begin{figure} [H]
\centering
\includegraphics[scale=0.4]{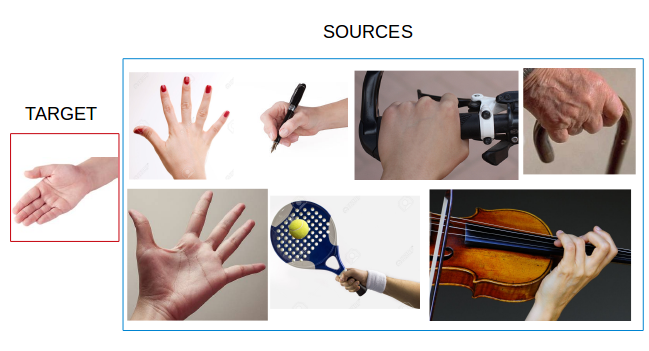} 
\caption{Each new target subject presents a distribution mismatch with respect to source subjects due to gender, age, hobbies, work, dominant hand, etc.}\label{fig:DAHand}
\end{figure}

The application on prosthetic hands treated in this work is an example of domain adaptation problem. In particular, the task is to catalogue different hand's postures of new target subject using, as sources, subjects that already performed the same movements. We refer about domain adaptation problem because biological signals of the same movement can be different for different users in terms of marginal data distribution. In fact, the EMG signal from different subjects varies depending on gender, age, muscular activity, dominant hand, position of the electrodes during the data collection and so on (see Figure \ref{fig:DAHand}).\\

The goal of a domain adaptation algorithm is to find the best way to exploit useful informations from the available sources, in order to solve the new target problem.\\
In the following we refer to $K$ as the total number of sources.\\

\section{Domain adaptation algorithms} \label{sec:DA}

\begin{figure} [H]
 \centering
 \includegraphics [scale = 0.50] {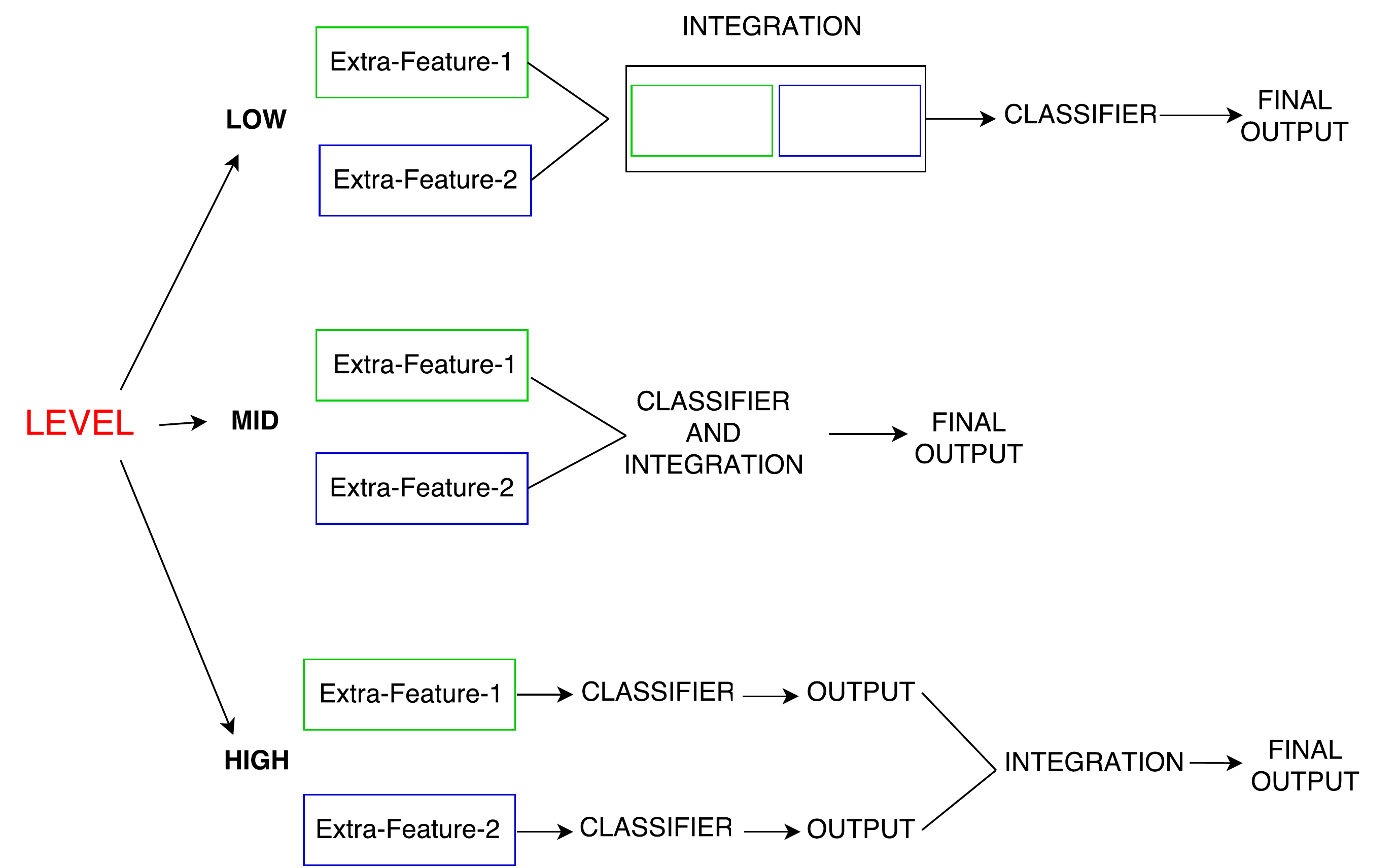}
 \caption{Example with only two sources of high, mid and low level of features integration. Source: \cite{Tom_CueInt_Medical}.}\label{fig:CueIntegration}
\end{figure}

We present here several different algorithms that use adaptive learning in order to boost the training phase of a prosthetic hand with the exploitation of previous experience. Each methods try to combine in different ways informations from target domain (represented by training labelled vectors) and from domains of sources (represented by source classification models).\\ 
In the following it is explained how informations coming from different sources can be combined. Each algorithm adds in different ways informations given by training vectors of target and this particular aspect is treated in the sessions dedicated to individual algorithms.\\
Some methods exploit the prior knowledge directly. These use a combination of the parameters ($\textbf{w},b$) of source models in order to obtain the new parameters of new model solving an optimization problem (see algorithm in section \ref{sec:MuliAdaAlg}).\\
In other methods the sources are seen like experts, or feature extractors. The outputs obtained using source's models for the classification of training data are considered as a descriptor, or as an extra-features. At this point there are different methods to combine informations derived from different sources. We can discern three different levels of integration: low, mid and high.\\
In the low level the extra-features, coming from each source, are combined into a single new vector. For this feature vector a classifier (for example a linear or non-liner SVMs) is trained. The different information from different sources influence in the same way the final decision. In fact, no weight factor is used.\\
In the middle level the extra-features of each source are kept separately, but a single classifier is used for the final hypothesis. This classifier is based on SVM algorithm with a new kernel that is a liner combination of kernels of sources appropriately weighed (see algorithm in section \ref{sec:MKAL-Alg}).\\
In the high level a classifier is trained for each extra-feature vectors. Each classifier produces a final hypothesis about training vectors. All these assumptions are combined together to produce the final output (see algorithm in section \ref{sec:H-L2LAlg}).\\ 
The different ways of integration explained above are called cue integration methods (\cite{Luo_thesis} and \cite{Tom_CueInt_Medical}) and are summarised in Figure \ref{fig:CueIntegration}.

With respect to the domain adaptation algorithms chosen, there are two baseline. 
The first is given by the solution of a classification problem using only the training data of the target. This method is called \textit{No Transfer} and it solves non-linear LS-SVM problem trained on the target data only (see section \ref{sec:NL-SVM}). The second is represented by the exploitation of training data of the source only. This method is called \textit{Prior Features} and it solves a linear LS-SVM problem trained on the source data only (see section \ref{sec:L-SVM}).\\

In the rest of the section we present the algorithms used in the simulations of this work for the classification problem of hand movements. The main structure of the code is the same for all of them. We have a training phase and a test phase. In the first one, the classification problem is tackled for each target subject and models are built. For each subject the models are calculated several times for an increasing number of training vectors. In the second phase models are tested using new vectors. In the following we explain the training phase of each algorithms. It is the point that distinguish one adaptive methods from the others.\\


\subsection{Multi Adapt algorithm} \label{sec:MuliAdaAlg}
\begin{figure} [H]
 \centering
 \includegraphics [scale = 0.45] {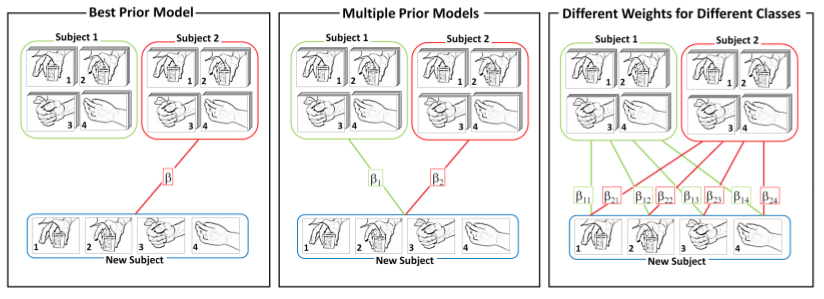}
 \caption{Best prior model: only a subject is take into account to create a new model for the new subject. Multiple prior models: a weighted linear combination of sources is used to built the new model. Different weights for different classes: it is similar to previous case but the classes of each source are weighted in different way. Source: \cite{Tom_thesis}.}\label{fig:MultiAdaptCases}
\end{figure}

The first form of \textit{Multi Adapt} algorithm was proposed in \cite{Orabona_DAforEMG} and after extended in \cite{Tom_hand_pros}. It solves a classification problem exploiting a combination of source models already known and properly weighted.\\
An optimization problem based on non-linear SVMs with Gaussian kernel is solved for every source. The result is a vector $ \hat{\textbf{w}} $ for each of these.\\
We begin to treat a binary problem to simplify the notation, in the end we extend the results to a multiclass problem.\\
As a starting point, it is possible to use one of the pre-trained models to build the new model for the new subject. The basic idea is to search a solution for the new problem that is close to the pre-trained one. The optimization problem that we tackle is the following:

\begin {equation}
\label{eq:OptimMultiAdaBest}
\begin{aligned}
&\underset{\textbf{w},b}{\text{max}}\> \dfrac{1}{2}\parallel \textbf{w}-\beta \hat{\textbf{w}} \parallel^{2}+\dfrac{C}{2} \sum\limits_{i=1}^N \xi_{i}^{2} \\
&\text{subject to:}\quad y_{i} = \textbf{w} \cdot \phi(\textbf{x}_{i}) + b + \xi_{i} .
\end{aligned}\end {equation}
\\As described in Section \ref{sec:SVM} we can use the Lagrangian formulation that leads to the following optimality conditions:

\begin{align} 
  \label{eq:KKTcondMA1}
  &\boldsymbol{\bigtriangledown}_{\textbf{w}}L =0 \> \Rightarrow \> \textbf{w} = \beta \hat{\textbf{w}} + \sum\limits_{i=1}^N \alpha_{i} \phi(\textbf{x}_{i}), \\
  \label{eq:KKTcondMA2}
  &\boldsymbol{\bigtriangledown}_{b}L =0 \> \Rightarrow \>  \sum\limits_{i=1}^N \alpha_{i} = 0, \\
  \label{eq:KKTcondMA3}
  &\boldsymbol{\bigtriangledown}_{\xi_{i}}L =0 \> \Rightarrow \> \alpha_{i} = C \xi_{i}, \\
  \label{eq:KKTcondMA4}
  &\boldsymbol{\bigtriangledown}_{\alpha_{i}}L =0 \> \Rightarrow \> \textbf{w} \cdot \phi(\textbf{x}_{i}) + b + \xi_{i} - y_{i} = 0,
\end{align} 
\\where $\boldsymbol{\alpha} = [\alpha_{1},\alpha_{2},...,\alpha_{N}]^{T}$ is the vector of Lagrangian multipliers. As in non-linear SVM algorithm, the new model $\textbf{w}$ is given by a combination of feature functions $\phi(\textbf{x}_{i})$. In addition, it appears the pre-trained model $\hat{\textbf{w}}$ weighted by $\beta$.\\
Previous equations can be combined and written in matrix form:

\begin{equation}
\label{eq:MulAdaSol}
    \begin{bmatrix}
        \textbf{K}+\frac{1}{c}\textbf{I} & \textbf{I} \\
        \textbf{I}^{T} & 0 \\
    \end{bmatrix}
    \begin{bmatrix}
        \boldsymbol{\alpha} \\
        b \\
    \end{bmatrix}   
    =  \begin{bmatrix}
        \textbf{y} - \beta \hat{\textbf{y}} \\
        0 \\
    \end{bmatrix},
\end{equation}
\\where $\textbf{y} = [y_{1},y_{2},...,y_{N}]^{T}$ is the vector that contains the real label of each training vector and $\hat{\textbf{y}} = [\hat{\textbf{w}}\phi(\textbf{x}_{1}),\hat{\textbf{w}}\phi(\textbf{x}_{2}),...,\hat{\textbf{w}}\phi(\textbf{x}_{N})]$ is the vector of labels predicted using the model $\parallel \hat{\textbf{w}} \parallel$. $\textbf{K}$ is the kernel matrix introduced in \ref{eq:K_matrix} and $\textbf{I}$ is the identity matrix.\\
Inverting equation (\ref{eq:MulAdaSol}) the solution is:

\begin{equation}
\label{eq:MulAdaSol2}
    \begin{bmatrix}
        \boldsymbol{\alpha} \\
        b \\
    \end{bmatrix}   
    =  \textbf{P} \begin{bmatrix}
        \textbf{y} - \beta \hat{\textbf{y}}\\
        0 \\
    \end{bmatrix},
\end{equation}
\\where $\textbf{P}$ is the inverse of the first matrix on the left of equation (\ref{eq:MulAdaSol}).\\
In order to evaluate $\beta$, the leave-one-out error is introduced. Let us define with $\tilde{y}_{i}$ the prediction obtained on \textit{i}-th sample when it is removed from the training set. Starting from equation (\ref{eq:MulAdaSol2}) and writing the Lagrangian multiplier as $\boldsymbol{\alpha} = \boldsymbol{\alpha}^{'} + \beta \boldsymbol{\alpha}^{''}$, it can be shown that the prediction vector has the following equation:

\begin {equation}
\label{eq:pediction-LOU}
\tilde{y}_{i} = y_{i} - \dfrac{\alpha^{'}_{i}}{P_{i,i}} + \beta \dfrac{\alpha^{''}_{i}}{P_{i,i}}.
\end {equation}
%
\\Thus, given a set of training vectors, the weight $\beta$, that gives the minimum distance between true and predicted label, is chosen in agreement with:

\begin {equation}
\label{eq:min_loss-LOU}
\underset{\beta}{\text{min}} \sum\limits_{i=1}^{N} l(\tilde{y}_{i}, y_{i}),
\end {equation}
where $l(\tilde{y}_{i}, y_{i})$ is a loss function.\\
The method described above exploits only one source, although many of them are available. The best source is selected by evaluating the minimal leave-one-out error.\\
The previous approach can be extended in case of multiple sources by defining the following learning problem:

\begin {equation}
\label{eq:OptimMultiAdaMulti}
\begin{aligned}
&\underset{\textbf{w},b}{\text{max}}\> \dfrac{1}{2}\parallel \textbf{w}-\sum\limits_{k=1}^K \beta^{k} \hat{\textbf{w}}^{k} \parallel^{2}+\dfrac{C}{2} \sum\limits_{i=1}^N \xi_{i}^{2} \\
&\text{subject to:}\quad y_{i} = \textbf{w} \cdot \phi(\textbf{x}_{i}) + b + \xi_{i} ,
\end{aligned}\end {equation}
\\where $K$ are the number of sources and $\boldsymbol{\beta} = [\beta_{1},\beta_{2},...,\beta_{K}]^{T}$ is the vector of the importance weight of each prior. Now the optimal solution is composed by a linear weighted combination of previous models:

\begin {equation}
\label{eq:w_MultiAdaMulti}
\textbf{w}= \sum\limits_{k=1}^{K} \beta^{k}\textbf{w}^{k} + \sum\limits_{i=1}^{N} \alpha_{i} \phi(\textbf{x}_{i}).
\end {equation}
\\The parameters of the new model are given by:

\begin{equation}
\label{eq:MulAdaSol2Multi}
    \begin{bmatrix}
        \boldsymbol{\alpha} \\
        b \\
    \end{bmatrix}   
    =  \textbf{P} \begin{bmatrix}
        \textbf{y} - \sum\limits_{k=1}^{K} \beta^{k} \hat{\textbf{y}}^{k}\\
        0 \\
    \end{bmatrix},
\end{equation}
\\where $\hat{\textbf{y}}^{k} = [\hat{\textbf{w}}^{k}\phi(\textbf{x}_{1}),\hat{\textbf{w}}^{k}\phi(\textbf{x}_{2}),...,\hat{\textbf{w}}^{k}\phi(\textbf{x}_{N})]^{T}$ is the vector of labels predicted using the $k$-th source model. As in the previous case, $\beta_{k}$ can be evaluated with live-one-out error, the only difference is that now the minimization problem (\ref{eq:min_loss-LOU}) is with respect to a vector ($\boldsymbol{\beta}$).\\
The last case is the more general and it is an extension to the multiclass problem. It consists in the use of different weights for different classes of the same source. Indeed, in a problem of domain adaptation, it can be reasonable to think that a subject learns better a task from a source and another task from on different source.\\ 
Starting from the previous cases, the new implementation is straightforward. Now the weight parameters are inside a matrix $\textbf{B}=\mathbb{R^{K \times G}}$. $K$ and $G$ are respectively the number of sources and classes, thus $\beta_{k,g}$ is the weight associated to class $g$ of source $k$. 
Given a sample $\textbf{x}_{i}$ we refer respectively to $\textbf{Y}_{i}$, $\hat{\textbf{Y}}_{i}$ and $\tilde{\textbf{Y}}_{i}$ as the column vector with real output (in one-vs-all approach), output predicted by prior and output predicted with live-one-out. The loss function used to evaluate the live-one-out prediction is reported in equation (\ref{eq:LossMulticlass}).\\ 
The three different cases are shown in Figure \ref{fig:MultiAdaptCases}.

\begin{figure} [H]
 \centering
 \includegraphics [scale = 0.53] {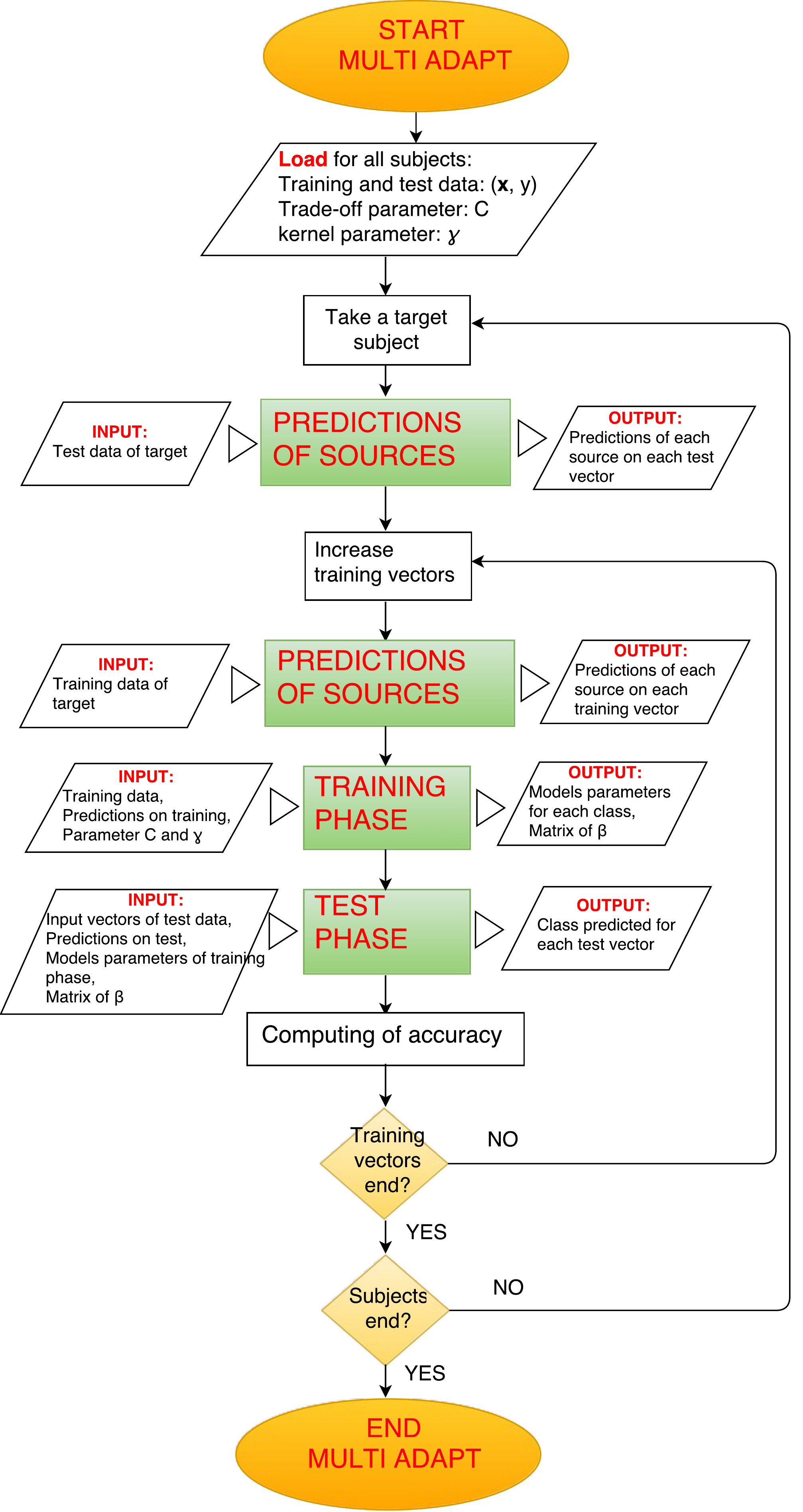}
 \caption{Flow chart of code of \textit{Multi Adapt}.}\label{fig:MultiAdapt}
\end{figure}

In this work we use the case of different weights for different classes.\\ 
During the test phase, for each test vector $\textbf{x}$, we choose the class $g$ that gives the maximum output in according with:

\begin {equation}
\label{eq:TestMultiClassMultiAdapt}
h(\textbf{x}) = \underset{g}{\text{argmax}} \left( \textbf{w}_{g} \phi(\textbf{x}) + b_{g} + \sum\limits_{k=1}^K B_{k,g} \hat{Y}_{k,g} \right).
\end {equation}
\\The flow char of the code that implements \textit{Multi Adapt} method is shown in Figure \ref{fig:MultiAdapt}. For each subject and class, during the training phase, is solved the optimization problem with different weights for different classes in order to find the parameters of new models. 

\subsection{Multi Kernel Adaptive Learning algorithm} \label{sec:MKAL-Alg}
\begin{figure} [H]
 \centering
 \includegraphics [scale = 0.50] {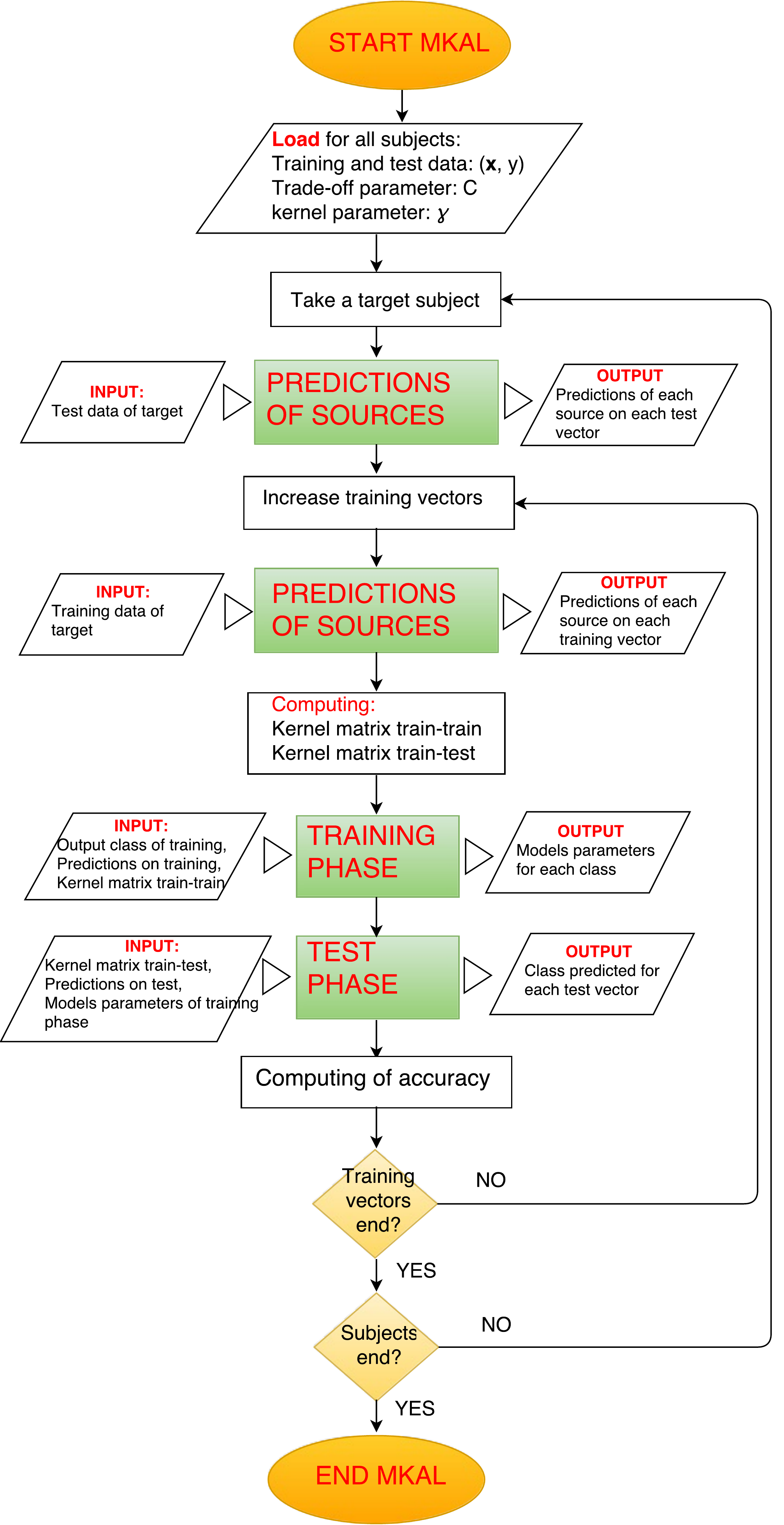}
 \caption{Flow chart of code of \textit{Multi Kernel Adaptive Learning}.}\label{fig:MKAL}
\end{figure}

There are several types of algorithms that exploit a kernel combination to obtain a better performance in test phase. The authors of \cite{Orabona_MKL_1}, \cite{Orabona_MKL_2} present the \textit{Multi Kernel Adaptive Learning} (MKAL) algorithm for a multiclass problem. This is described in detail also in \cite{Luo_thesis}.\\
The MKAL algorithm belongs to middle level of features integration, it is based on a linear combination of kernels. Kernels may be different for type, or these may be of the same type with different parameters. The weights of this linear combination reflect the importance of each kernel.\\
In order to introduce this algorithm, we define its final prediction as:

\begin{equation}
\label{eq:pred_MKAL}
h(\textbf{x}) = \underset{g}{\text{argmax}} \sum\limits_{k=1}^K \beta^{k}_{g} s^{k}(\textbf{x},y),
\end{equation}
\\where the index $g$ runs over classes and the index $k$ over sources. $\beta^{k}_{g}$ are the weights of the score function $s^{k}(\textbf{x},y)$ defined as:

\begin{equation}
\label{eq:score_func}
s^{k}(\textbf{x},y)=\textbf{w}^{k} \phi^{k}(\textbf{x},y) = \textbf{w}^{k}_{g} \phi^{' k}_{g}(\textbf{x}).
\end{equation}
\\In the previous equation $\textbf{w}^{k} = [\textbf{w}^{k}_{1},\textbf{w}^{k}_{2},...,\textbf{w}^{k}_{G}]$ is the hyperplane referred to source $k$. It is composed by $G$ blocks, one for each class. The feature map $\phi^{k}(\textbf{x},y) = [0,..,0,\phi^{' k}_{g}(\textbf{x}),0,...,0]$ is represented as a vector with all the elements null except that for the $g$-th position, that corresponds to the output class of vector $\textbf{x}$. The kernel corresponding to the source $k$ is defined as: $K^{k}((\textbf{x},y),(\textbf{x}^{'},y^{'}))$.\\
Let us define with: 

\begin{equation}
\label{eq:WBar_PhiBar}
\begin{aligned}
&\bar{\textbf{w}} = [\textbf{w}^{0},\textbf{w}^{1},\textbf{w}^{2},...,\textbf{w}^{K}] \quad \text{and} \\
&\bar{\phi}(\textbf{x},y) = [\phi^{0}(\textbf{x},y),\phi^{1}(\textbf{x},y),\phi^{2}(\textbf{x},y),...,\phi^{K}(\textbf{x},y)]
\end{aligned}
\end{equation}
\\respectively the concatenated vectors of hyperplanes and feature maps. These are both composed by $(K + 1)$ blocks: one for each source plus the block labelled with $0$ that is referred to the original training vectors.\\
Let as define also the (2,$p$) group norm $\parallel \bar{\textbf{w}} \parallel_{2,p} $ of vector $\bar{\textbf{w}}$ as:

\begin{equation}
\label{eq:p_norm}
\parallel \bar{\textbf{w}} \parallel_{2,p} = \parallel [\parallel \textbf{w}^{0} \parallel_{2},\parallel \textbf{w}^{1} \parallel_{2},...,\parallel \textbf{w}^{K} \parallel_{2}]\parallel_{p},
\end{equation}
\\that is the $p$-norm of a vector with $K+1$ elements, where each element is given by the 2-norm of the vector $\textbf{w}^{k}$ composed by $G$ elements. The dual norm of $\parallel \cdot \parallel_{p}$ is defined as $\parallel \cdot \parallel_{q}$ where $p$ and $q$ satisfy $\frac{1}{p} + \frac{1}{q} = 1$.\\
At this point we introduce the optimization problem in order to find the vector $\bar{\textbf{w}}$:

\begin{equation}
\label{eq:MinProb_MKAL}
\begin{aligned}
&\underset{\bar{\textbf{w}}}{\text{min}} \dfrac{\lambda}{2} \parallel \bar{\textbf{w}} \parallel^{2}_{2,p} + \dfrac{1}{N} \sum\limits_{i=1}^N \xi_{i} \\
&\text{subject to:}\quad \bar{\textbf{w}} \cdot (\bar{\phi} (\textbf{x}_{i}, y_{i})- \bar{\phi} (\textbf{x}_{i}, y)) \geq 1 - \xi_{i},\quad \forall \textit{i},\quad y \neq y_{i} ,
\end{aligned}
\end{equation}
\\where $\lambda = \frac{1}{C N}$ is a regularization term and $y$ is the predicted class of the vector $\textbf{x}_{i}$. The previous equation can be written as:

\begin{equation}
\label{eq:MinProbLoss_MKAL}
\underset{\bar{\textbf{w}}}{\text{min}} \dfrac{\lambda}{2} \parallel \bar{\textbf{w}} \parallel^{2}_{2,p} + \dfrac{1}{N} \sum\limits_{i=1}^N l(\bar{\textbf{w}},\textbf{x}_{i},y_{i}).
\end{equation}
\\The function that appears in the second term is the loss function for a multiclass problem. It is defined as:

\begin{equation}
\label{eq:lossMCfunction}
l(\textbf{w},\textbf{x}_{i},y_{i}) = \underset{y \neq y_{i}}{\text{max}} \lbrace 0,1-\bar{\textbf{w}} \cdot (\bar{\phi} (\textbf{x}_{i},y_{i}) - \bar{\phi} (\textbf{x}_{i},y) ). \rbrace
\end{equation}
\\The element $p$ that appears in the first term of equation (\ref{eq:MinProbLoss_MKAL}) can vary in the range $[1,2]$. By changing the value of $p$, the level of sparsity of the solution changes. With $p = 1$ we have only the sum of all the 2-norm of each hyperplane $ \textbf{w}^{k} $ of each source. Thus a solution with few hyperplanes is favoured. The $l_{1}$ norm introduces sparsity in the solution by reducing the complexity, but often it produces a non-convex problem. To select $p = 2$ is equivalent to choose an unweighted sum of kernels. To overcame these problems usually a $1 < p < 2$ is chosen.\\
Summarizing, the task of finding the vector $\bar{\textbf{w}}$ that minimize equation (\ref{eq:MinProbLoss_MKAL}) is equivalent to search for the best kernels combination that minimizes the loss function for a given set of training vectors.\\
There are several algorithms that solve computationally the minimal problem (\ref{eq:MinProbLoss_MKAL}). The one used in our implementation is called \textit{Obscure} (\cite{Orabona_MKL_1}, \cite{Orabona_MKL_2}).\\
The described process is iterated for $T$ times for all the training vectors. At each step the algorithm takes a random sample of the training set. The output of the training vector is evaluated using equation (\ref{eq:pred_MKAL}). At the first step all the parameters are null, in the others the parameters of previous iteration are taken.\\ 
For each training vector the loss function is evaluated with equation (\ref{eq:lossMCfunction}). The minimization problem that is tackled if the loss function is greater than zero is slightly different to the one earlier presented. It is formulated as follows:

\begin{equation}
\label{eq:MinProb_Obsc}
\bar{\textbf{w}}_{t+1} = \underset{\bar{\textbf{w}}}{\text{argmin}} [h(\bar{\textbf{w}}) + \bar{\textbf{w}} \sum\limits_{i=1}^{t} \eta_{i} \partial l (\bar{\textbf{w}}_{i},\textbf{x}_{i},y_{i}) ].
\end{equation}
\\The loss function is replaced with its sub-gradient, in order to linearise the problem. The first term of the equation is the regularization term\\ $h(\textbf{x})=\frac{\lambda}{2} \parallel \bar{\textbf{w}} \parallel^{2}_{2,p}$. In the second term the $\eta_{i}$ are a set of trade-off parameters that balance the speed of convergence and the precision.\\
The solution of above equation is:

\begin{equation}
\label{eq:Wt+1}
\bar{\textbf{w}}_{t+1} = \triangledown h^{*} \left(-\sum\limits_{i=1}^{t} \eta_{i} \partial l (\bar{\textbf{w}}_{i},\textbf{x}_{i},y_{i}) \right),
\end{equation}
\\where $h^{*}$ is the Fenchel conjugate of $h$. In the considered case the earlier equation is written as follows:

\begin{equation}
\label{eq:Wt+1_sostituita}
\textbf{w}^{k}_{t+1}= \dfrac{1}{q} \left( \dfrac{\parallel \boldsymbol{\theta}^{k}_{t+1} \parallel_{2}}{\parallel \bar{\boldsymbol{\theta}}_{t+1} \parallel_{2,q}} \right)^{q-2} \boldsymbol{\theta}^{k}_{t+1}, \quad \forall \textit{k} 
\end{equation}
\\where $\bar{\boldsymbol{\theta}} = (-\sum\limits_{i=1}^{t} \eta_{i} \partial l (\bar{\textbf{w}}_{i},\textbf{x}_{i},y_{i}) )$ is the dual weight of $\bar{\textbf{w}}$ and $q$ is the dual coefficient of $p$. If we have a positive loss function the vector $\bar{\textbf{w}}$ is updated, if the loss is null the vector doesn't change.\\
In order to exploit the kernel definition in final prediction (\ref{eq:pred_MKAL}) we can express as follow the vector $\bar{\textbf{w}}$: 

\begin{equation}
\label{eq:KernelInMKL}
\textbf{w}_{t+1}^{k} \propto \boldsymbol{\theta}_{t+1}^{k} = -\sum\limits_{i=1}^{t} \eta_{i} ((\phi^{k} (\textbf{x}_{i},y_{i}) - \phi^{k} (\textbf{x}_{i},y) )).
\end{equation}
\\The flow char of the code that implements \textit{Multi Kernel Adaptive Learning} method is shown in Figure \ref{fig:MKAL}. For each subject, during the training phase, is solved the optimization problem (\ref{eq:MinProbLoss_MKAL}) in order to find the parameters of the new models. 

\subsection{High Level-Learning2Learn algorithm} \label{sec:H-L2LAlg}
\begin{figure} [H]
 \centering
 \includegraphics [scale = 0.45] {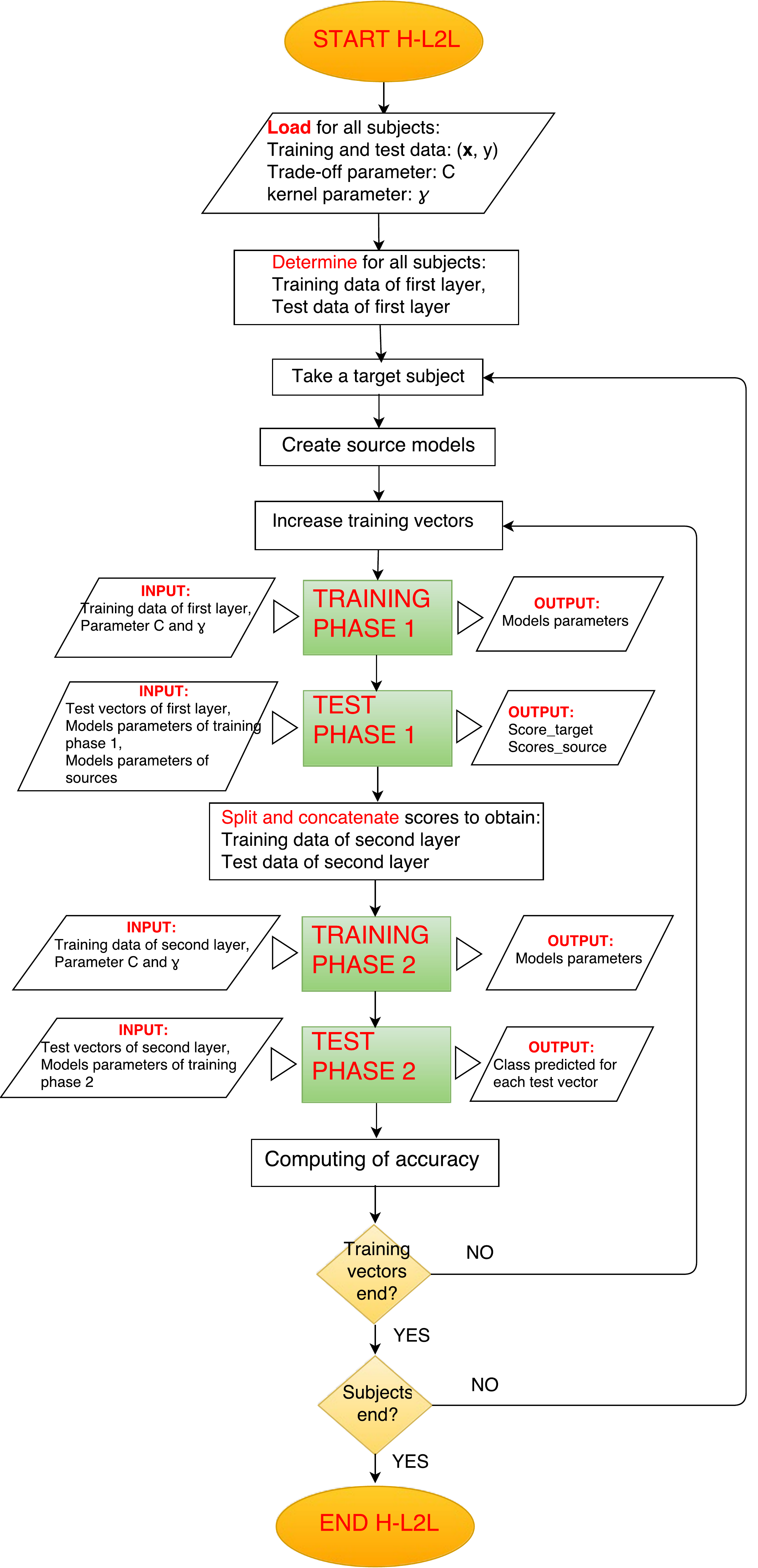}
 \caption{Flow chart of code of \textit{High Level-Learning2Learn algorithm}.}\label{fig:H-L2L}
\end{figure}

The algorithm \textit{High Level-Learning2Learn} (H-L2L) belongs to the high level cue integration methods (\cite{Tom_CueInt_Medical} and \cite{Novi_Cue}).\\ 
This algorithm is composed by two layers. In the first one, it is calculated the confidence score on each source model and on a target model built with a multiclass LS-SVM. The vectors used to train the target model are a part of the original training vectors ($63 \%$ for each class). The ones exploited in the calculation of confidence are all the test vectors and the remaining training vectors ($37 \%$ for each class).\\ 
Thus, given a vector $\textbf{x}$, for each output class $g$ we have the score $s^{target}(\textbf{x},g)$ and $K$ different scores $s^{source}(\textbf{x},g)$ in according to:

\begin{equation}
\label{eq:score_func_first_lay}
s(\textbf{x},g) = \textbf{w}_{g} \phi(\textbf{x}) + b_{g}.
\end{equation}
\\The vectors used to train the new model in the second layer are the concatenation of confidence scores of target and sources for the $37 \%$ of the original training vectors. These are exploited to solve a multiclass LS-SVM problem and the final score is given by:

\begin{equation}
\label{eq:score_fun_sec_layer}
s(\textbf{x},g) = \textbf{w}^{0} \phi(s^{target}(\textbf{x},g)) + \sum\limits_{k=1}^{K} \textbf{w}^{k} \phi(s^{sources}(\textbf{x},g)),
\end{equation}
\\where the the index $0$ refers to the target model. 
The concatenation of confidence scores of target and sources of the original test vectors are used to test the model of second layer.\\
The flow chart of the code that implements \textit{High Level-Learning2Learn algorithm} method is shown in Figure \ref{fig:H-L2L}. In the two training phases two different LS-SVM problems are solved, the first one exploiting original vectors and the second one using the output scores obtained previously.

\clearpage{\pagestyle{empty}\cleardoublepage}
\chapter{Results} \label{cha:Results}

In this chapter we evaluate and compare the performance and the results obtained using the state of the art algorithms described in chapter \ref{cha:Algorith} for data processed as explained in chapter \ref{cha:Data}.\\ 
Our goal is to understand how and if, for a new target subject, prior knowledge coming from other subjects can boost the control of prosthesis and reduce the time needed to learn how to control the device. Furthermore, when we consider a new amputee that is learning how to control the prosthesis, it is interesting to investigate which sources he/she exploits. In particular our task is to understand how the learning curve for an user might change when using intact or amputees as sources.\\

We present in Section \ref{sec:ExpProt} the experimental protocol used for each experiment.\\
Sections \ref{sec:II}, \ref{sec:AA} and \ref{sec:AI} focus on results obtained in the three different experiments done. In the first we consider an intact subject that exploits prior knowledge from other intact subjects. In the second we have an amputated subject that uses prior knowledge from other amputated subjects. In the third we consider an amputated subject that exploits prior knowledge from intact subjects.\\
We conclude with an overall discussion of our findings in Section \ref{sec:CompariosonRes}.

\section{Experimental protocol} \label{sec:ExpProt}
As said above, in this work we ran three different experiments. In this section we describe the experimental settings that are common to all of them.\\ 
We consider a target subject that exploits prior knowledge of source subjects in order to learn how to perform hand postures. Each experiment is repeated for several targets: 20 in the first and 9 in the second and third. The considered source models are 19, 8 and 20 in, respectively, the first, second and third set-up. Each protocol will be explained in detail in the following sections.\\
The goal that a target subjects aims to achieve is to learn to perform the same movements that source subjects are already able to do. For all the experiments we consider the postures of Exercise B, i.e. 8 hand configurations and 9 basic movements of the wrist (see section \ref{sec:Ninapro_dataset}). Thus, we are solving for each target a classification problem with 18 classes: 17 movements and rest posture.\\
Each experiment for each target subject has two phases: training and test. During the first, starting from few training vectors and prior models, each new target builds a classification model. During the second the model is tested with new vectors unused during the training phase.\\

The classification models of sources are built with a non-linear SVM with Gaussian kernel using the library \textit{LIBSVM} available online (\url{https://www.csie.ntu.edu.tw/~cjlin/libsvm/}).\\
In all settings the target model is built with an increasing number of training vectors, up to a maximum of $2160$. For each subject the experiment is repeated $18$ times where at each step we increase the training vectors of $120$.\\
The algorithms used in the training phase of each experiment are the same and represent the state of the art in adaptive learning. These are described in section \ref{sec:DA} and are: \textit{Multi Kernel Adaptive Learning} (MKAL), \textit{High Level-Learning2Learn} (H-L2L) and \textit{Multi Adapt}. We consider as reference two baseline: \textit{No Transfer} and \textit{Prior Features} (see section \ref{sec:DA}).\\
In the test phase, the final performance is evaluated using the formula of balanced accuracy that takes into account a possible imbalance in the number of vectors in different classes. We report the formulation for a binary problem:

\begin{equation}
\label{eq:BalAcc}
\dfrac{0.5 TP}{TP + FN} + \dfrac{0.5 TN}{TN+FP}.
\end{equation}
\\We refer to the positive or negative samples classified correctly with TP (True Positive) and TN (True Negative). With FP (False Positive) and FN (False Negative) we indicate the negative or positive samples misclassified. This formulation can be straightforwardly extended to multiclass case. An average performance is calculated for each algorithm, it comes from the mean between the performance obtained for all the target subjects. For each subject the performance is evaluated for each considered set of training vectors. Thus, we evaluate the trend of mean performance as the number of training vectors increases.\\

All the classification models used are based on non-linear SVM with Gaussian kernel (see equation \ref{eq:GausKer}). These need the setting of the hyperparameters $C$ and $\gamma$. The method generally used to determine them is called cross-validation. A grid with the following values for parameters is taken into account: $C = (0.01,0.1,1,10,100,1000)$ and $\gamma = (0.01,0.1,1,10,100,1000)$. For each subject and for each possible combination of $C$ and $\gamma$, a classification model is built using the library \textit{LIBSVM}. The couple of parameters that gives the best result in performance is chosen.\\

All the obtained results for each experiment are available in \url{https://sites.google.com/site/noninvasiveprosthetichand/}.\\

\section{Intact-Intact} \label{sec:II}

\paragraph{Setup.} The first experiment that we tackle is close to the experiments already performed in the literature (\cite{Novi_Nina1} and \cite{Tom_hand_pros}). It involves 20 intact random subjects from the second sub-database of NinaPro (see Table in section \ref{sec:Ninapro_dataset}). In this case we consider intact subjects as target and source. In particular, one by one, each of the 20 subjects is the new target problem. The remaining 19 intact subjects are considered as sources.\\
The data taken into account are those of the Exercise B: 8 hand configurations and 9 basic movements of the wrist.\\
During the training phase the classification models are built for each target subject and for each of the five algorithms. We repeat the process for an increasing number of training vectors with steps of $120$ up to a maximum of $2160$. In the test phase we evaluate the performance of each model.\\  
In this work we performed the same experiment for 10, 20 and 30 intact random subjects. Taking into account that the final performance slightly increases from 10 to 20 but doesn't change from 20 to 30, in this section we report only the results and analysis for 20 subjects. The interested readers can find all results in \url{https://sites.google.com/site/noninvasiveprosthetichand/}.\\ 
In the Tables \ref{tab:DatasetII} and \ref{tab:DatasetIIAlg} we report the principal characteristics of the first experiment.

\begin{table}[H]
	
	\resizebox{13,5cm}{!}	{
	
	\begin{tabular} {|| l | l | p{7cm} | l ||}
	
	\hline
	\hline
	  & \textbf{Database} & \textbf{Subjects} & \textbf{Postures} \\ 		\hline
	\hline
	\textbf{Target} & 2: Intact & 20 random: 1, 3, 4, 5, 9, 10, 12, 20, 21, 22, 24, 25, 29, 30, 31, 34, 35, 38, 39, 40  & 17 + rest\\    \hline
	\textbf{Source} & 2: Intact & 20 random: 1, 3, 4, 5, 9, 10, 12, 20, 21, 22, 24, 25, 29, 30, 31, 34, 35, 38, 39, 40 & 17 + rest\\   \hline
	\hline
	\end{tabular} \\
	}
	\caption{Characteristics of used data.}
	\label{tab:DatasetII}
\end{table}

\begin{table}[H]
	\quad
	
	\resizebox{14cm}{!}	{
	
	\begin{tabular}{|| l | p{6,5cm} | p{4cm} ||}
	
	\hline
	  & \textbf{Adaptive} & \textbf{Baseline}  \\ 		\hline
	\hline	
	\textbf{Algorithms} & Multi Kernel Adaptive Learning, High Level-Learning2Learn, Multi Adapt & No Transfer, Prior Features \\ \hline 
	\hline

	\end{tabular}

	}
	
	\caption{Used algorithms.}
	
\label{tab:DatasetIIAlg}
\end{table}
 
\begin{figure} [H]
\centering
\includegraphics [scale = 0.4] {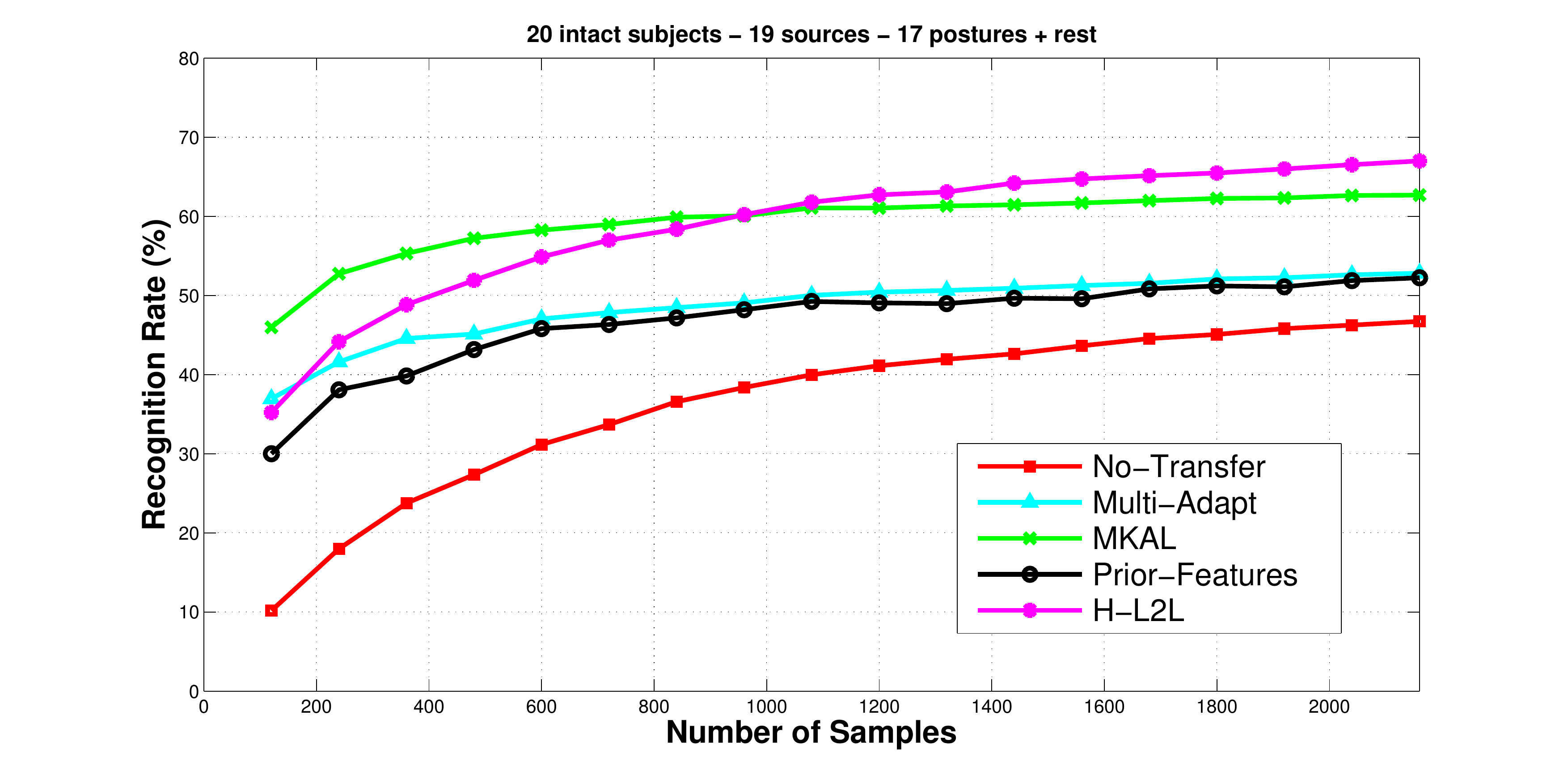}
\caption{Classification rate obtained by averaging over all the subjects as a function of the number of samples in the training set. }\label{fig:20IntactSubj}
\end{figure}

\paragraph{Results: Recognition Rate.} For all the algorithms, the trend of the mean performance as a function of the training vectors of the target problem is reported in Figure \ref{fig:20IntactSubj}. It comes from the mean between performance obtained for all the target subjects.

\begin{figure} [H]
\centering
\subfigure
   {\includegraphics[scale=0.38]{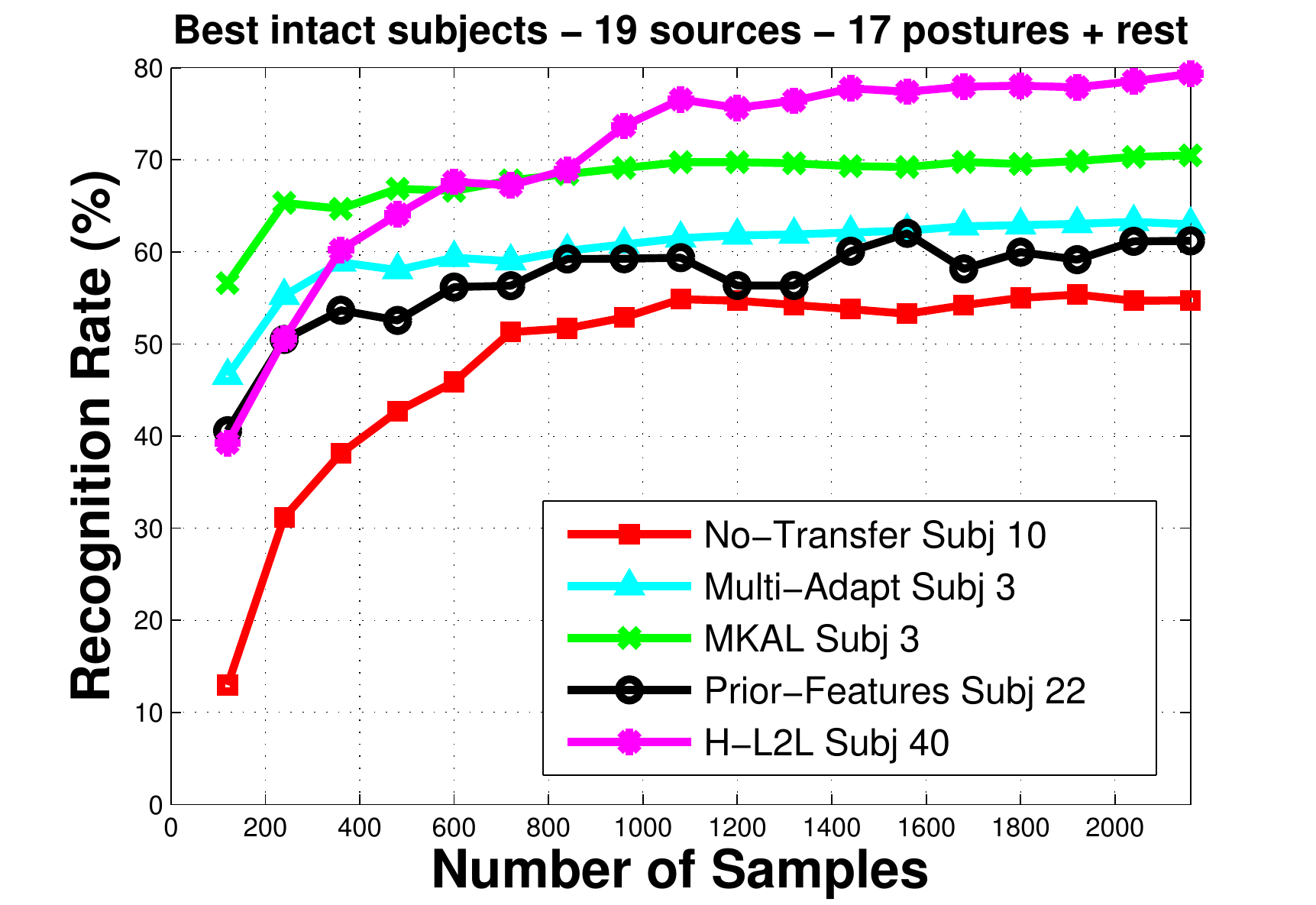}}
   \enspace
 \subfigure
   {\includegraphics[scale=0.38]{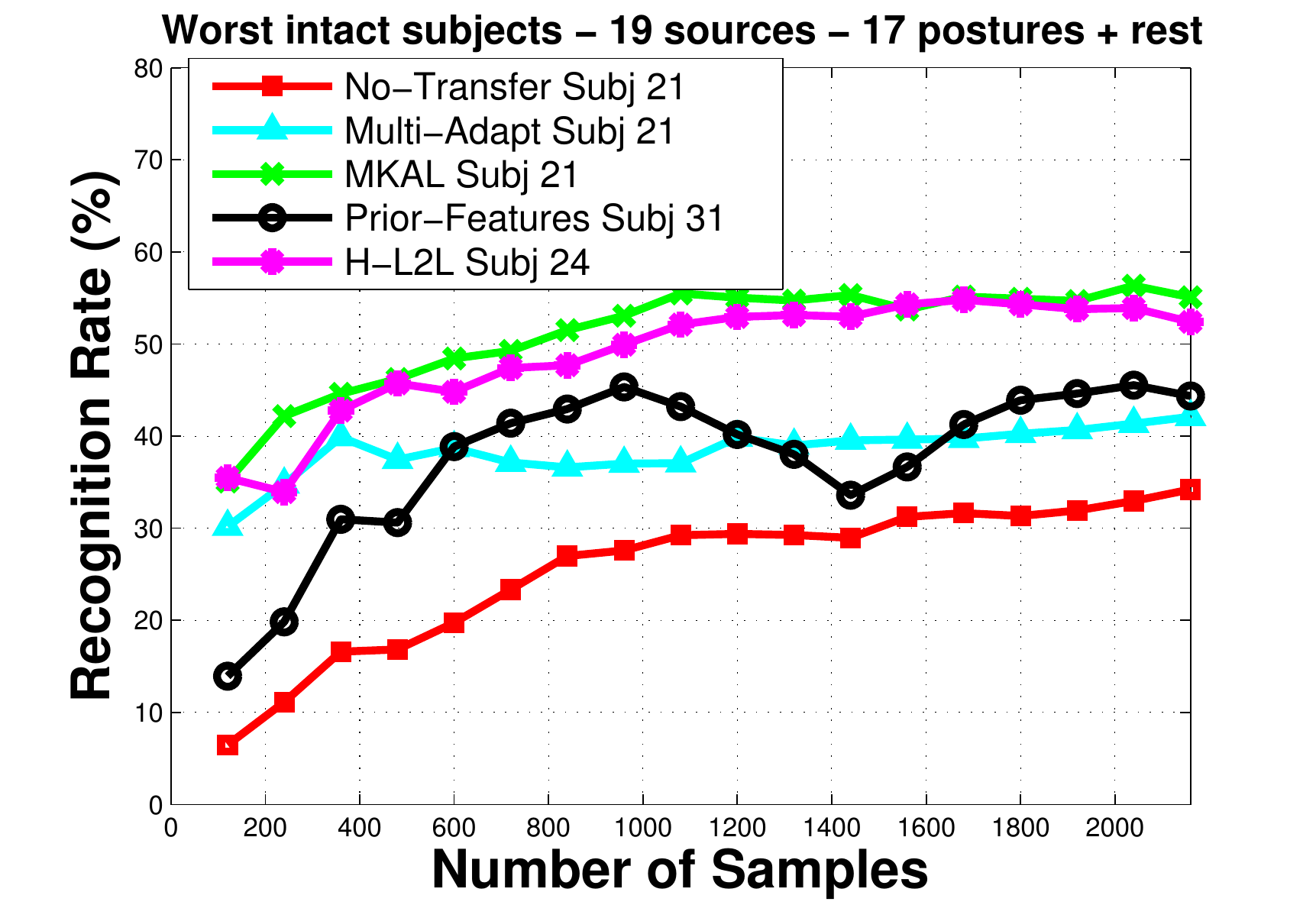}}
  \caption{Classification rate for the best and worst subjects as a function of the number of samples in the training set.}\label{fig:20IntactSubjBW} 
\end{figure}

In Figure \ref{fig:20IntactSubjBW} the best and worst cases are reported. These are respectively the subjects for which each algorithm gives the best and worst result in performance. 

MKAL and H-L2L achieve the best performance followed by Multi Adapt, Prior Features and No Transfer. The same order is preserved in the best and worst case, but a component of noise is present for the single subject and the trend appears not very smooth.\\
As shown in Figure \ref{fig:20IntactSubj} H-L2L outperforms MKAL for more than 1000 training
samples of about 3  $\%$ ($p < 0.05 $). 
The difference between MKAL and Multi Adapt shows an average advantage in recognition rate of around 10 $\%$ ($p < 0.05 $). 
Multi Adapt has an average gain of 2 $\%$ with respect to Prior Features ($p < 0.05 $).\\ 
No Transfer is the method that shows the lowest performance. Multi Adapt, MKAL and H-L2L outperform No Transfer with an average of about 12 $\%$, 23 $\%$ and 22 $\%$ respectively. At the last step, i.e. 2160 training vectors, No Transfer achieves a performance in recognition of about 47 $\%$. The adaptive methods reach before this goal: 57 $\%$ for MKAL at 240 training samples, 49 $\%$ for H-L2L at 360 and 47 $\%$ for Multi Adapt at 600. This result means that the use of prior knowledge allows us to reduce by one order of magnitude the training time.\\
The adaptive methods achieve faster than No Transfer the asymptotic performance. In fact, passing from 600 training samples to 2160 the performance of No Transfer, Multi Adapt, MKAL and H-L2L increases of respectively: 16 $\%$, 5 $\%$, 5 $\%$ and 12 $\%$.\\

\begin{figure} [H]
\centering
\subfigure
   {\includegraphics[scale=0.38]{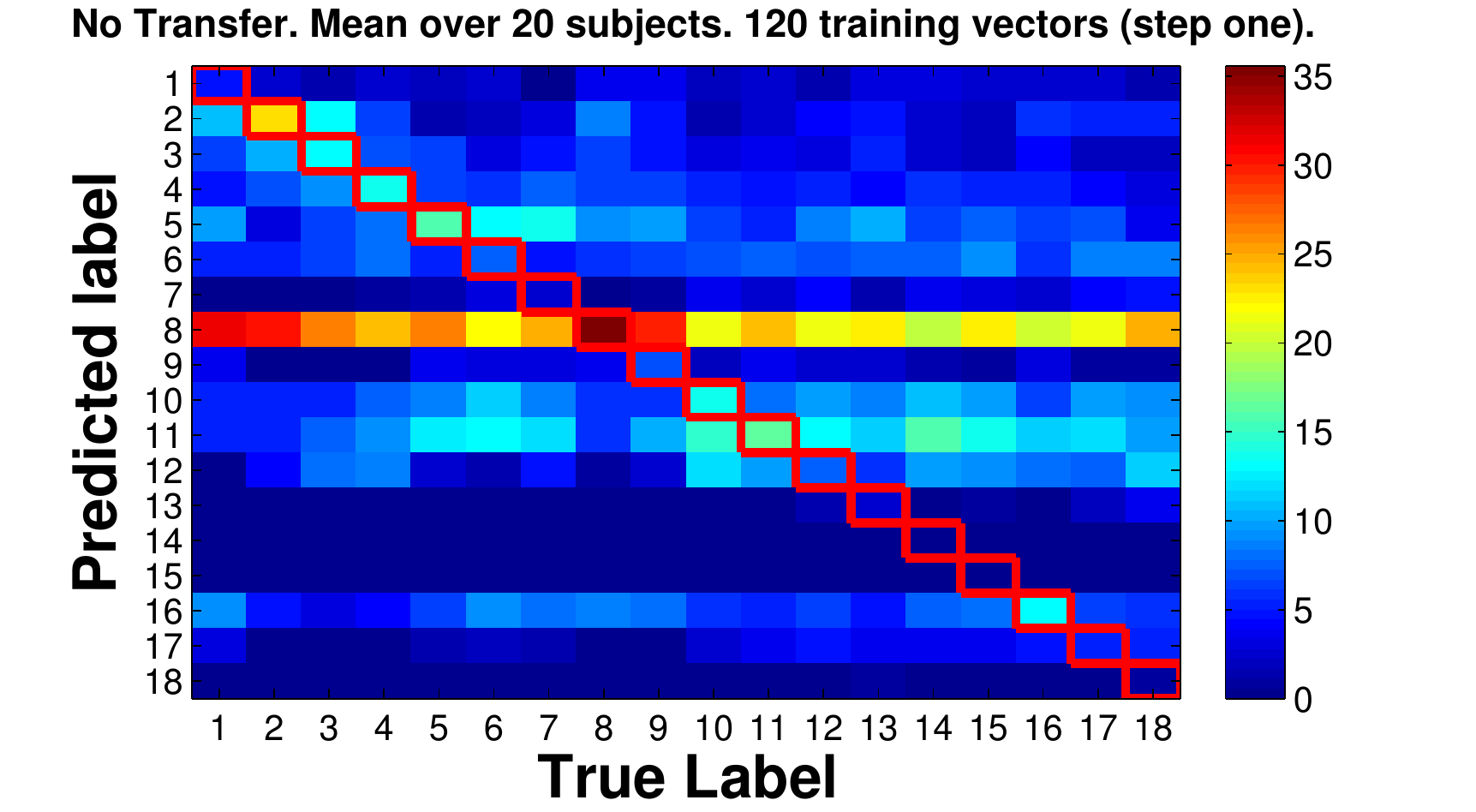}}
 \enspace
 \subfigure
   {\includegraphics[scale=0.38]{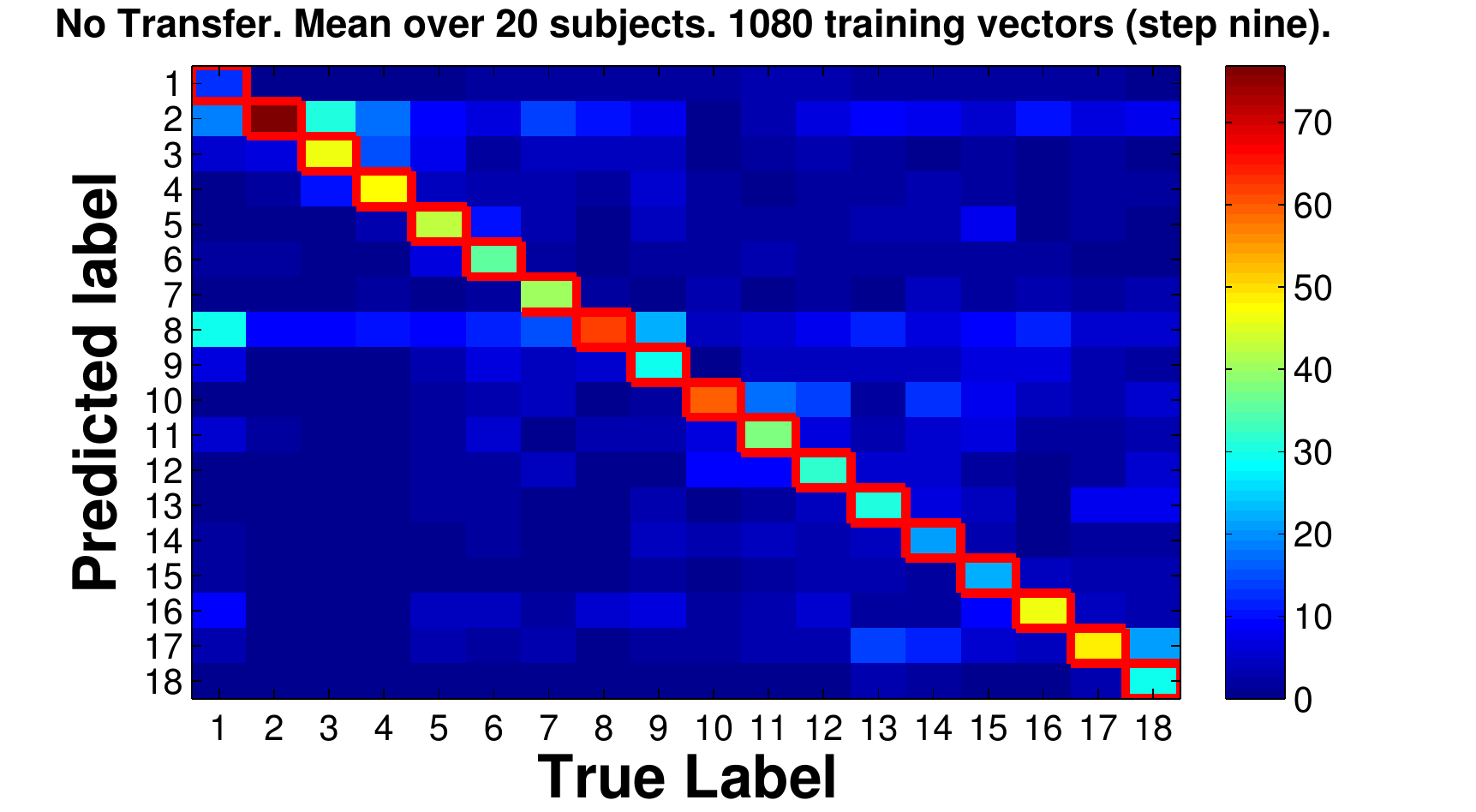}}
   \enspace
 \subfigure
   {\includegraphics[scale=0.38]{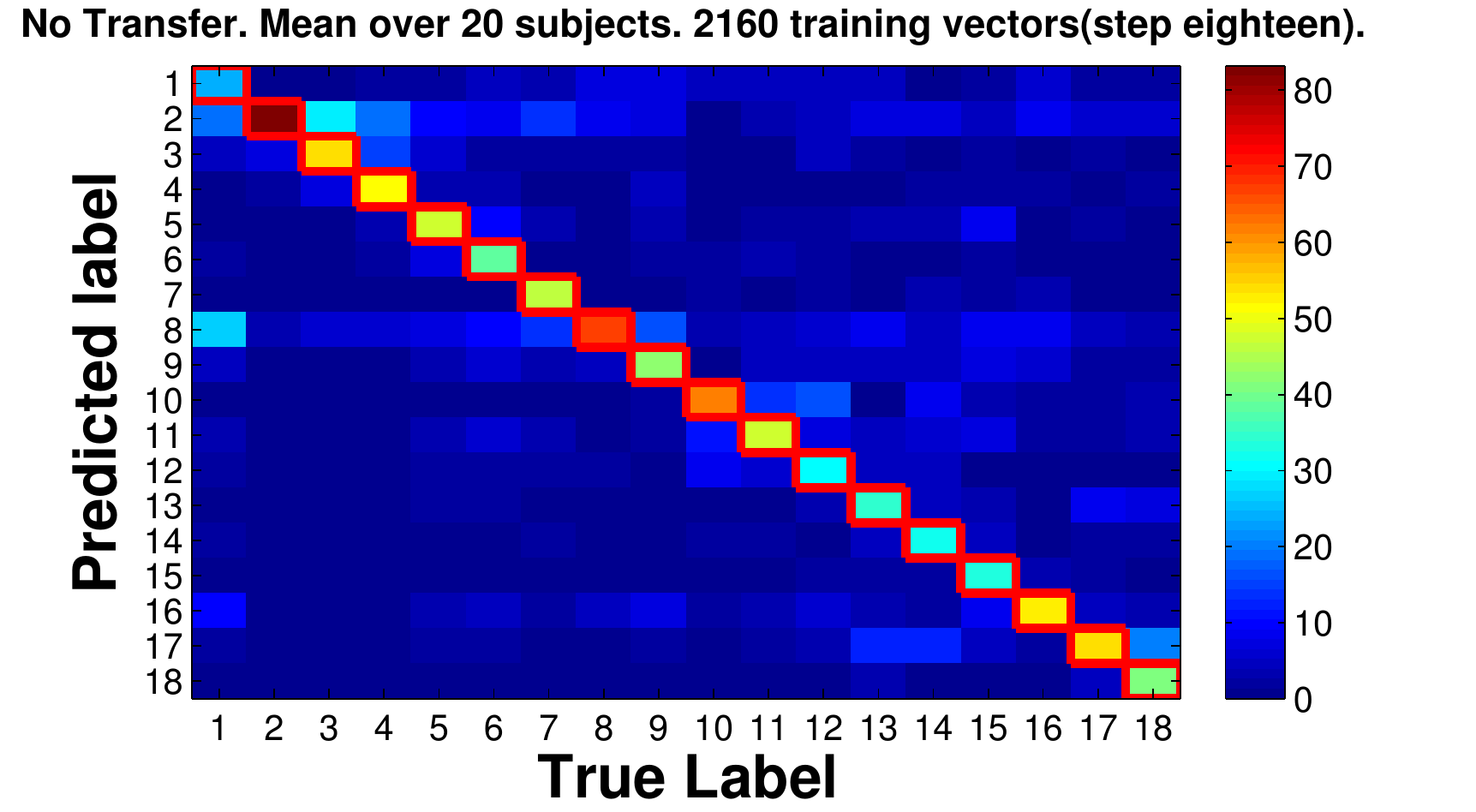}}
  \caption{Confusion matrices for No Transfer method 120, 1080 and 2160 training vectors.}\label{fig:NTConfMat} 
\end{figure}

\begin{figure} [H]
\centering
\subfigure
   {\includegraphics[scale=0.38]{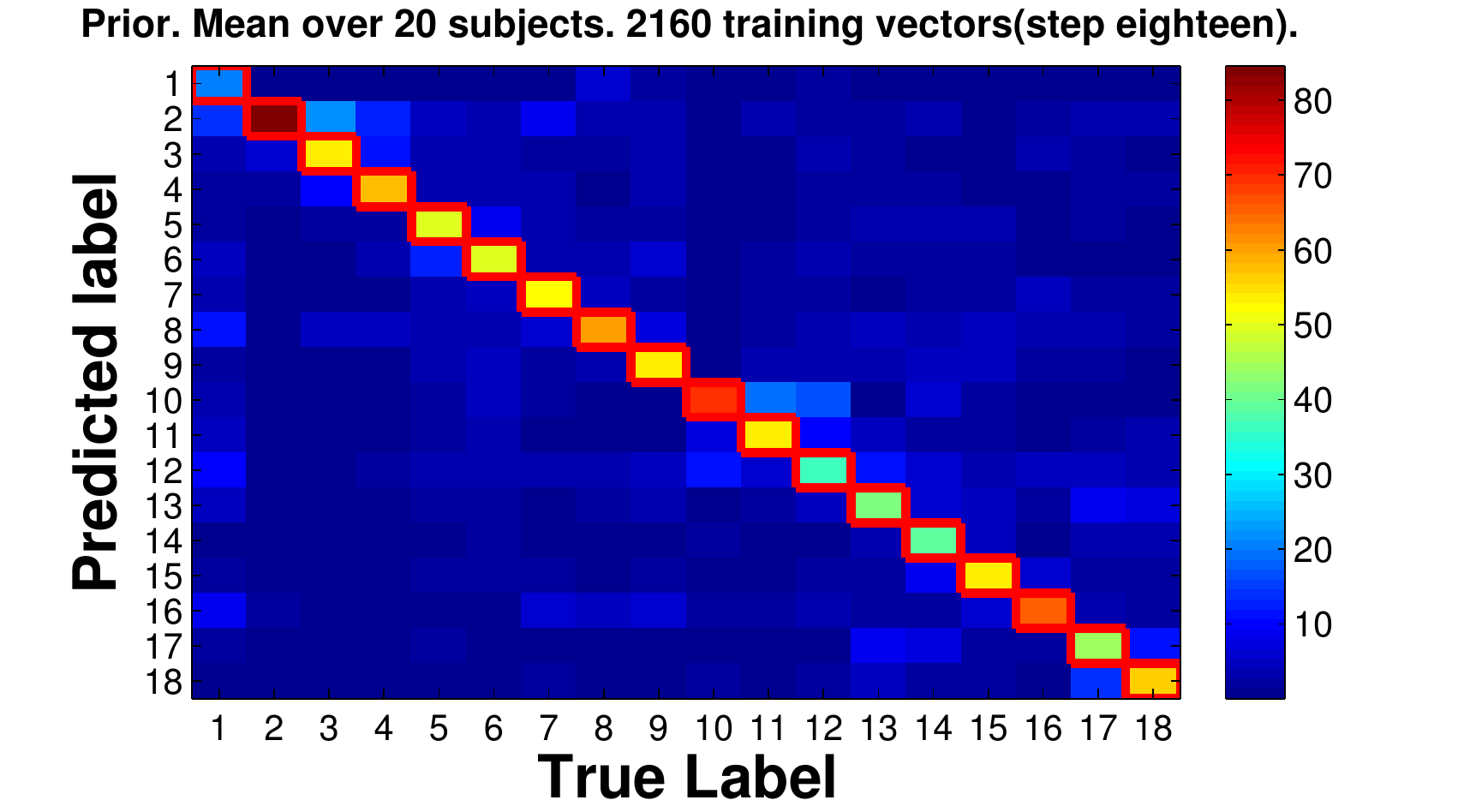}}
 \enspace
 \subfigure
   {\includegraphics[scale=0.38]{Prior_2160TrainPIC.pdf}}
   \enspace
 \subfigure
   {\includegraphics[scale=0.38]{Prior_2160TrainPIC.pdf}}
  \caption{Confusion matrices for Prior method 120, 1080 and 2160 training vectors.}\label{fig:PriorConfMat} 
\end{figure}

\begin{figure} [H]
\centering
\subfigure
   {\includegraphics[scale=0.38]{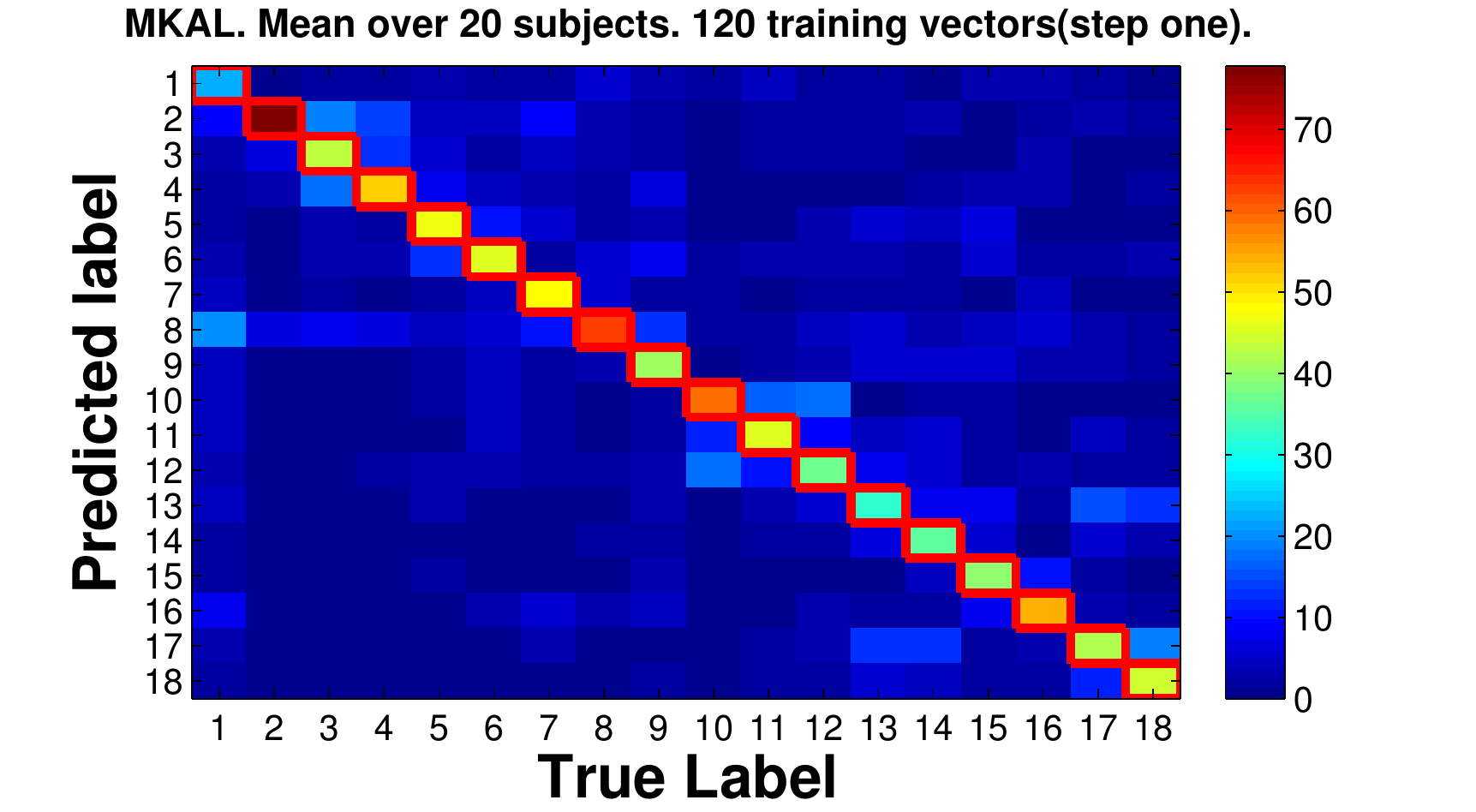}}
 \enspace
 \subfigure
   {\includegraphics[scale=0.38]{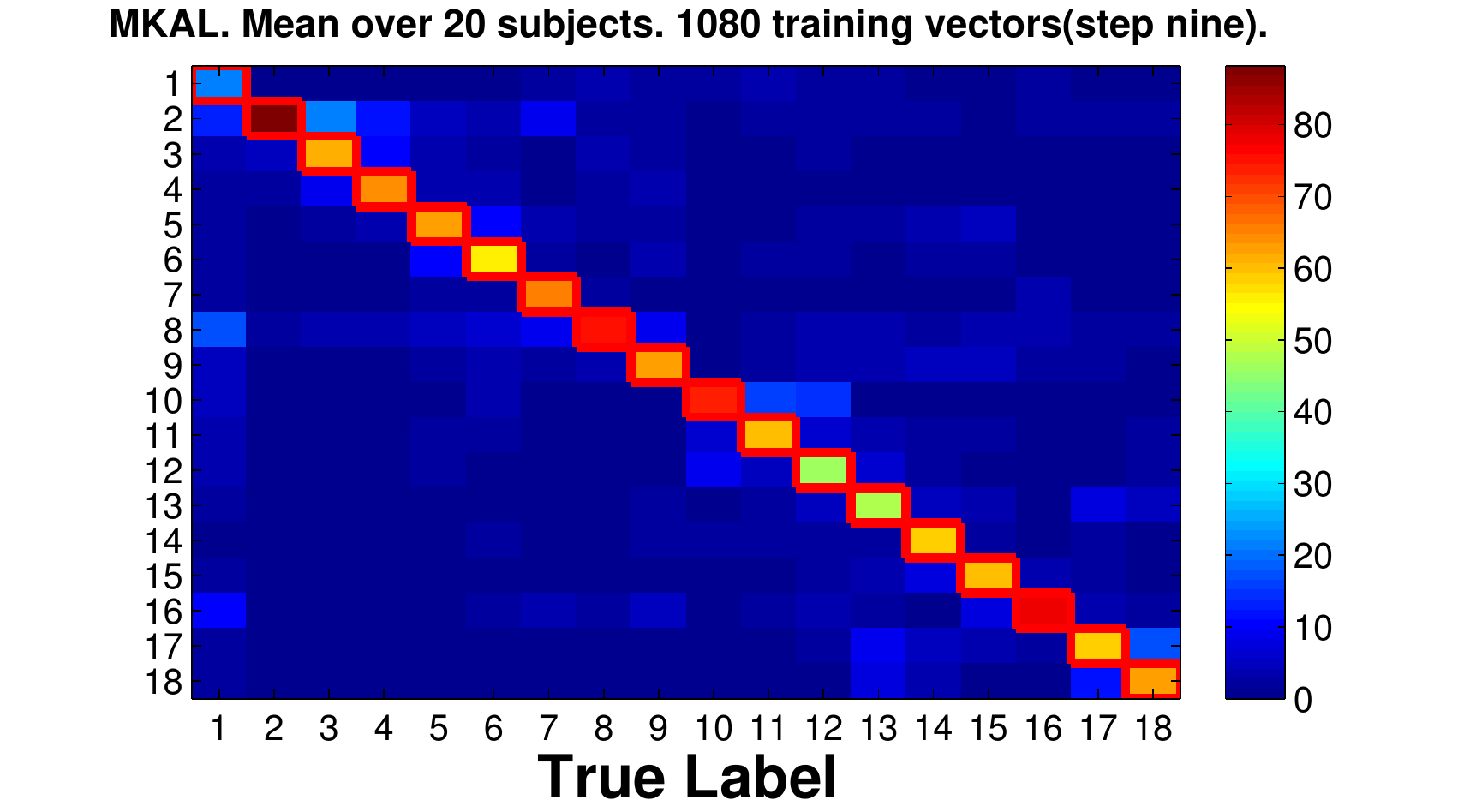}}
   \enspace
 \subfigure
   {\includegraphics[scale=0.38]{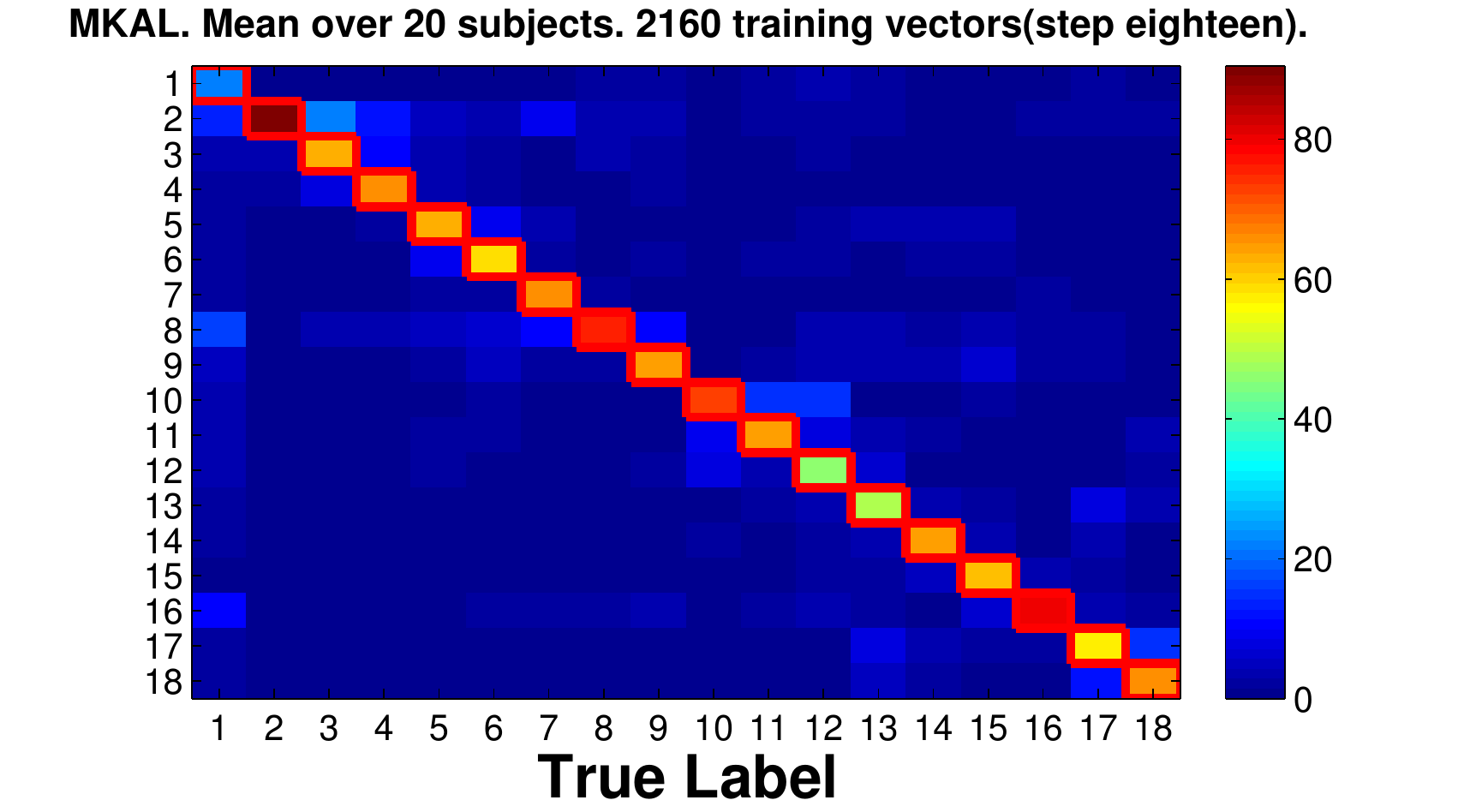}}
  \caption{Confusion matrices for MKAL method 120, 1080 and 2160 training vectors.}\label{fig:MKALConfMat} 
\end{figure}

\paragraph{Results: Confusion Matrices.} Let us introduce the confusion matrix. This matrix contains information about real labels and labels predicted by a classification model. Each row represents the predicted values associated to a class for each real label. Each column represents the prediction given for each real class, so the cumulative is 1.\\
The analysis of confusion matrices shows several aspects of recognition in a classification problem that the single performance analysis hides. We can check if a set of classes are better recognize than others or if there are differences in recognition of a single posture changing algorithm. Thus we obtain a statistic of recognition about single class, changing algorithm and number of training samples.\\
We report the confusion matrices for No Transfer (Figure \ref{fig:NTConfMat}), Prior Features (Figure \ref{fig:PriorConfMat}) and MKAL (Figure \ref{fig:MKALConfMat}) for 120 (i.e. initial step), 1080 (i.e. middle step) and 2160 (i.e. final step) training vectors. The label 1 is associated to rest posture.\\
We can find the confusion matrices for all algorithms in \url{https://sites.google.com/site/noninvasiveprosthetichand/}.\\

As we can evaluate from colorbar, the warm colors are associated to high probability and cool colors to low probability. The ideal case is represented by red colour into the diagonal of matrix (i.e. the prediction labels are equal to the true ones) and blue colours outside (i.e. wrong prediction is equal to zero). In all three cases, augmenting the number of training vectors, the warm colours move towards the diagonal.\\
It is interesting to analyse the case of No Transfer. It doesn't exploit any type of source and predictions are based only on training vectors of the target used at each step. At first one the classes with higher predictions are 8, 10, 11 and 12. It means that the 120 training vectors of first step in majority belong to these classes.\\ 
The situation at first step is different for Prior Features and MKAL, it means that the recognition task is helped by the presence of the source knowledge. With No Transfer, at the first step, each class has as output the predicted value of 8. With Prior Features all the classes, except 1 and 14, have the highest prediction associated to the true label. With MKAL for each true label the highest prediction is always on the diagonal.\\
The increasing of training samples gives stability in all cases. At the last step there aren't attractors, i.e. there are no rows with higher cumulative probability than others. For all methods the highest predictions are inside the diagonal except one class for No Transfer.\\
In all cases we can note that the posture with lowest accuracy on the diagonal is the 1, i.e. the rest. We recall that, in data acquisition process, each movement is inter-spaced by rest. It makes this posture different than others and easily confused.\\

Another analysis can be done to evaluate if there is an adaptive method that recognizes a class better than others and if this statistic changes increasing the number of training vectors. It is based on the results obtained with confusion matrices. For each column (i.e. each real class) we consider the four classes with highest predictions.\\
The details of this analysis are reported in the Appendix \ref{sec:HistI}.\\
From this analysis we can conclude that, given a class, different methods, generally, misclassify it with the same wrong classes. The situation doesn't change as we increase the number of training vectors. Thus, different algorithms have different mean performances but, paying our attention to a single class, the type of misclassification are the same.\\

\section{Amputees-Amputees} \label{sec:AA}

\paragraph{Setup.} The second experiment involves 9 amputated subjects from the third sub-database of NinaPro (see Table in section \ref{sec:Ninapro_dataset}).\\
In this set-up we consider amputees as target and sources. In particular, one by one, each of the 9 subjects is the new target problem. The remaining 8 amputated subjects are considered as sources. Thus, we have an amputee that learns to use the prosthetic hand with the help of prior knowledge from other amputees.\\
The data taken into account are those of the Exercise B: 8 hand configurations and 9 basic movements of the wrist.\\
In the training phase a model, using each of the five algorithms, is built for each target subject. The process is repeated for an increasing number of training vectors with steps of $120$ up to a maximum of $2160$. In the test phase we evaluate the performance of each model.\\  
The 9 subjects of this experiment are not chosen randomly. As explained in section \ref{sec:Data_acq}, $2$ of the $11$ amputees had only $10$ electrodes instead of $12$ because of insufficient space in the stump. At the beginning we tested all the 11 subjects, but those with less electrodes had a lower performance than others. Thus, in the following we report only the results for 9 subjects; the results for 11 subjects are available in \url{https://sites.google.com/site/noninvasiveprosthetichand/}.\\
In Tables \ref{tab:DatasetAA} and \ref{tab:DatasetAAAlg} we report the principal characteristics of second experiment.\\

\begin{table}[H]
	
	\resizebox{13,5cm}{!}	{
	
	\begin{tabular} {|| l | l | p{7cm} | l ||}
	
	\hline
	\hline
	  & \textbf{Database} & \textbf{Subjects} & \textbf{Postures} \\ 		\hline
	\hline
	\textbf{Target} & 3: Amputees & 9 no-random: 1, 2, 3, 4, 5, 6, 9, 10, 11  & 17 + rest\\    \hline
	\textbf{Source} & 3: Amputees & 9 no-random: 1, 2, 3, 4, 5, 6, 9, 10, 11  & 17 + rest\\   \hline
	\hline
	\end{tabular} \\
	}
	\caption{Characteristics of used data.}
	\label{tab:DatasetAA}
\end{table}

\begin{table}[H]
	\quad
	
	\resizebox{14cm}{!}	{
	
	\begin{tabular}{|| l | p{6,5cm} | p{4cm} ||}
	
	\hline
	  & \textbf{Adaptive} & \textbf{Baseline}  \\ 		\hline
	\hline	
	\textbf{Algorithms} & Multi Kernel Adaptive Learning, High Level-Learning2Learn, Multi Adapt & No Transfer, Prior Features \\ \hline 
	\hline

	\end{tabular}

	}
	
	\caption{Used algorithms.}
	
\label{tab:DatasetAAAlg}
\end{table}

\begin{figure} [H]
\centering
\includegraphics [scale = 0.4] {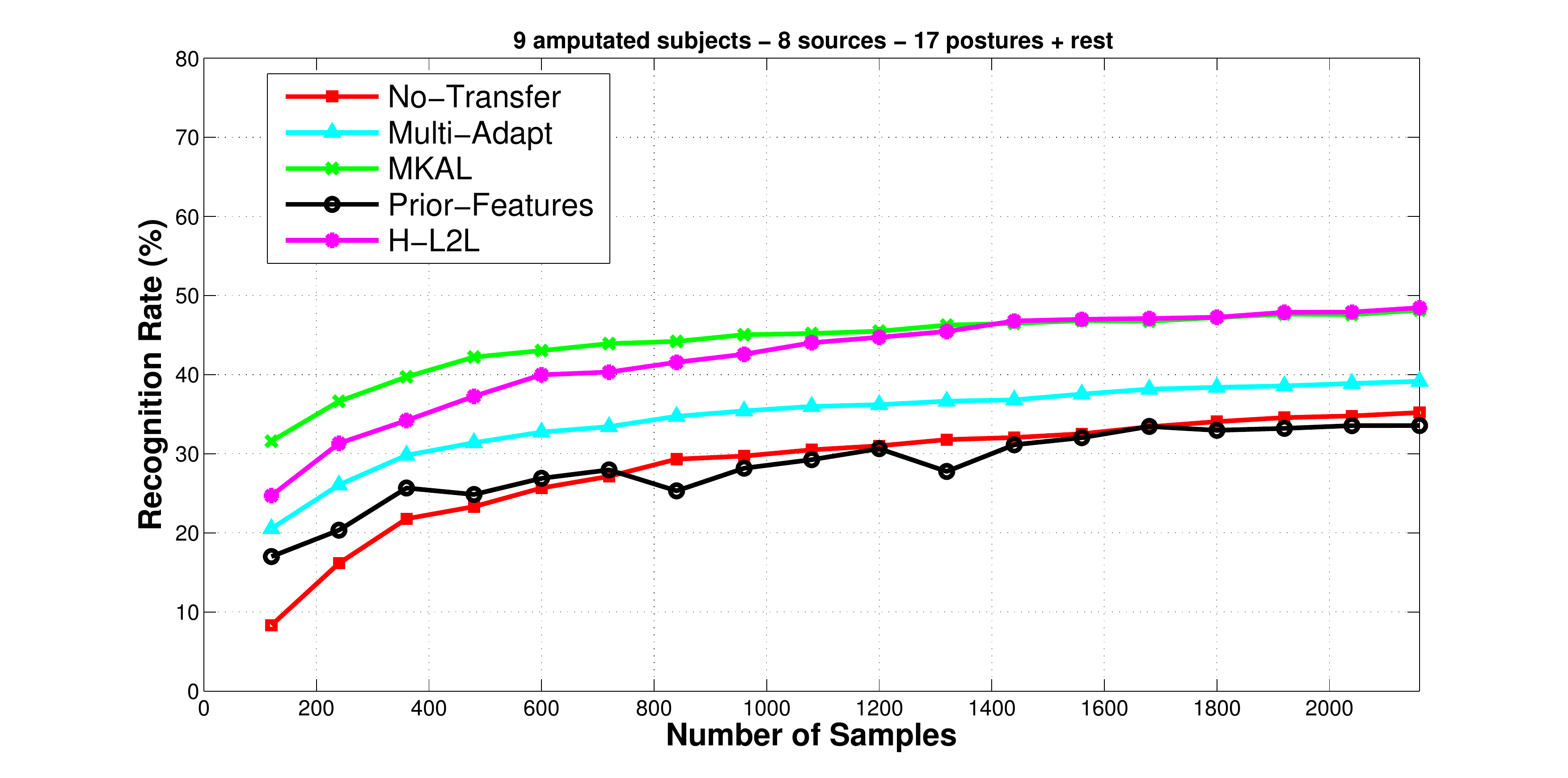}
\caption{Classification rate obtained by averaging over all the subjects as a function of the number of samples in the training set. }\label{fig:9AmputatedSubj}
\end{figure}

\paragraph{Results: Recognition Rate.} The trend of the performance averaged over all the subjects as a function of the training vectors of the target problem is reported in Figure \ref{fig:9AmputatedSubj} for all the algorithms. It comes from the mean between performance obtained for all the target subjects.\\

\begin{figure} [H]
\centering
\subfigure
   {\includegraphics[scale=0.38]{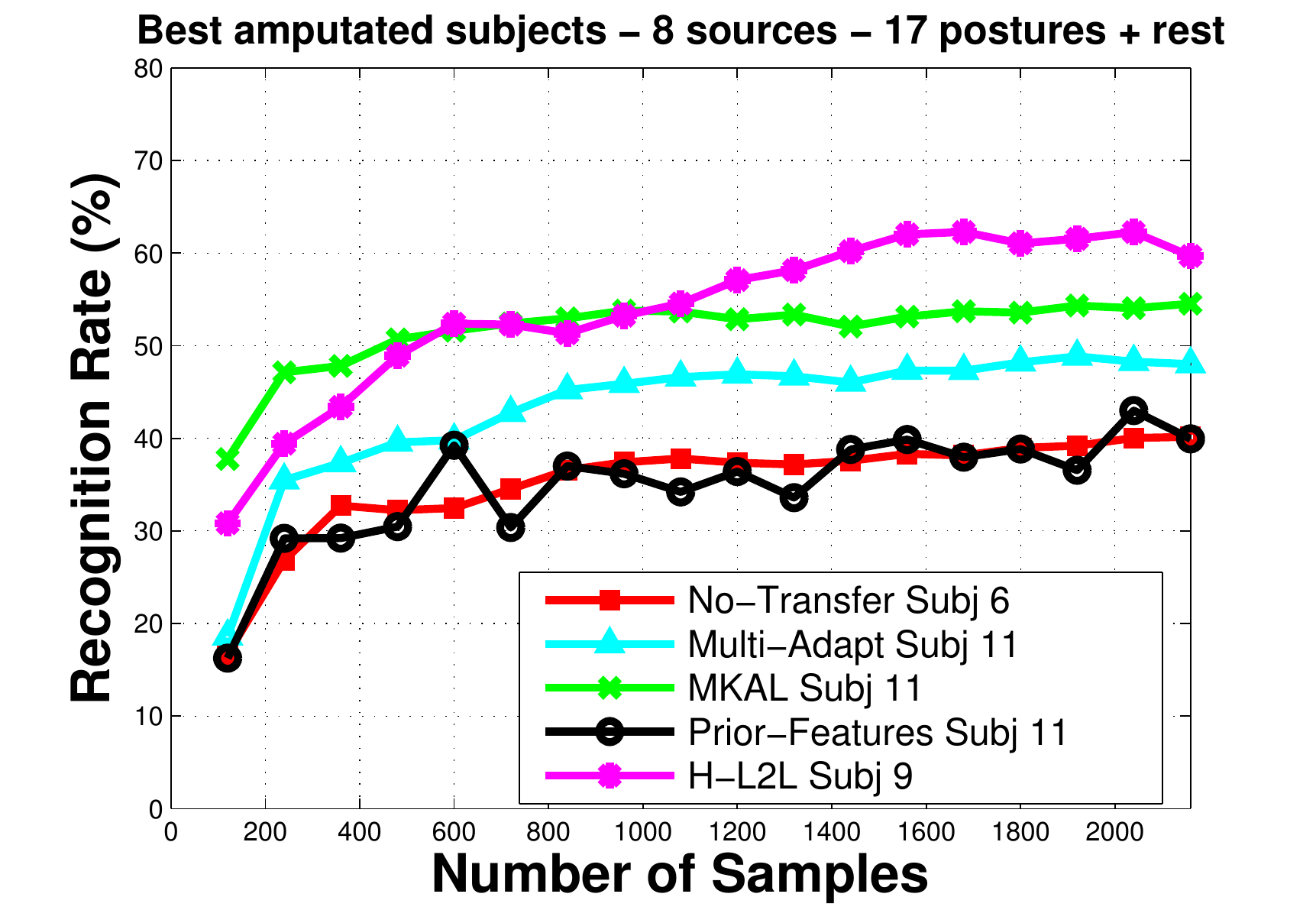}}
 \enspace
 \subfigure
   {\includegraphics[scale=0.38]{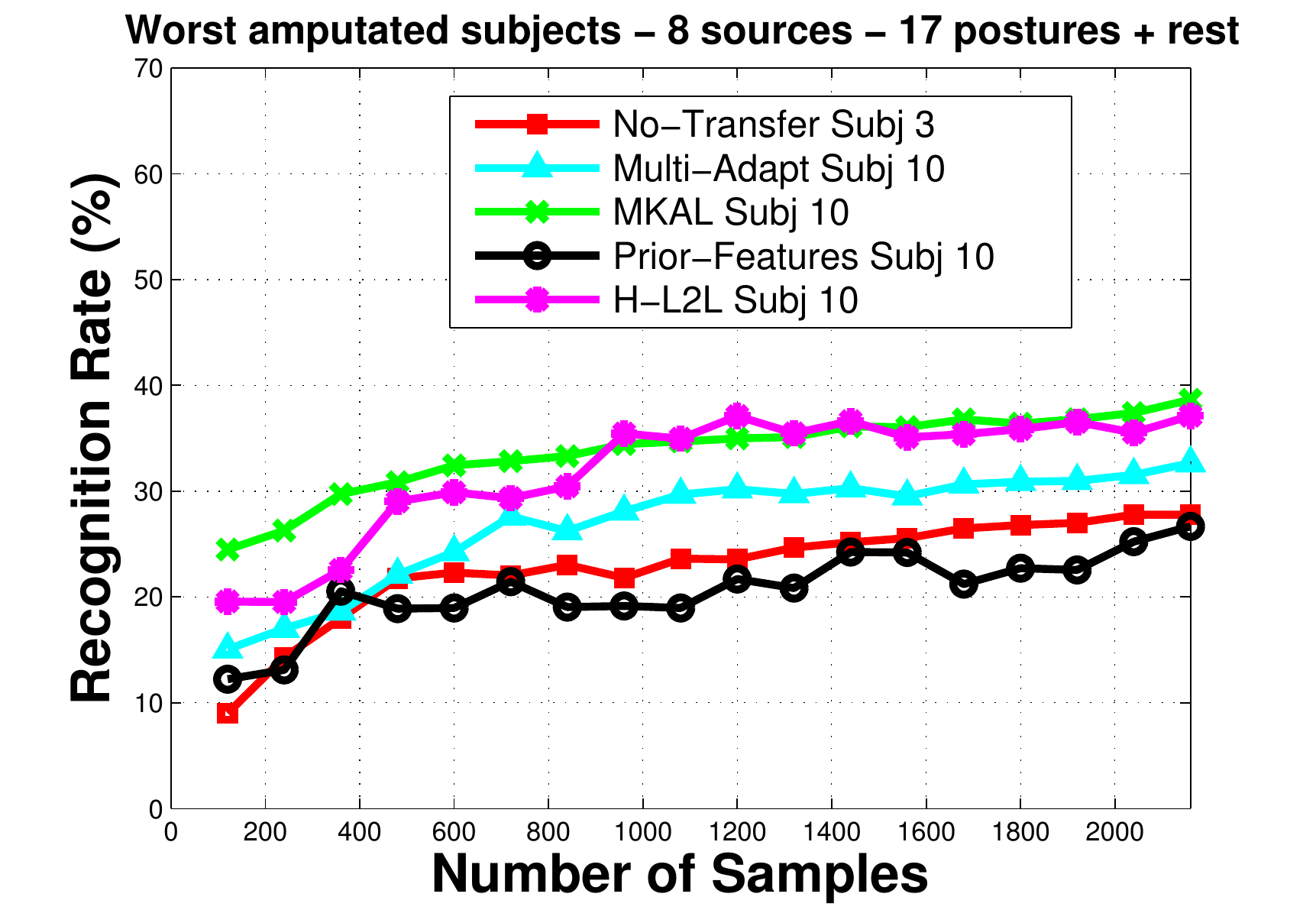}}
  \caption{Classification rate for the best and worst subjects as a function of the number of samples in the training set.}\label{fig:9AmputatedSubjBW} 
\end{figure}

In Figure \ref{fig:9AmputatedSubjBW} the best and worst cases are reported. These are respectively the subjects for which each algorithm gives the best and worst result in performance. 

Asymptotically MKAL and H-L2L achieve the best performance followed by Multi Adapt and Prior Features with No Transfer. A similar order is preserved in the best and worst case but a component of noise is present for single subject and the trend appears not very smooth. The only difference is that Prior Features outperforms No Transfer of about 2 $\%$ in the worst case ($p < 0.05$).\\ 
As shown in Figure \ref{fig:9AmputatedSubj} MKAL performs better than H-L2L with an average of 4 $\%$ until 1080 training vectors ($p < 0.05$), after these curves give the same result in performance. 
The difference between MKAL and Multi Adapt shows an average advantage in recognition rate of around 10 $\%$ for first curve ($p < 0.05 $). Multi Adapt has an average gain of 6 $\%$ with respect to Prior Features ($p < 0.05 $).\\  
No Transfer and Prior show the lowest performance and their difference is not statistically significant ($p > 0.05 $). Multi Adapt, MKAL and H-L2L outperform No Transfer with an average of about 6 $\%$, 16 $\%$ and 14 $\%$ respectively. At 2160 training samples, i.e. the last step, No Transfer achieves a performance of 35 $\%$. The same performance is reached by Multi Adapt, MKAL and H-L2L at only 840, 240 and 480 training vectors, respectively. This result shows that the use of prior knowledge allows us to reduce by one order of magnitude the training time.\\
The adaptive methods achieve faster than No Transfer the asymptotic performance. Passing from 600 training samples to 2160 the performance of No Transfer, Multi Adapt, MKAL and H-L2L increases of respectively: 9 $\%$, 6 $\%$, 5 $\%$ and 8 $\%$.\\

\begin{figure} [H]
\centering
\subfigure
   {\includegraphics[scale=0.38]{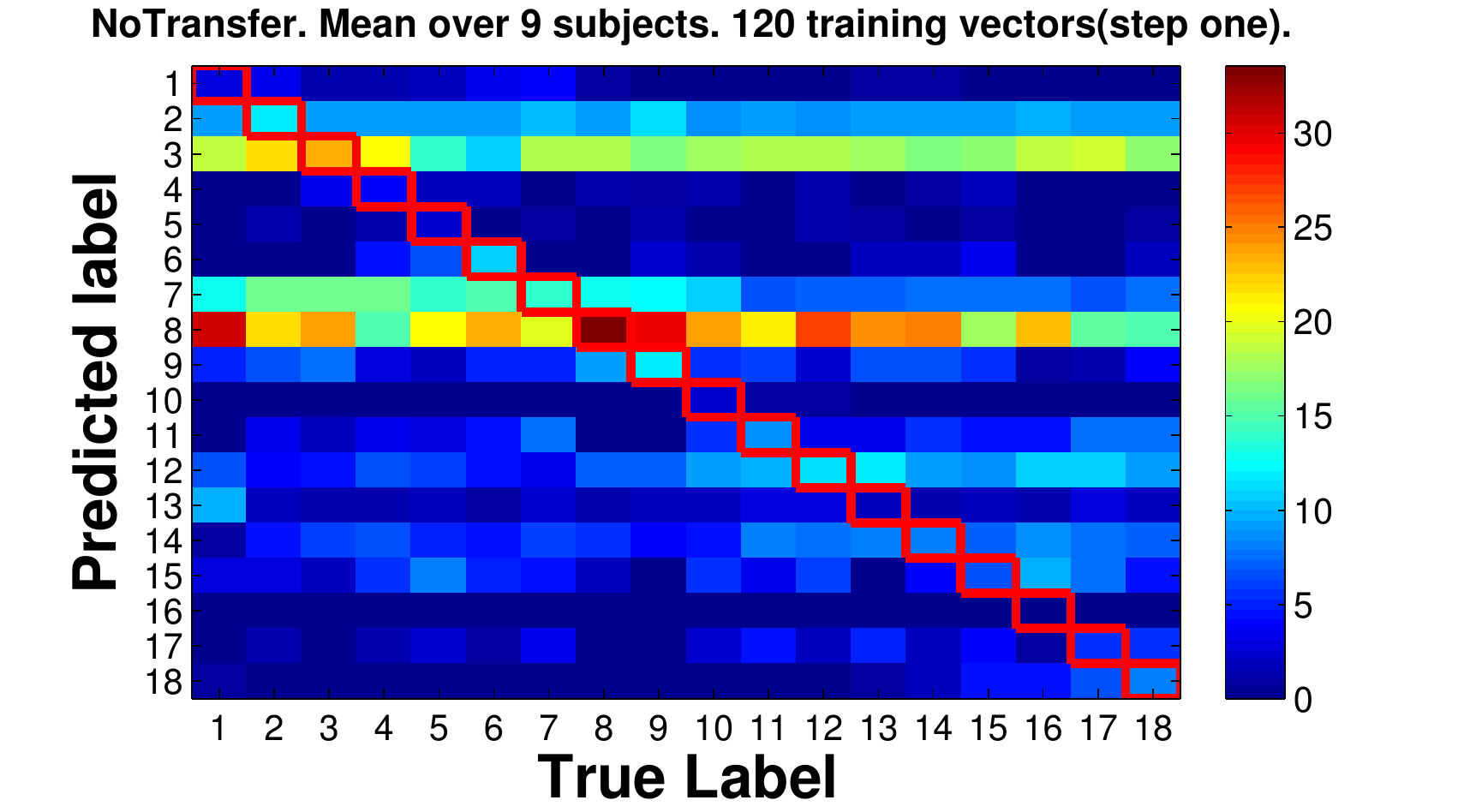}}
 \enspace
 \subfigure
   {\includegraphics[scale=0.38]{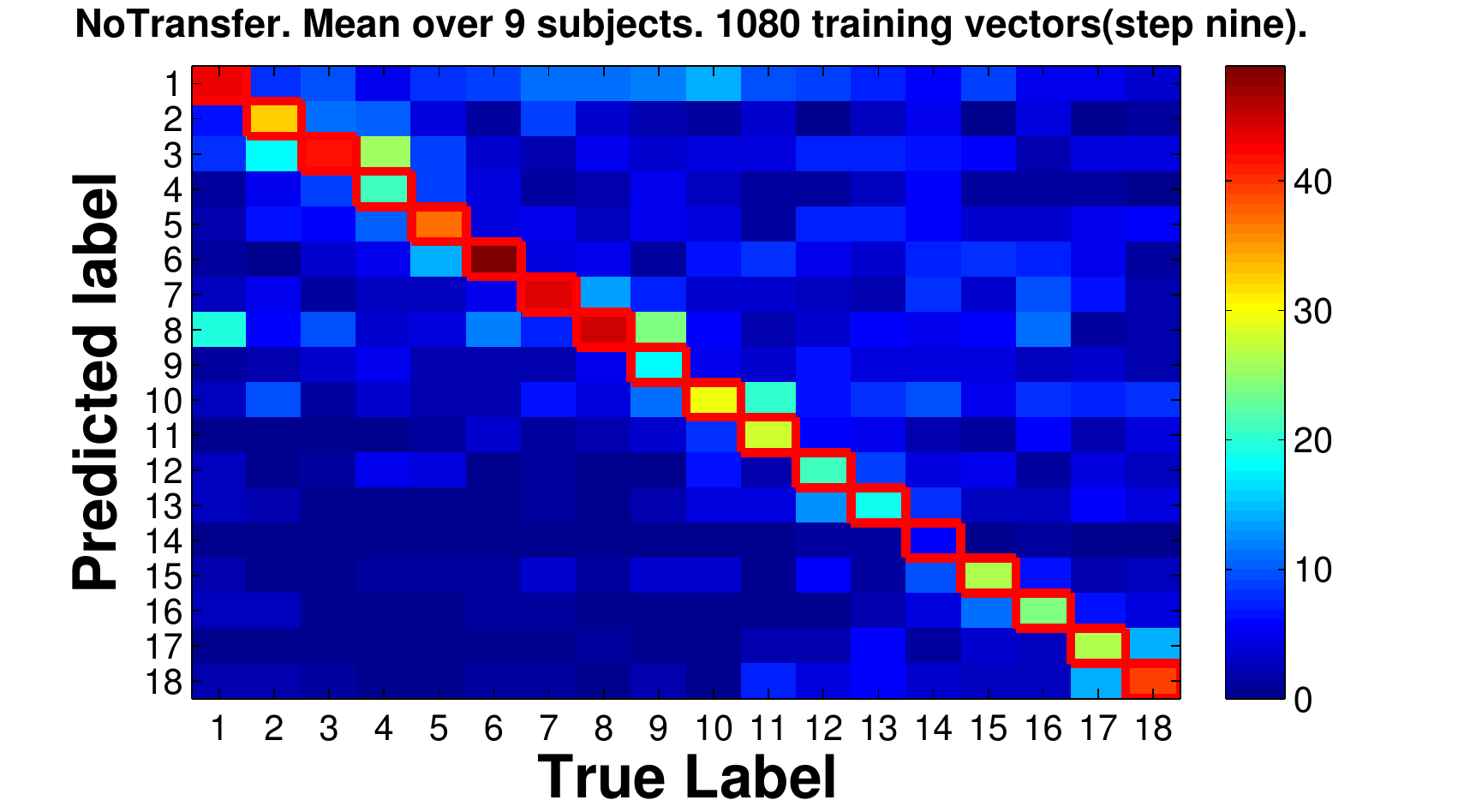}}
   \enspace
 \subfigure
   {\includegraphics[scale=0.38]{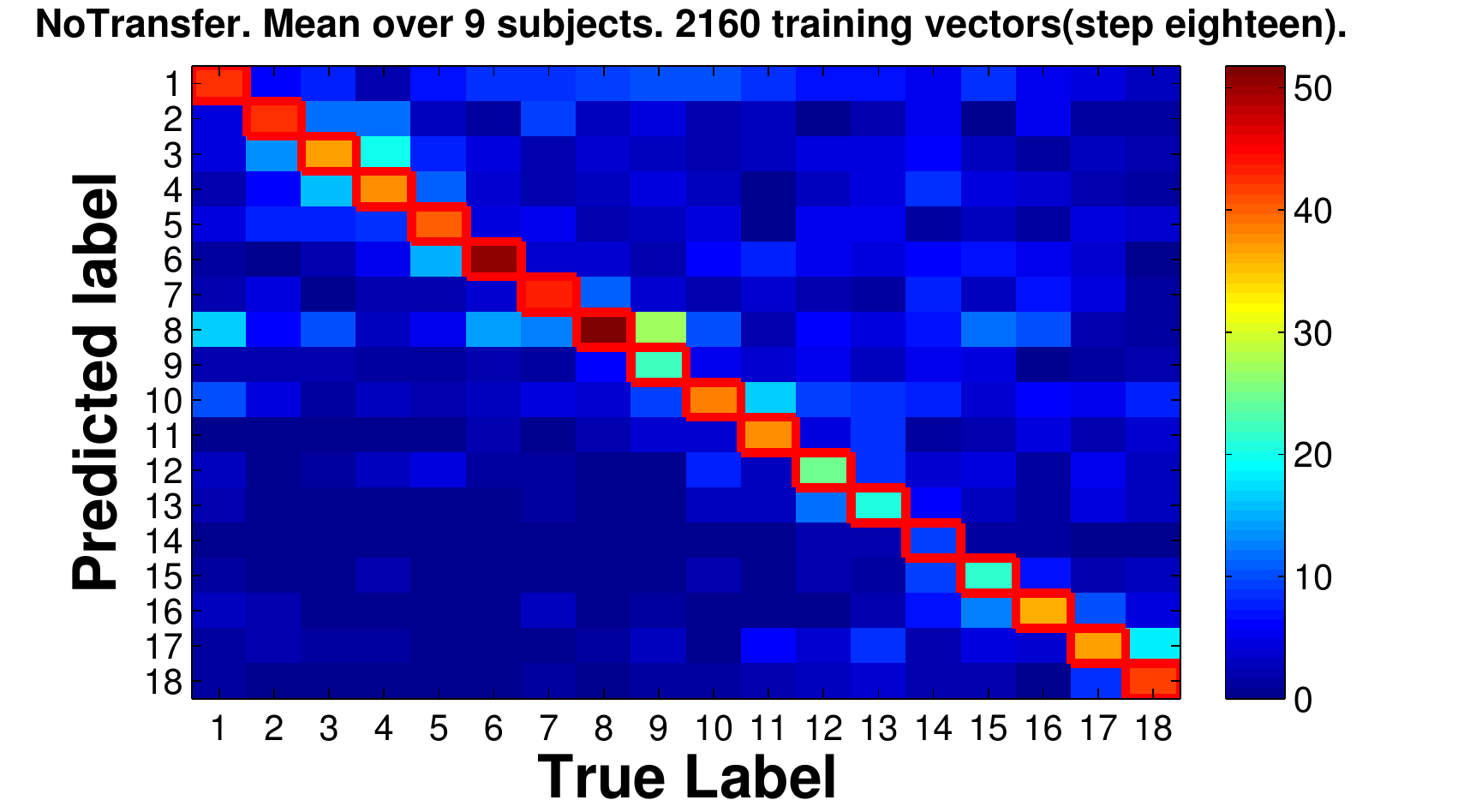}}
  \caption{Confusion matrices for No Transfer method 120, 1080 and 2160 training vectors.}\label{fig:NTConfMatA} 
\end{figure}

\begin{figure} [H]
\centering
\subfigure
   {\includegraphics[scale=0.38]{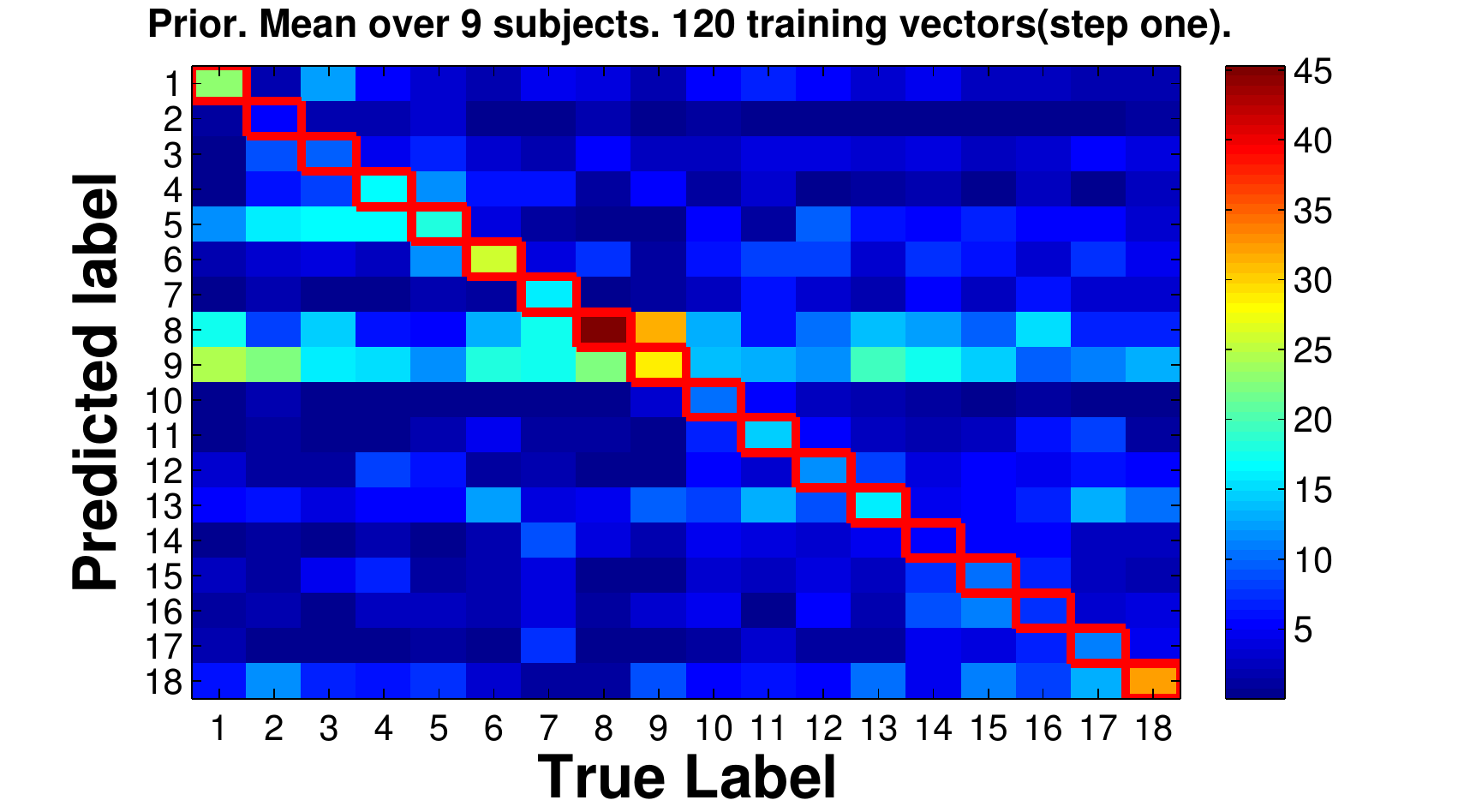}}
 \enspace
 \subfigure
   {\includegraphics[scale=0.38]{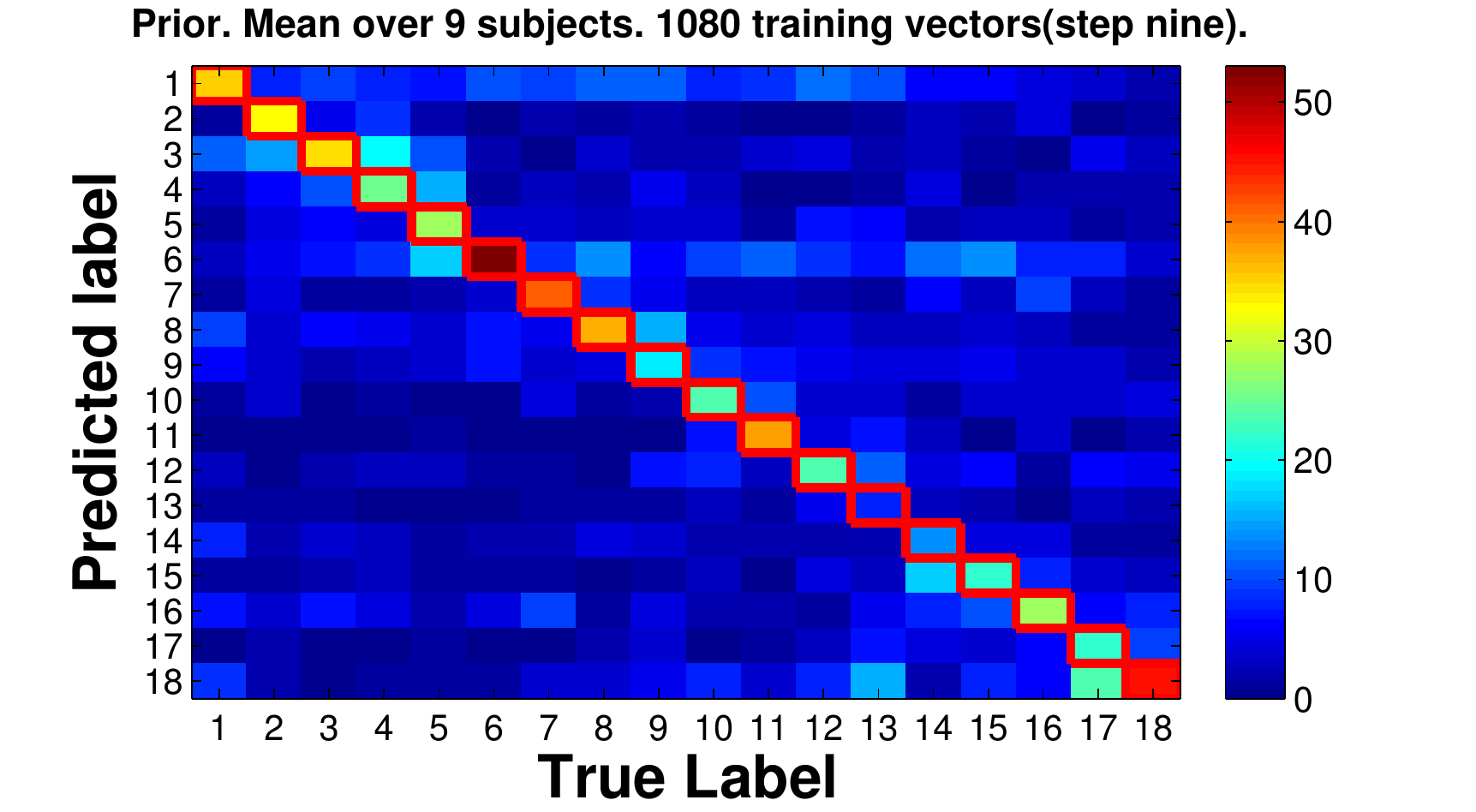}}
   \enspace
 \subfigure
   {\includegraphics[scale=0.38]{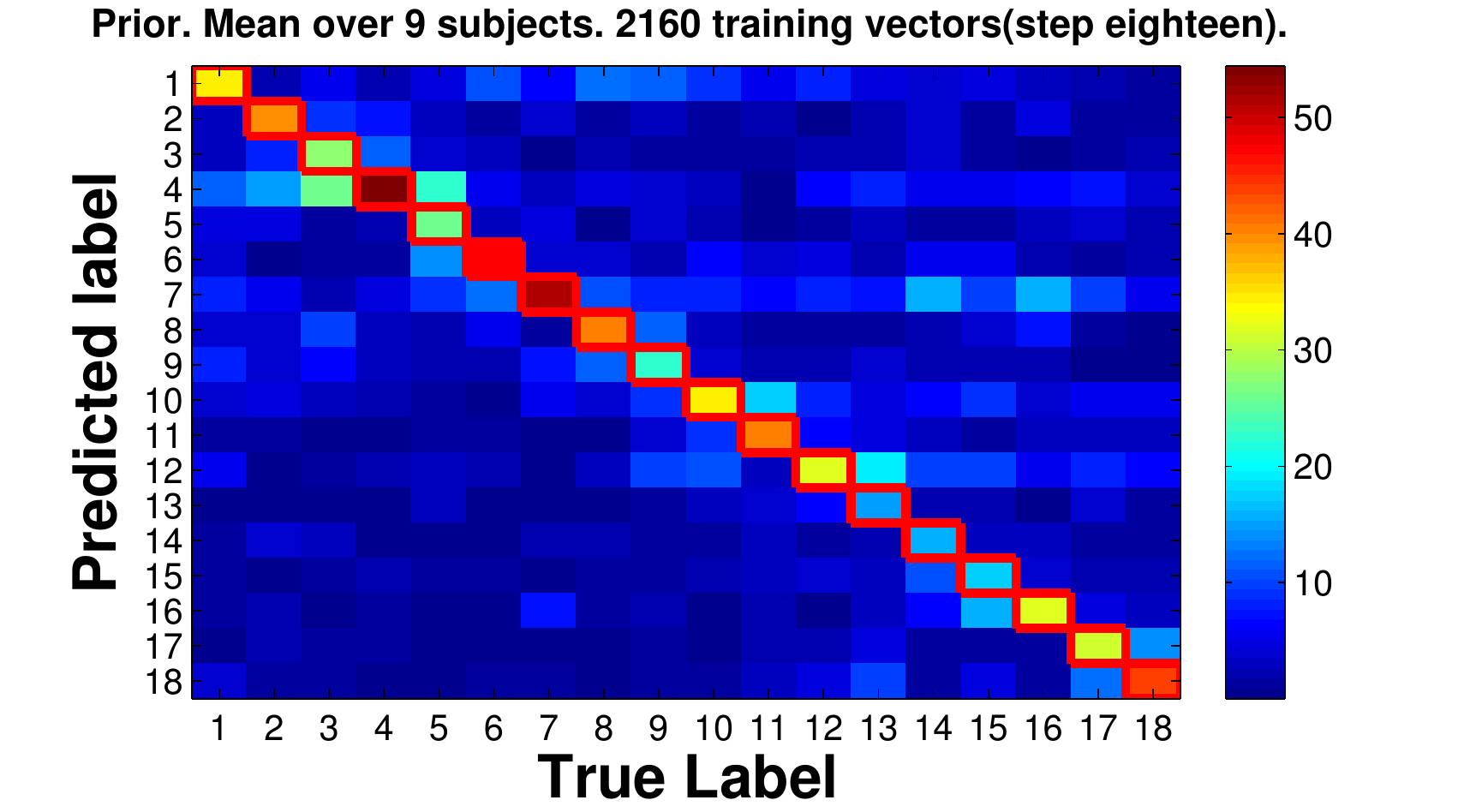}}
  \caption{Confusion matrices for Prior method 120, 1080 and 2160 training vectors.}\label{fig:PriorConfMatA} 
\end{figure}

\begin{figure} [H]
\centering
\subfigure
   {\includegraphics[scale=0.38]{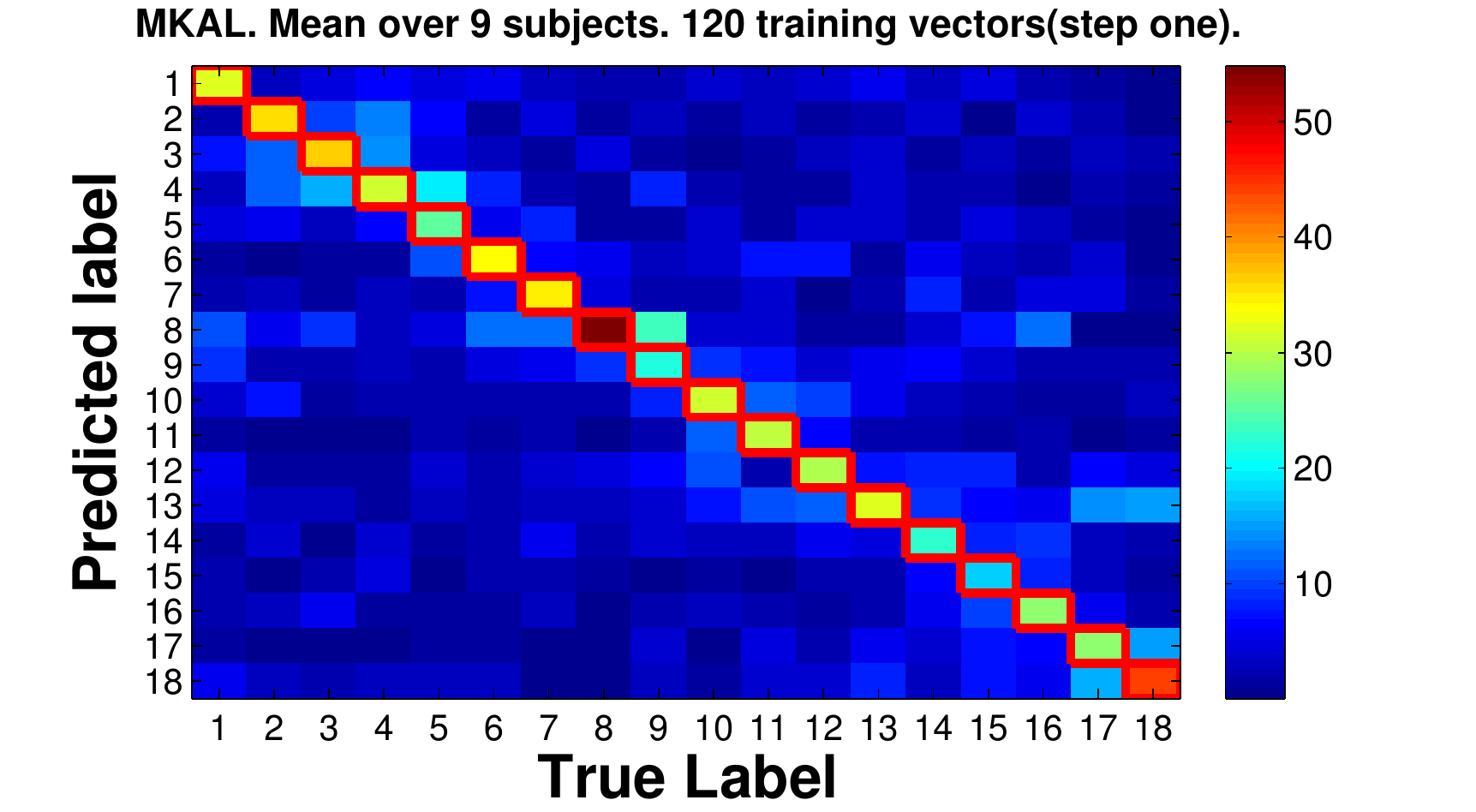}}
 \enspace
 \subfigure
   {\includegraphics[scale=0.38]{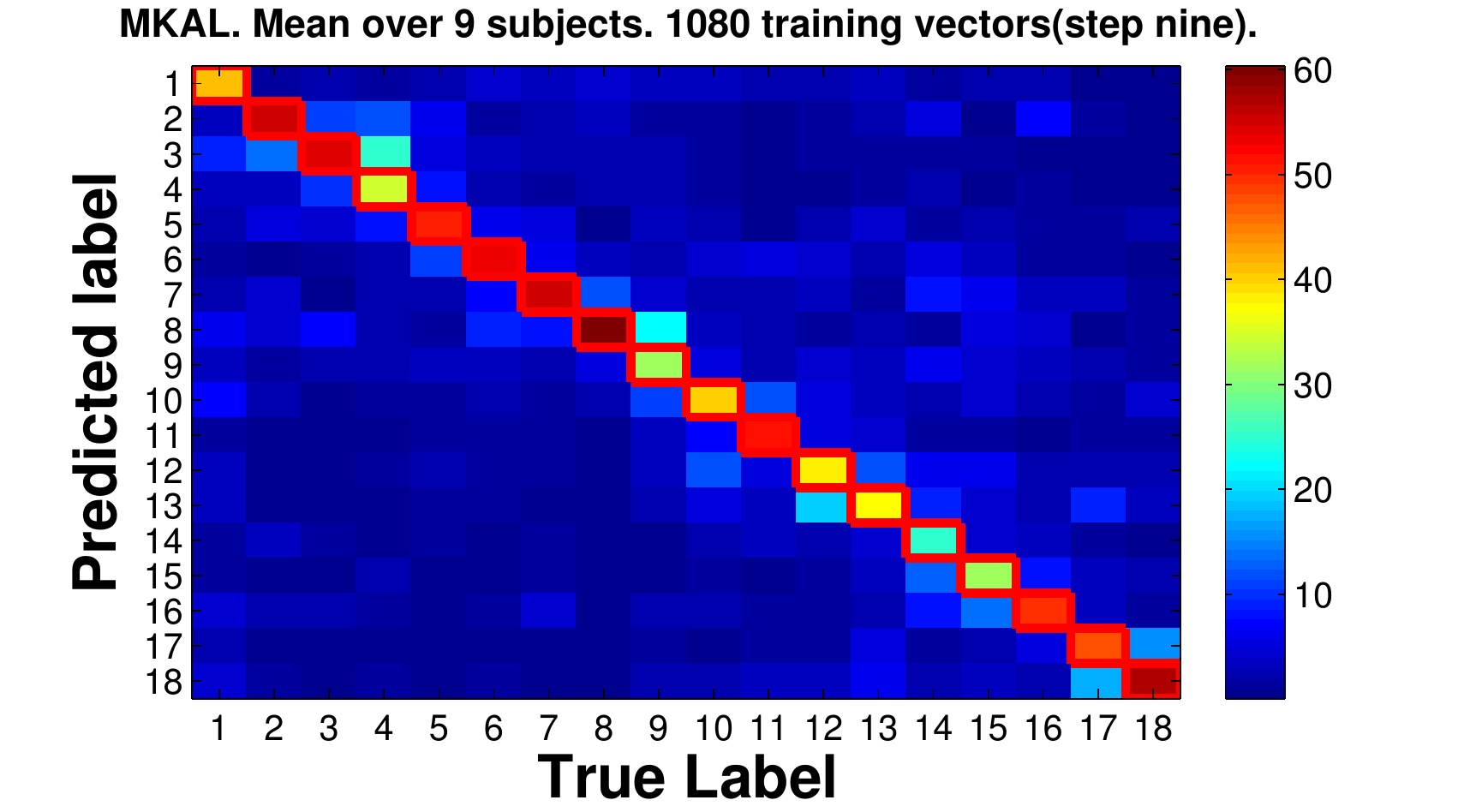}}
   \enspace
 \subfigure
   {\includegraphics[scale=0.38]{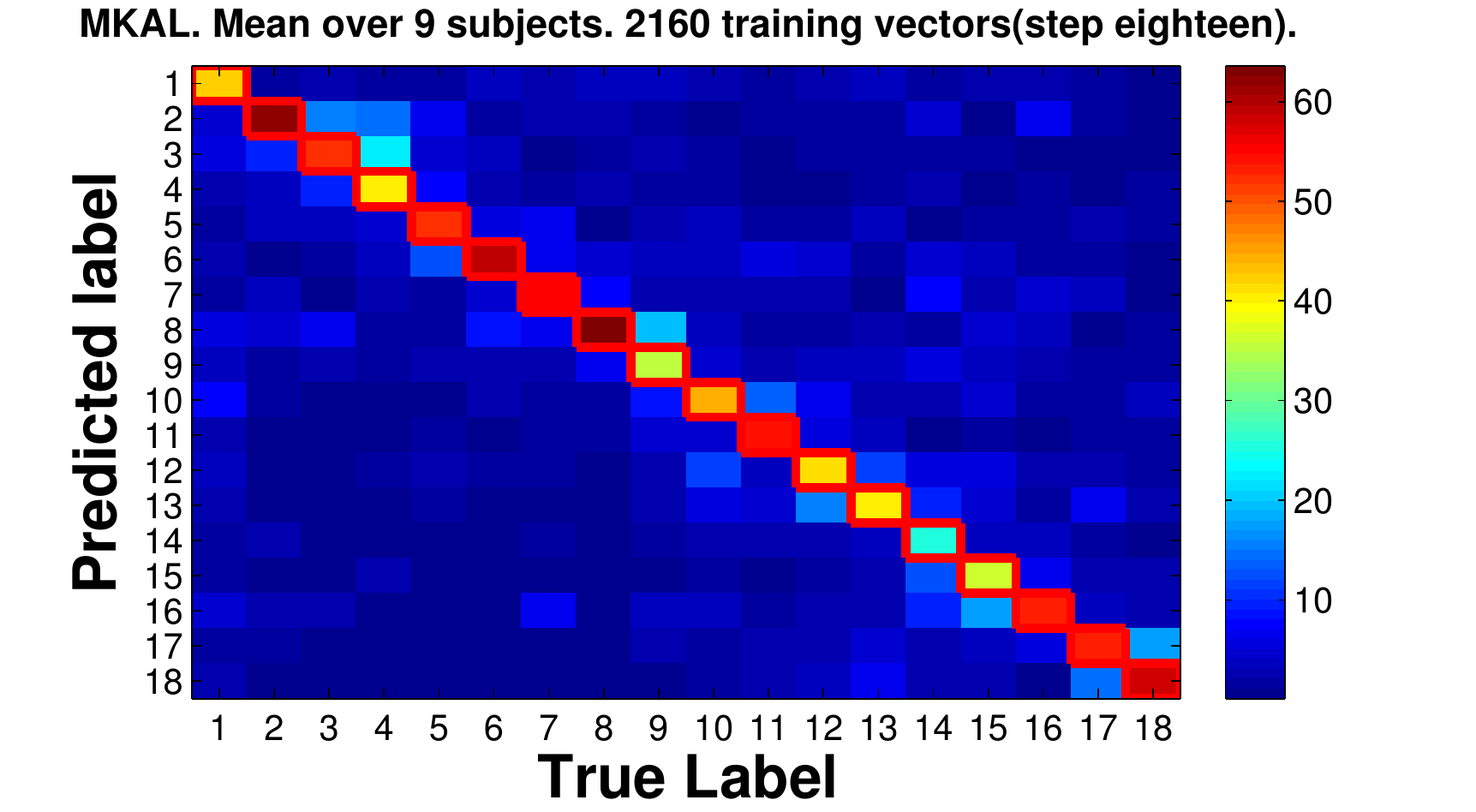}}
  \caption{Confusion matrices for MKAL method 120, 1080 and 2160 training vectors.}\label{fig:MKALConfMatA} 
\end{figure}

\paragraph{Results: Confusion Matrices.} As done for intact subjects in previous section we can analyse the confusion matrices in order to evaluate the level of recognition and misclassification of a single class changing the number of training vectors and the algorithm.\\
We report the confusion matrices for No Transfer (Figure \ref{fig:NTConfMatA}), Prior Features (Figure \ref{fig:PriorConfMatA}) and MKAL (Figure \ref{fig:MKALConfMatA}) for 120 (i.e. initial step), 1080 (i.e. middle step) and 2160 (i.e. final step) training vectors. The label 1 is associated to rest posture.\\
One can find the confusion matrices for all algorithms in \url{https://sites.google.com/site/noninvasiveprosthetichand/}.\\

As explained previously the warm colors are associated to high probability and the cool colors to low probability. In ideal case the prediction labels are equal to the true ones with the maximum probability and, graphically, we have red cells on the diagonal and blue outside. In all the three reported cases, augmenting the number of training vectors, the warm colours move towards the diagonal.\\
No Transfer makes prediction exploiting only the training vectors. At the first step there is an imbalance due to the training samples considered. From first image of Figure \ref{fig:NTConfMatA} it is evident that the majority of first 120 training vectors belong to classes 2, 3, 7 and 8. At the last step, with 2160 training vectors, this displacement is solved. Only the predictions associated to class 8 are higher than others out of the diagonal, but the highest prediction is always associated to the right classes except two cases.\\ 
With Prior, using only 120 training vectors, the highest prediction is associated to classes 8 and 9. Passing to 1080 training vectors the highest prediction is inside the diagonal for all classes except three. At the last step only a class suffers of this problem.\\
With MKAL only a class has the highest prediction outside the diagonal using only 120 training vectors, but the problem is solved at the next steps.\\

Also in this case we evaluate for each real class the first four classes with highest predictions. This analysis is done in order to understand if there is an adaptive method that recognizes a class better than others, or if the misclassification changes increasing the number of training vectors.\\
The details of this analysis are reported in the Appendix \ref{sec:HistA}.\\
The obtained results show that, generally, a class is always misclassified with the same wrong classes and it is independent from the number of training vectors and the adaptive algorithm used.\\

\section{Amputees-Intact} \label{sec:AI}

\paragraph{Setup.} The third experiment involves 9 amputated subjects and 20 intact subjects from respectively the third and the second sub-database of NinaPro (see Table in section \ref{sec:Ninapro_dataset}). Considering previous results we used the same amputated and intact subjects exploited in the first and second experiment.\\
In this set-up we consider amputees as target and intact subjects as sources. In particular the sources are fixed and, one by one, we consider as target an amputated subject. Thus, we have an amputee that learns to use the prosthetic hand with the help from the prior knowledge of intact subjects.\\
The data taken into account are those of the Exercise B: 8 hand configurations and 9 basic movements of the wrist.\\
In the training phase we build a classification model with each of the five algorithms for each target subject. The process is repeated for an increasing number of training vectors with steps of $120$ up to a maximum of $2160$. In the test phase we evaluate the performance of each model.\\
In the Tables \ref{tab:DatasetAI} and \ref{tab:DatasetAIAlg} we report the principal characteristics of third experiment.\\

\begin{table}[H]
	
	\resizebox{13,5cm}{!}	{
	
	\begin{tabular} {|| l | l | p{7cm} | l ||}
	
	\hline
	\hline
	  & \textbf{Database} & \textbf{Subjects} & \textbf{Postures} \\ 		\hline
	\hline
	\textbf{Target} & 3: Amputees & 9 no-random: 1, 2, 3, 4, 5, 6, 9, 10, 11  & 17 + rest\\    \hline
	\textbf{Source} & 3: Intact & 20 no-random: 1, 3, 4, 5, 9, 10, 12, 20, 21, 22, 24, 25, 29, 30, 31, 34, 35, 38, 39, 40  & 17 + rest\\   \hline
	\hline
	\end{tabular} \\
	}
	\caption{Characteristics of used data.}
	\label{tab:DatasetAI}
\end{table}

\begin{table}[H]
	\quad
	
	\resizebox{14cm}{!}	{
	
	\begin{tabular}{|| l | p{6,5cm} | p{4cm} ||}
	
	\hline
	  & \textbf{Adaptive} & \textbf{Baseline}  \\ 		\hline
	\hline	
	\textbf{Algorithms} & Multi Kernel Adaptive Learning, High Level-Learning2Learn, Multi Adapt & No Transfer, Prior Features \\ \hline 
	\hline

	\end{tabular}

	}
	
	\caption{Used algorithms.}
	
\label{tab:DatasetAIAlg}
\end{table}

\begin{figure} [H]
\centering
\includegraphics [scale = 0.4] {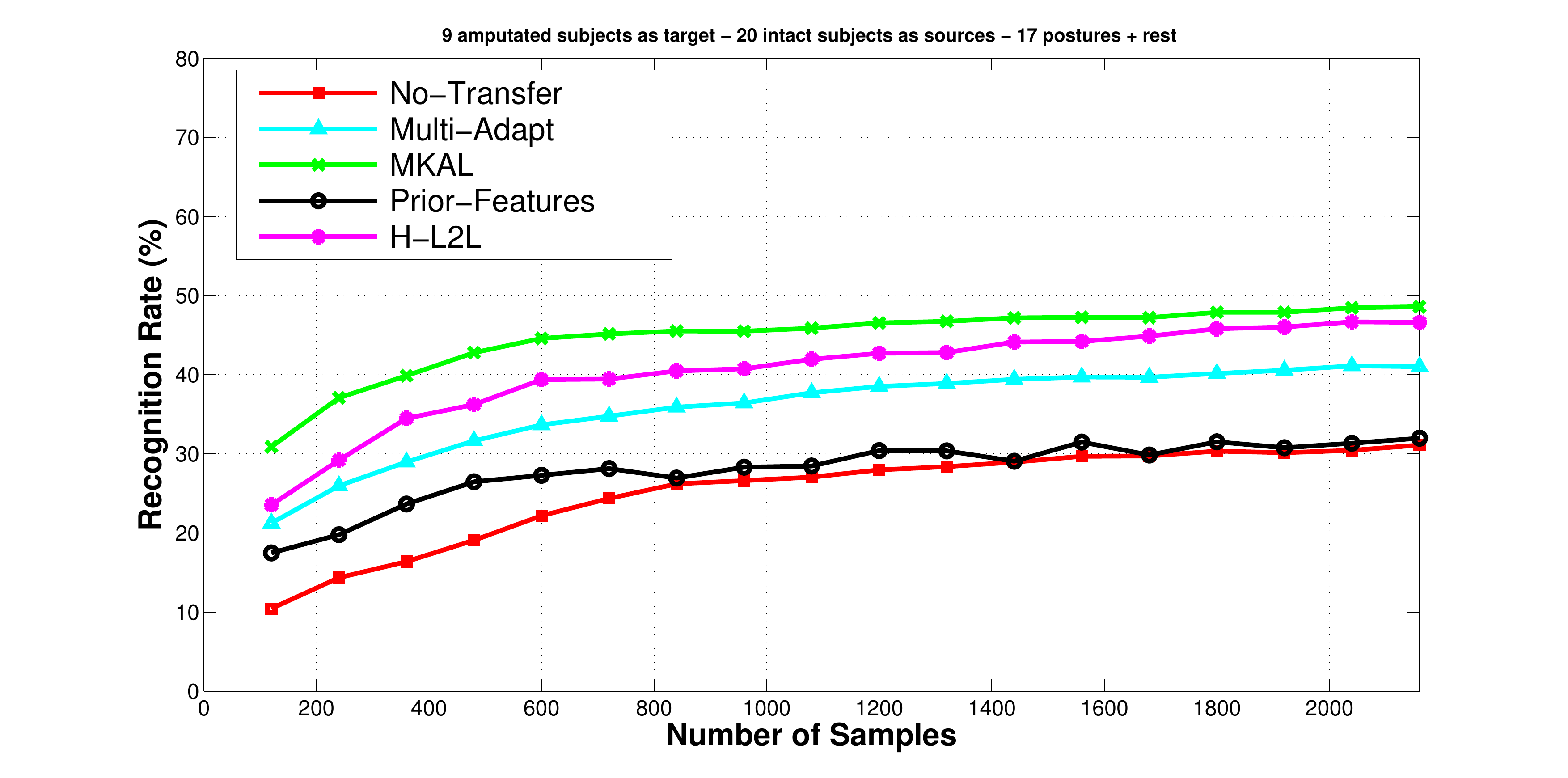}
\caption{Classification rate obtained by averaging over all the subjects as a function of the number of samples in the training set. }\label{fig:9Amp20IntSubj}
\end{figure}

\paragraph{Results: Recognition Rate.} The trend of performance averaged over all the subjects as a function of the training vectors of the target problem is reported in Figure \ref{fig:9Amp20IntSubj} for all the algorithms. It comes from the mean between performance obtained for all the target subjects.\\

\begin{figure} [H]
\centering
\subfigure
   {\includegraphics[scale=0.38]{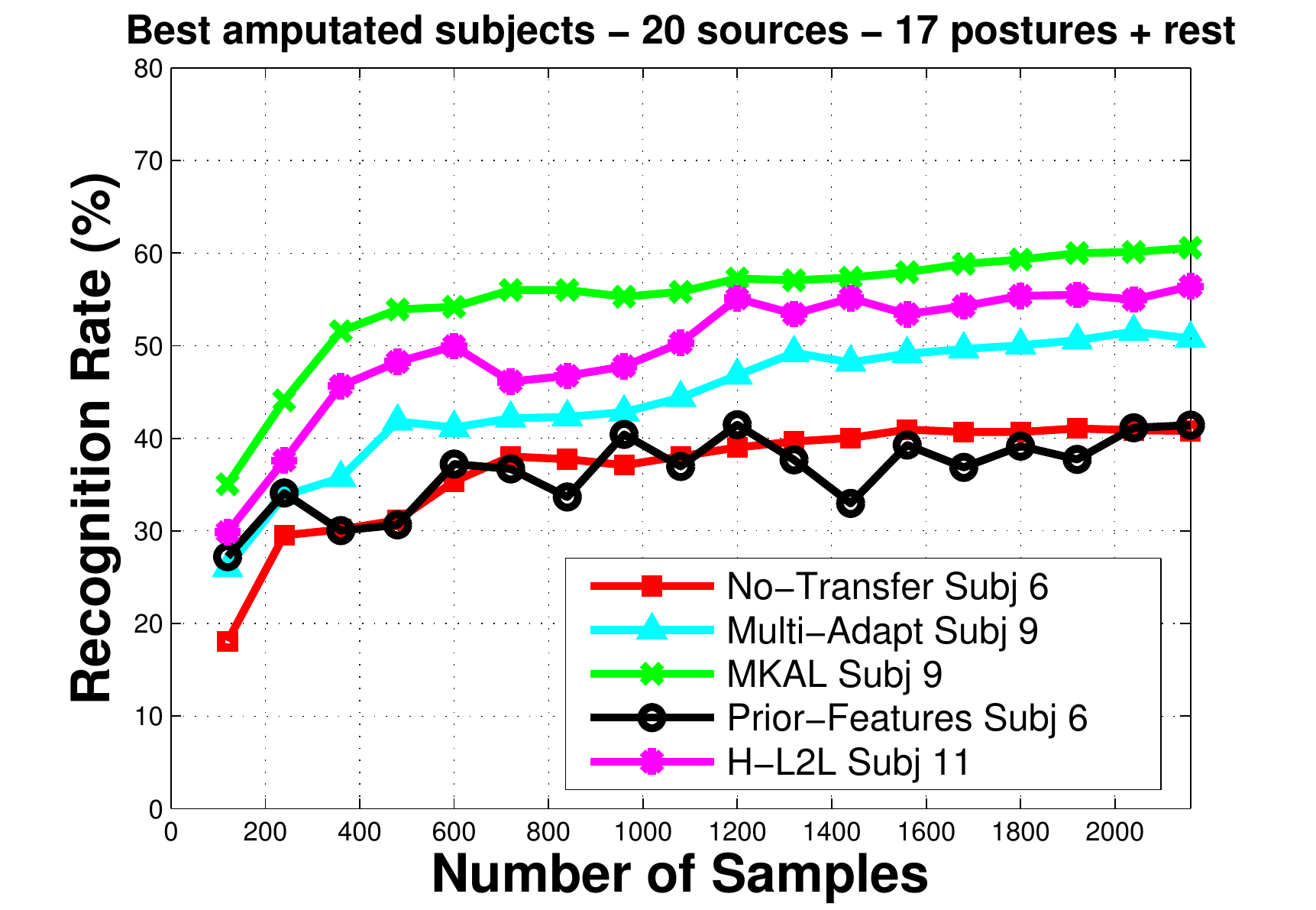}}
 \enspace
 \subfigure
   {\includegraphics[scale=0.38]{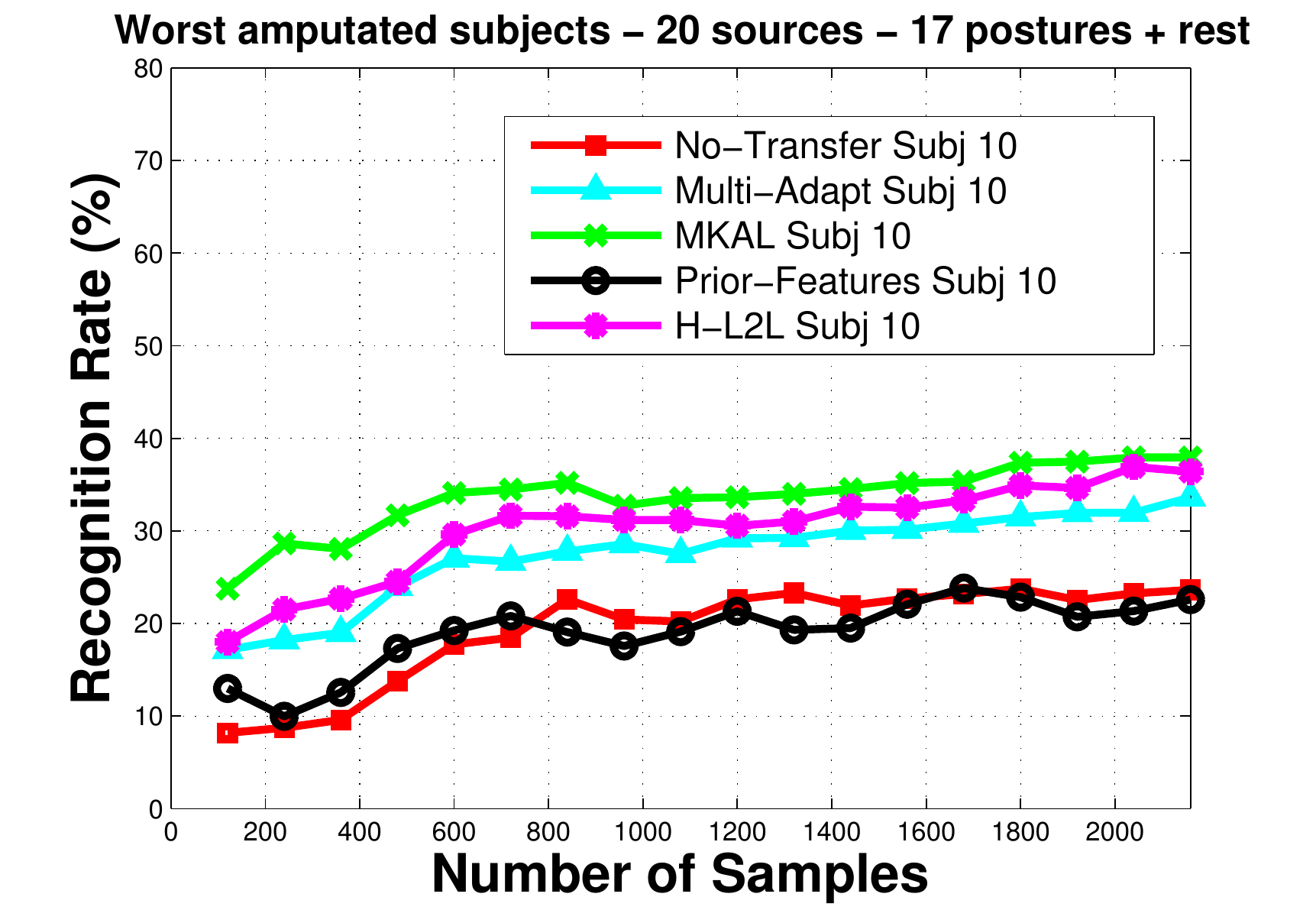}}
  \caption{Classification rate for the best and worst subjects as a function of the number of samples in the training set.}\label{fig:9Amp20IntSubjBW} 
\end{figure}

In Figure \ref{fig:9Amp20IntSubjBW} the best and worst cases are reported. These are respectively the subjects for which each algorithm gives the best and worst result in performance. 

MKAL achieves the best performance asymptotically and considering the whole trend. It is followed by H-L2L, Multi Adapt, Prior Features and No Transfer. A similar order is preserved in the best and worst case but a component of noise is present for single subject and the trend appears not very smooth. We can note that No Transfer and Prior Features perform in the same way and their differences are not statistically significant ($p > 0.05 $).\\
As shown in Figure \ref{fig:9Amp20IntSubj} MKAL outperforms H-L2L with an average of 4 $\%$ ($p < 0.05$). The difference between H-L2L and Multi Adapt shows an average advantage in recognition rate of around 5 $\%$ ($p < 0.05 $). Multi Adapt has an average gain of 8 $\%$ with respect to Prior Features ($p < 0.05 $).\\ 
Until 720 training vectors Prior Features outperforms No Transfer with an average of 6 $\%$, after the two curves keep the same performance. Multi Adapt, MKAL and H-L2L outperform No Transfer with an average of about 11 $\%$, 20 $\%$ and 15 $\%$ respectively. At 2160 training samples, i.e. the last step, No Transfer achieves a performance of 31 $\%$. The same performance is reached by Multi Adapt, MKAL and H-L2L with only 480, 240 and 360 training vectors, respectively. Also in this case, the use of prior knowledge allows us to reduce by one order of magnitude the training time.\\
The adaptive methods achieve faster than No Transfer the asymptotic performance. In fact, passing from 600 training samples to 2160 the performance of No Transfer, Multi Adapt, MKAL and H-L2L increases of respectively: 10 $\%$, 7 $\%$, 4 $\%$ and 7 $\%$.\\

\begin{figure} [H]
\centering
\subfigure
   {\includegraphics[scale=0.38]{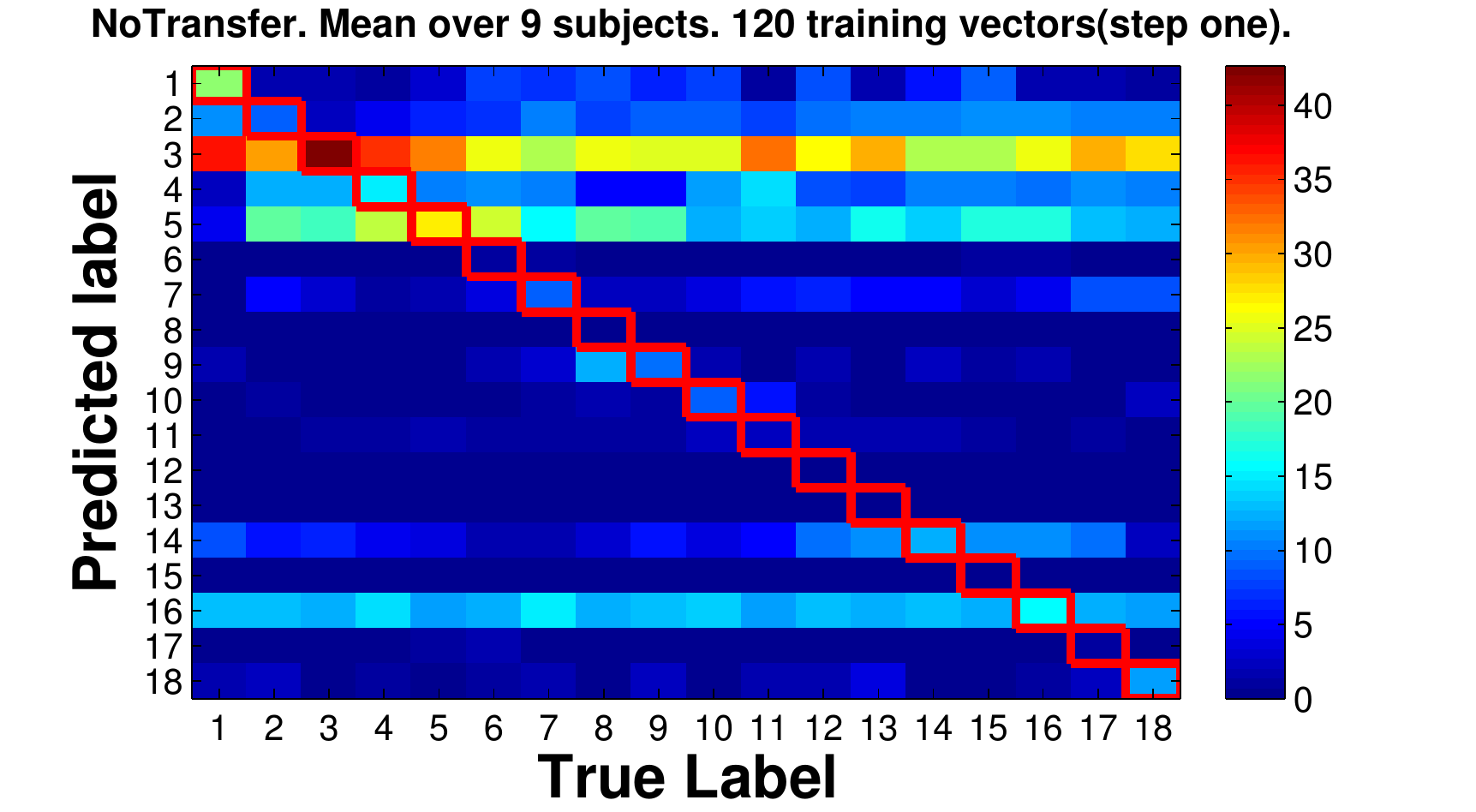}}
 \enspace
 \subfigure
   {\includegraphics[scale=0.38]{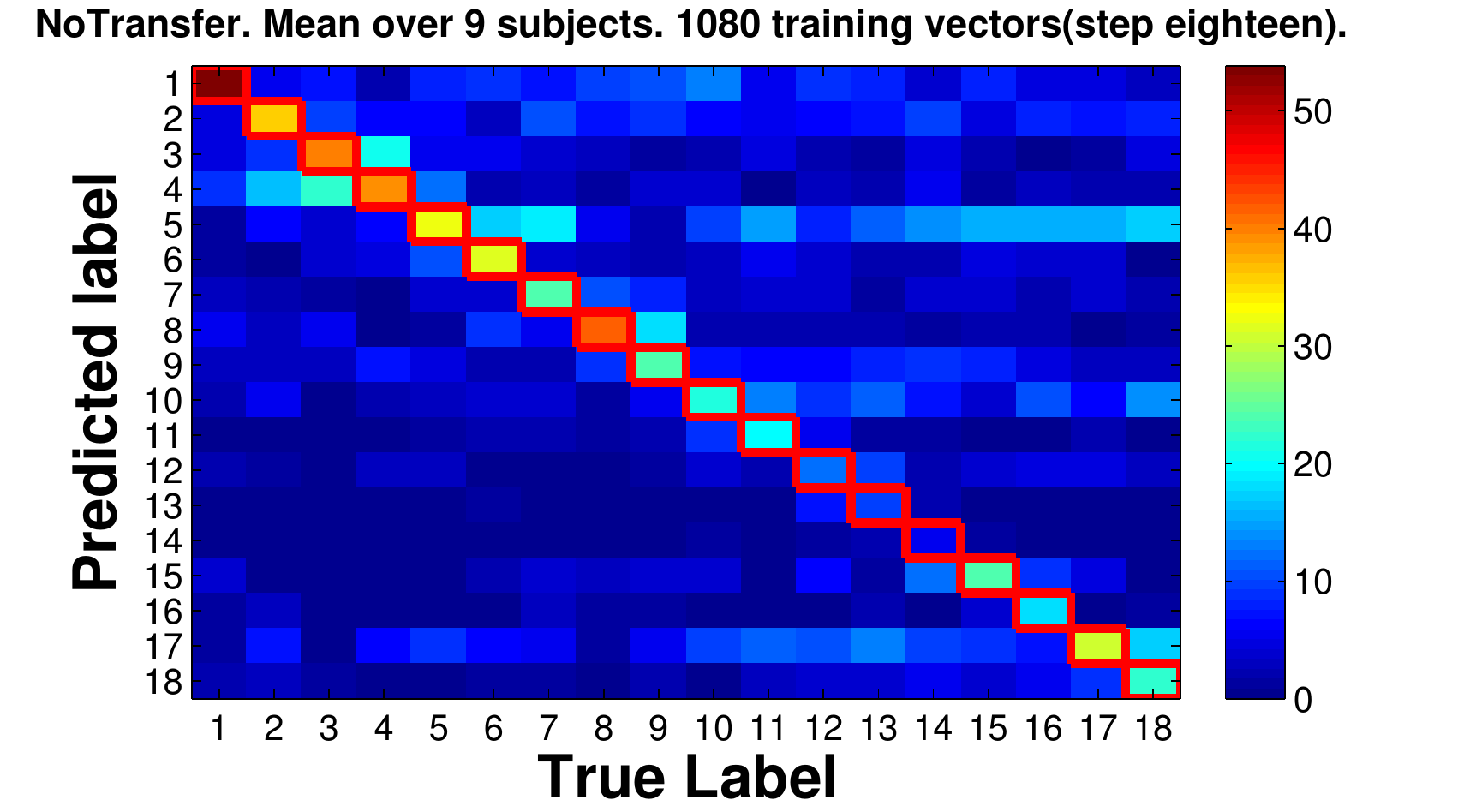}}
   \enspace
 \subfigure
   {\includegraphics[scale=0.38]{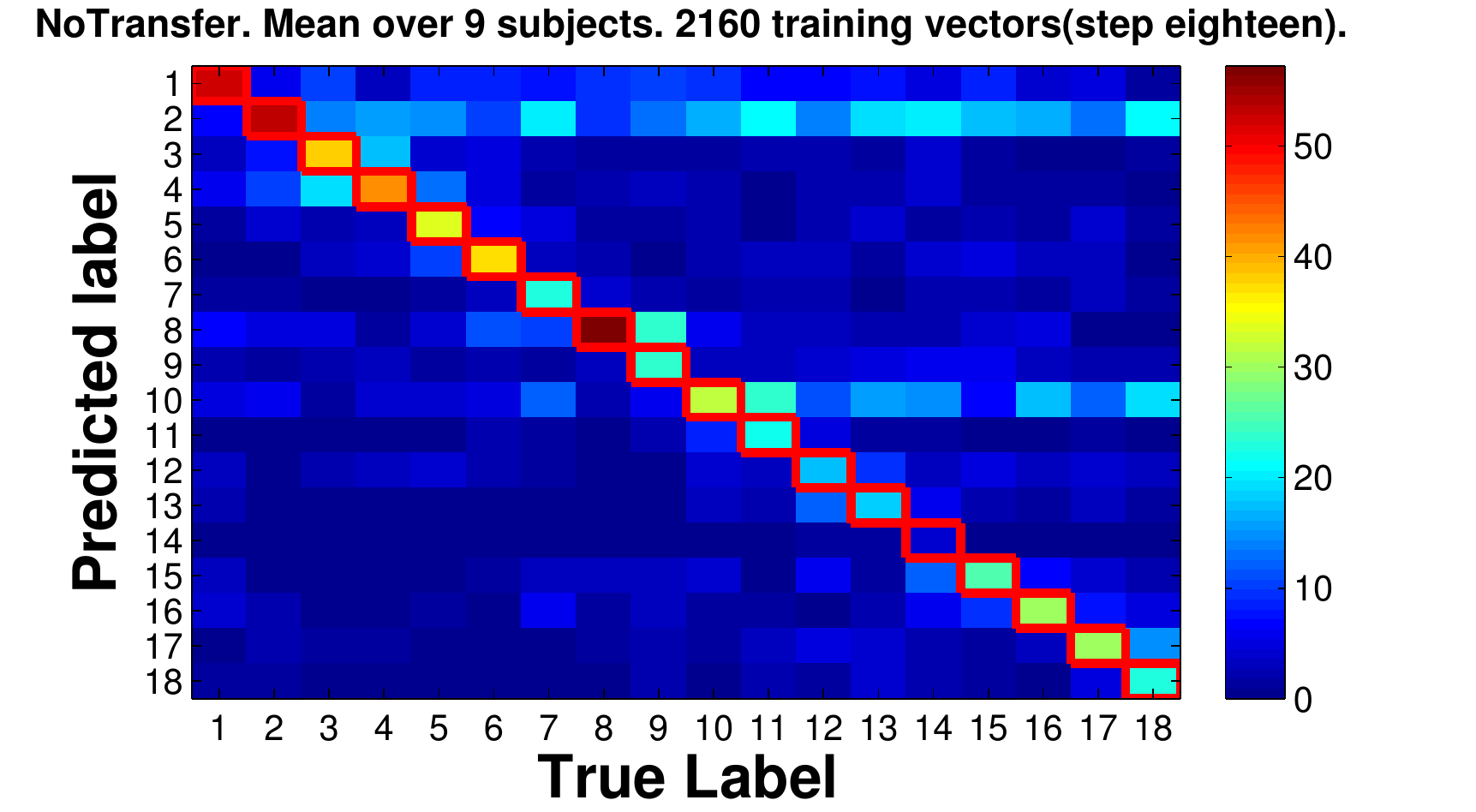}}
  \caption{Confusion matrices for No Transfer method 120, 1080 and 2160 training vectors.}\label{fig:NTConfMatAI} 
\end{figure}

\begin{figure} [H]
\centering
\subfigure
   {\includegraphics[scale=0.38]{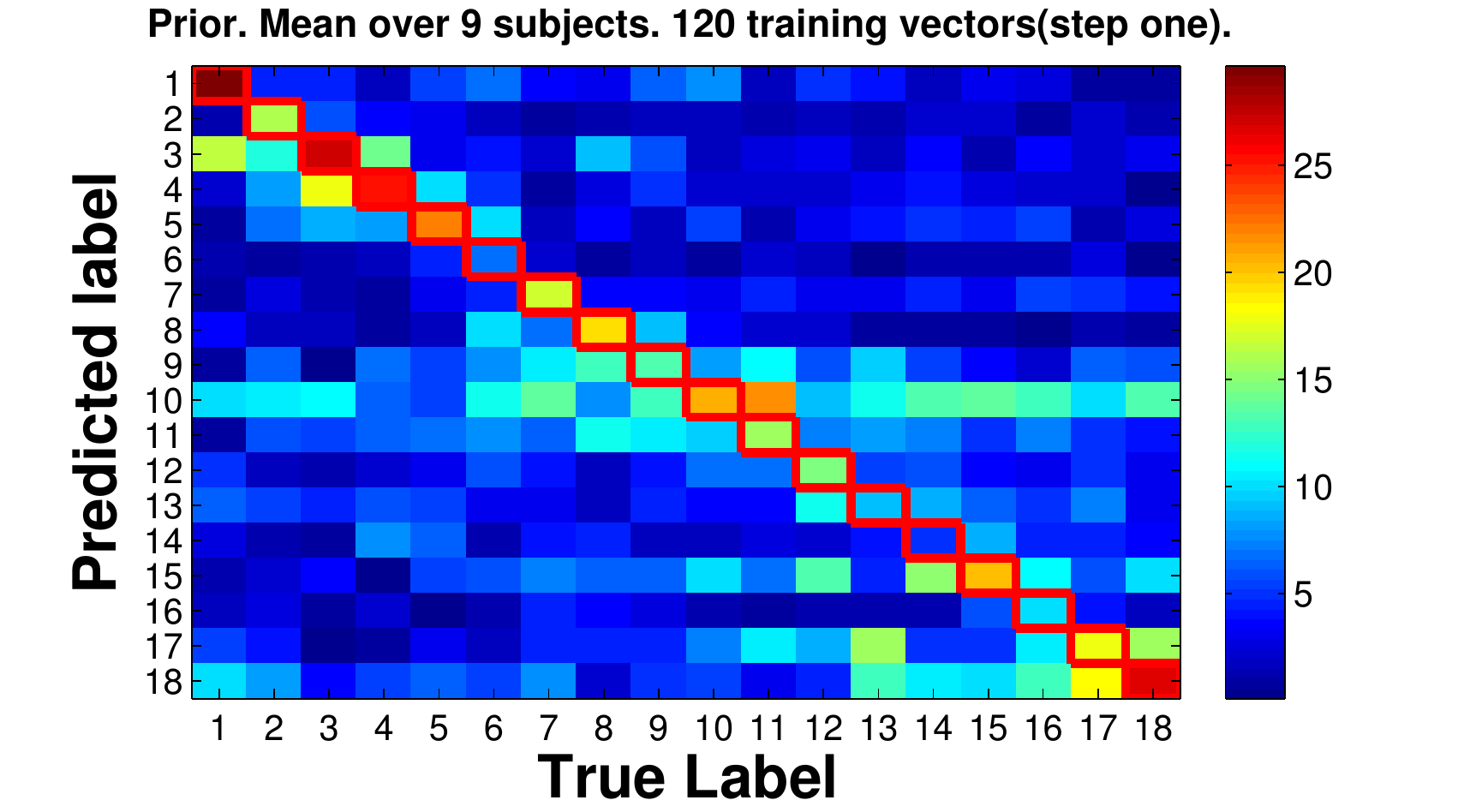}}
\enspace
 \subfigure
   {\includegraphics[scale=0.38]{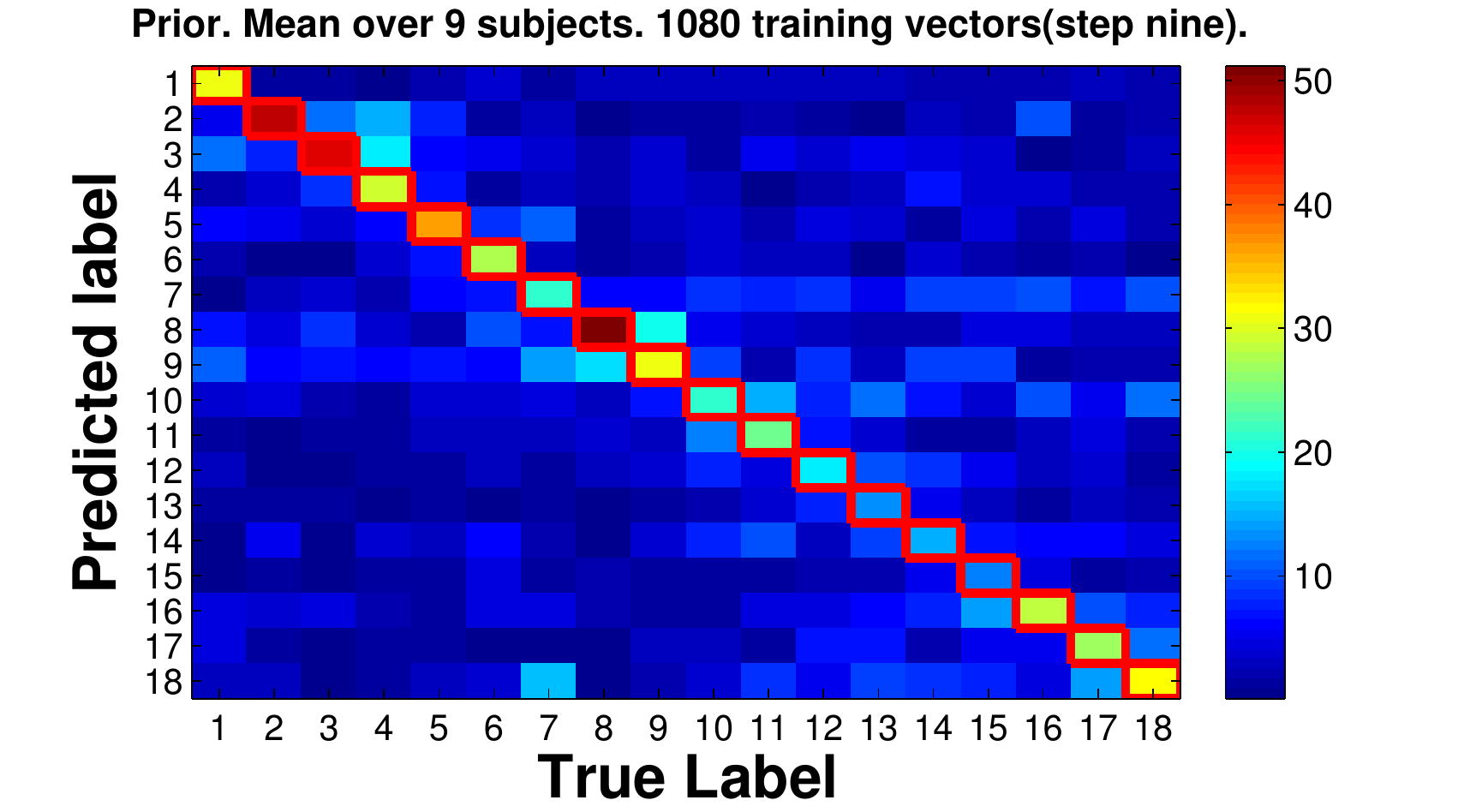}}
   \enspace
 \subfigure
   {\includegraphics[scale=0.38]{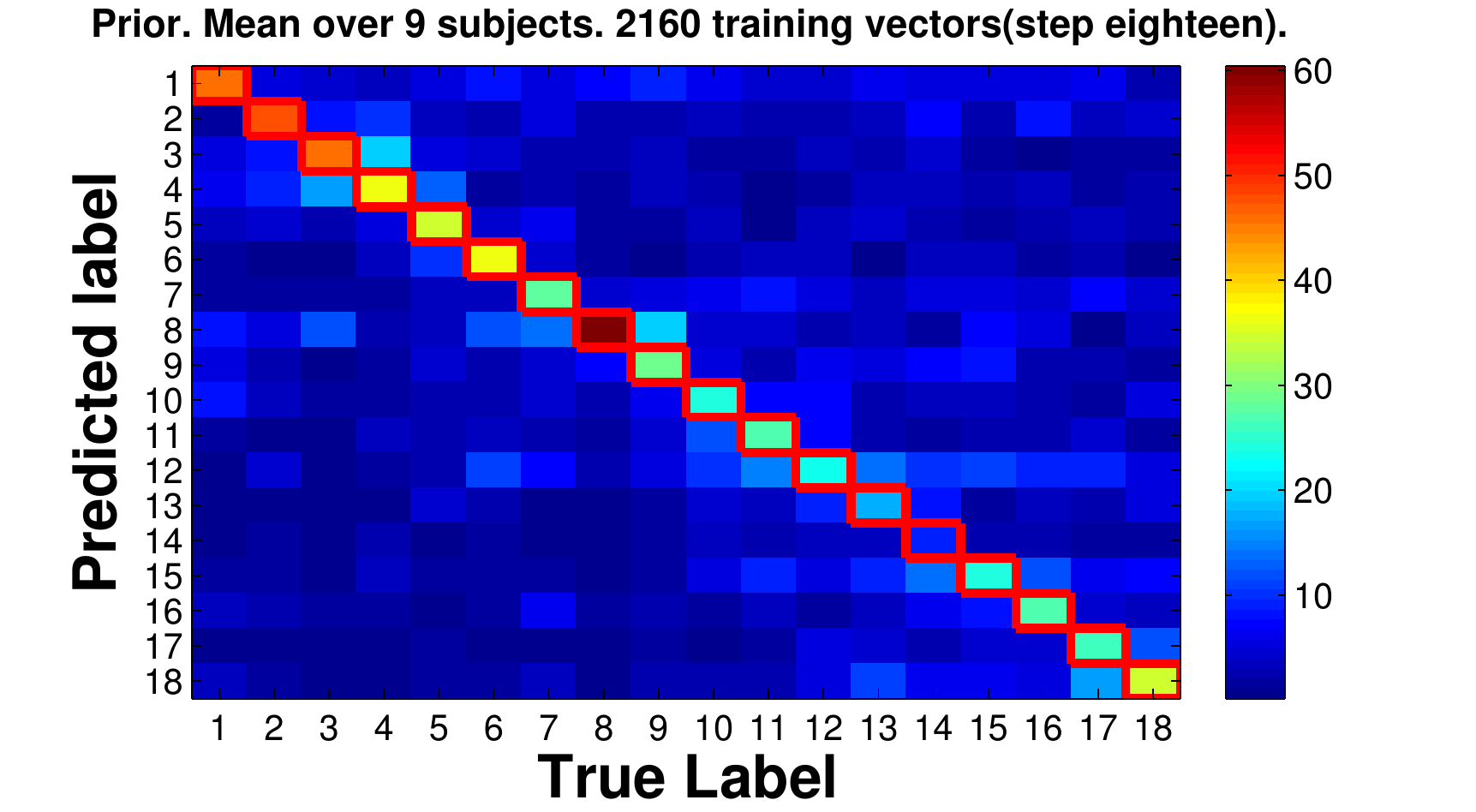}}
  \caption{Confusion matrices for Prior method 120, 1080 and 2160 training vectors.}\label{fig:PriorConfMatAI} 
\end{figure}

\begin{figure} [H]
\centering
\subfigure
   {\includegraphics[scale=0.38]{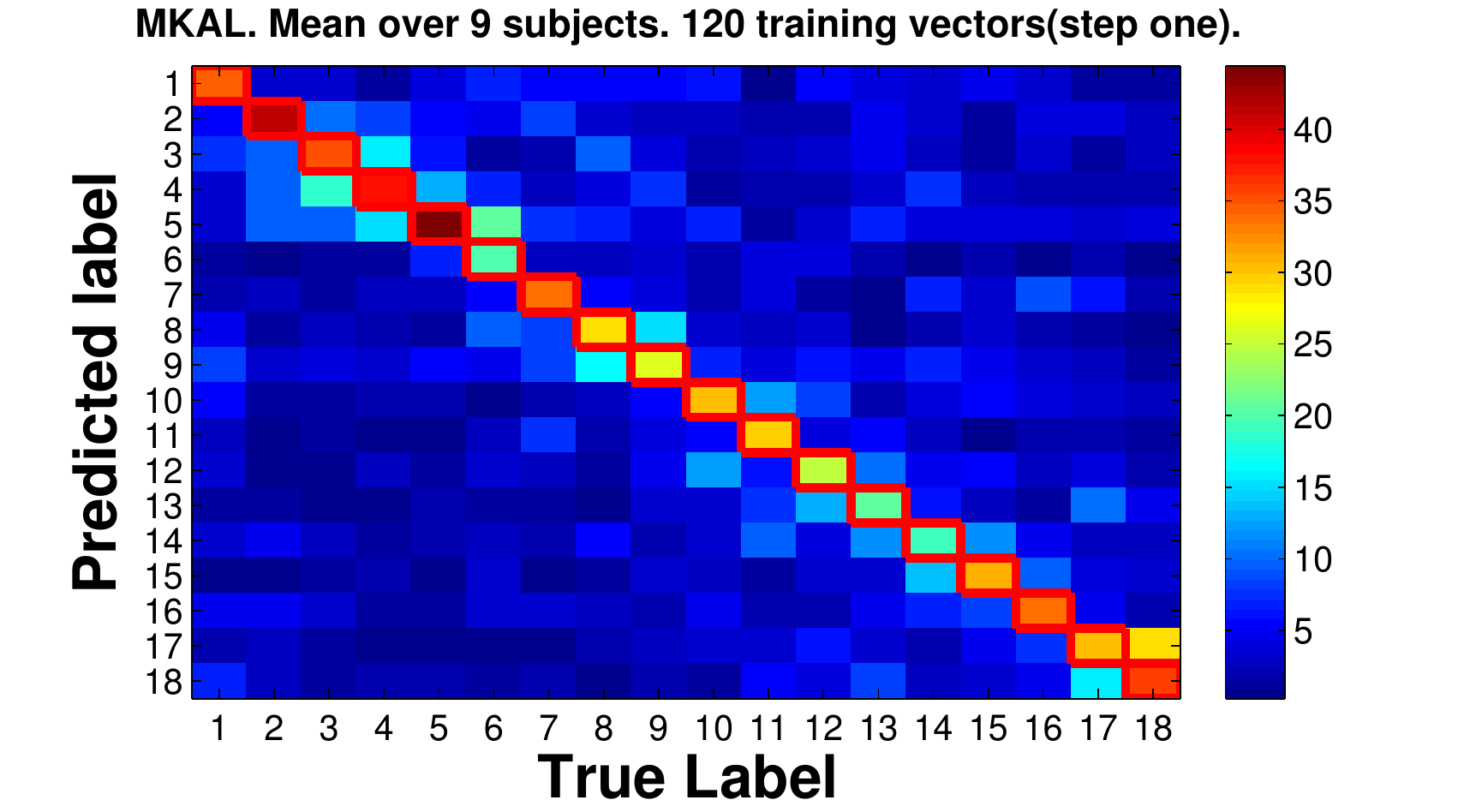}}
 \enspace
 \subfigure
   {\includegraphics[scale=0.38]{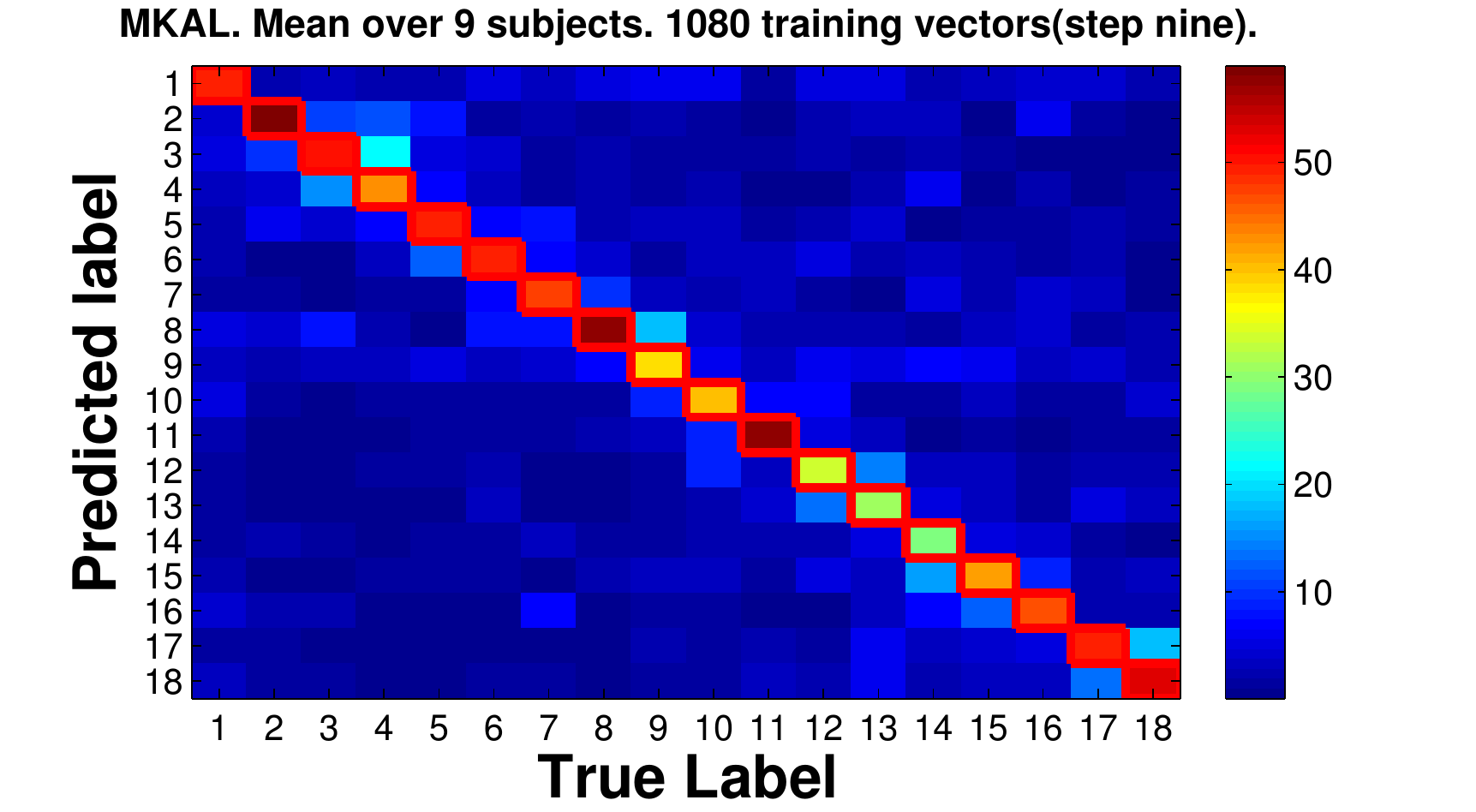}}
   \enspace
 \subfigure
   {\includegraphics[scale=0.38]{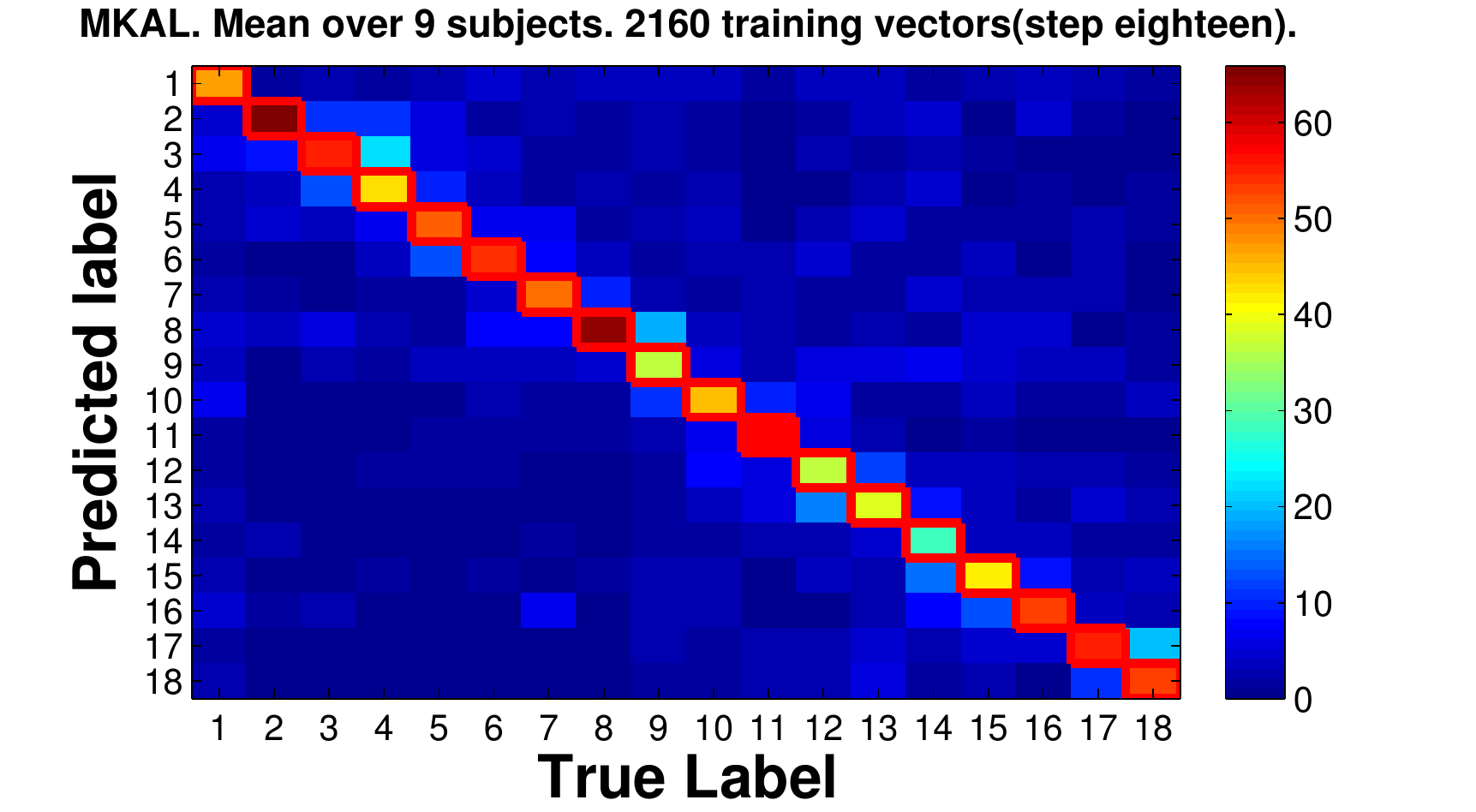}}
  \caption{Confusion matrices for MKAL method 120, 1080 and 2160 training vectors.}\label{fig:MKALConfMatAI} 
\end{figure}

\paragraph{Results: Confusion Matrices.} As done in previous sections we evaluate the level of recognition and misclassification of all classes from analysis of confusion matrices for each algorithm changing the number of training vectors.\\
We report the confusion matrices for No Transfer (Figure \ref{fig:NTConfMatAI}), Prior Features (Figure \ref{fig:PriorConfMatAI}) and MKAL (Figure \ref{fig:MKALConfMatAI}) for 120 (i.e. initial step), 1080 (i.e. middle step) and 2160 (i.e. final step) training vectors. The label 1 is associated to rest posture.\\
One can find the confusion matrices for all algorithms in \url{https://sites.google.com/site/noninvasiveprosthetichand/}.\\

In the matrices the warm and cool colors represent respectively an high and low probability. In the case of a perfect prevision we have red cells on diagonal and blue outside. In all three reported cases, augmenting the number of training vectors, the warm colours move towards the diagonal.\\
No Transfer at first step presents an imbalance in predictions because the models are built with few training vectors. From first image of Figure \ref{fig:NTConfMatA} it is evident that the majority of first 120 training vectors belong to classes 3, 4, 5 and 16. With 120 training samples the predictions associated to classes 2 and 10 are higher than others out of the diagonal. In this case the highest prediction is always associated to right classes except three cases.\\
With Prior Features the highest prediction is associated to classes 10, 11 and 18 at the first step. Passing to 1080 and after to 2160 training vectors the highest prediction is inside the diagonal for all classes except one.\\
With MKAL only a class has the highest prediction outside the diagonal using only 120 training vectors and the problem is solved at next steps.\\

Also in this case we evaluate for each real class the first four classes with highest predictions. From this analysis we can evaluate if there is an adaptive method than in the recognition of a particular class is is better than others.\\
The details of this analysis are reported in the Appendix \ref{sec:HistAI}.\\
As in previous cases, the obtained results show that, generally, a class is always misclassified with the same wrong classes and it is independent from the number of training vectors and the adaptive algorithm used.\\

\section{Discussion and comparison} \label{sec:CompariosonRes}

\begin{figure} [H]
\centering
\subfigure
   {\includegraphics [scale = 0.44] {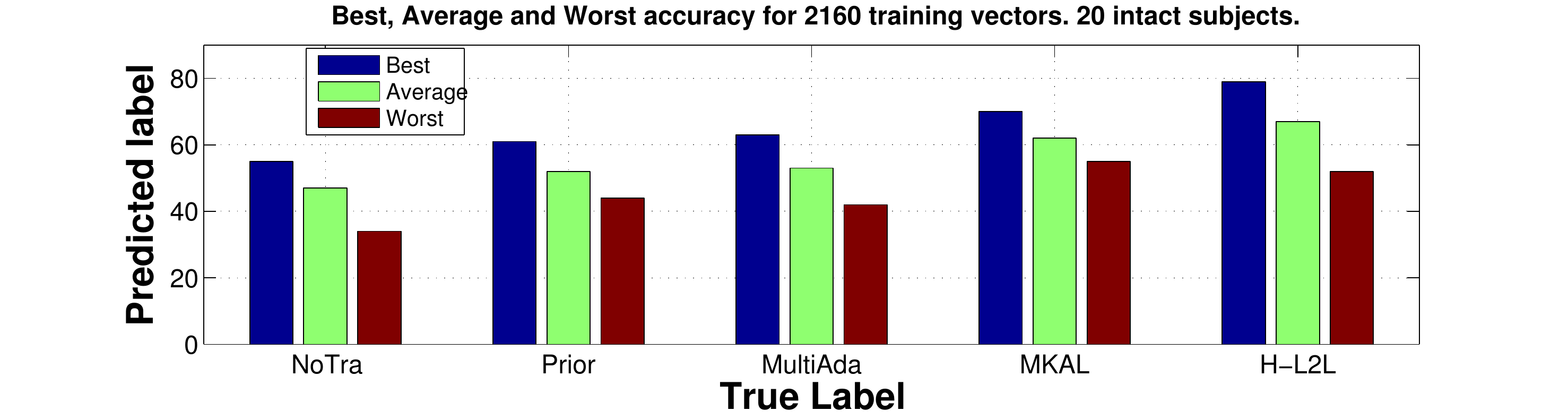}}
 \quad
 \subfigure
   {\includegraphics [scale = 0.44] {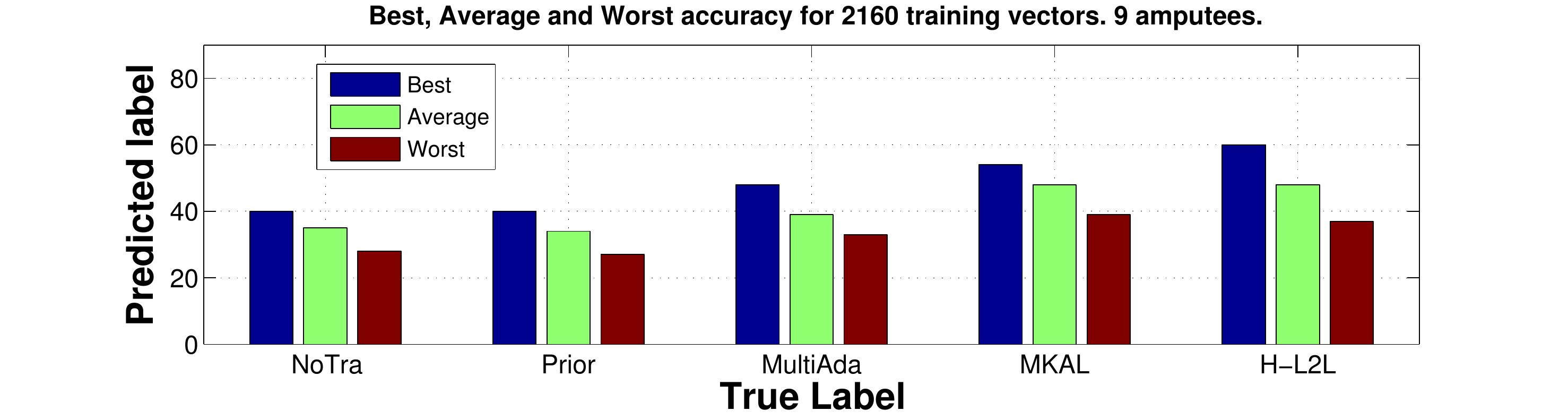}}
   \hspace{5mm}
 \subfigure
   {\includegraphics [scale = 0.44] {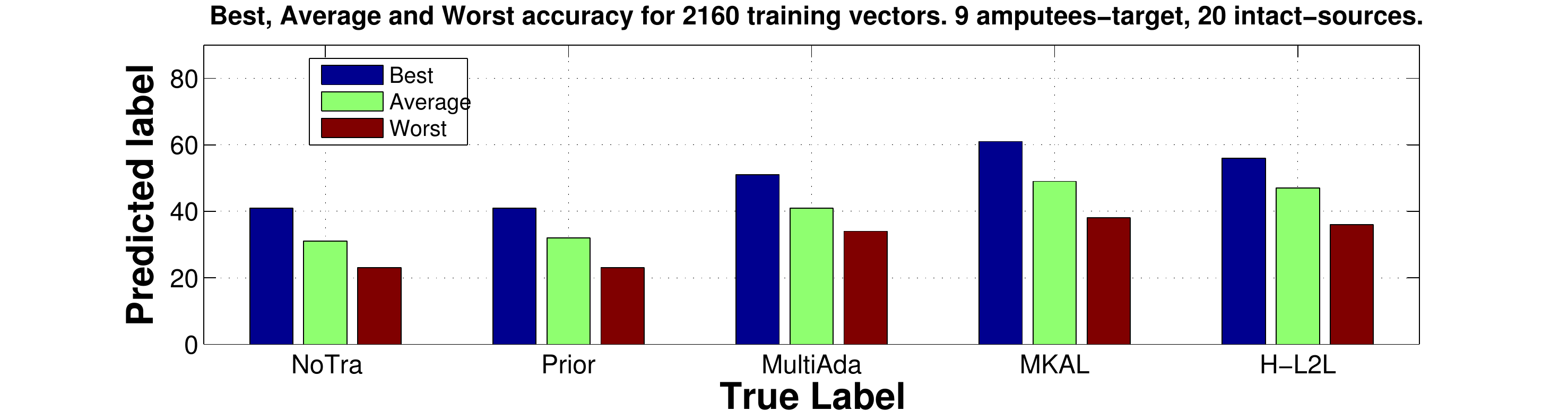}}
  \caption{Classification rate comparison.}\label{fig:ComparisonPerformance} 
\end{figure}

\paragraph{Comparison: Recognition Rate.} In this section we compare results obtained in different experiments.\\
For the comparison in performance of all the algorithms in the three experiments we report in Figure \ref{fig:ComparisonPerformance} the asymptotic classification rate (i.e. 2160 training vectors) obtained averaging over all the subjects, in the best and in the worst case.\\

The order in performance of the different algorithms is the same in the three experiments, thus the average behaviours are maintained. Only H-L2L presents some differences: it outperforms MKAL in the first case, instead in the second and third case the two algorithms reach the same final performance.\\
The performance achieved for intact target (first experiment) is higher than the performance obtained for amputated target (second and third experiment). It means that an intact is able to adapt his domain on domain of other intact subjects successfully (MKAL and H-L2L reach a performance of respectively 63 $\%$ and 67 $\%$ at 2160 training samples). The same performance is not reached by amputees, in fact MKAL and H-L2L achieve an asymptote of about 50 $\%$. It means that, with respect to intact, for an amputee is more difficult to extract useful informations from different domains.\\
The most interesting point concerns the comparison between the results from the second and the third experiment. The obtained performance appears unchanged in the two cases for all the algorithms. It means that an amputee learns in the same way from intact and amputated sources. It is a very important result, in fact as it is difficult to recruit amputees and to perform an experiment with them. Amputees are obviously outnumbered with respect to intact subjects and, for them, the proposed exercises are usually difficult and painful to perform.\\

\begin{figure} [H] 
\centering
\includegraphics [scale = 0.45] {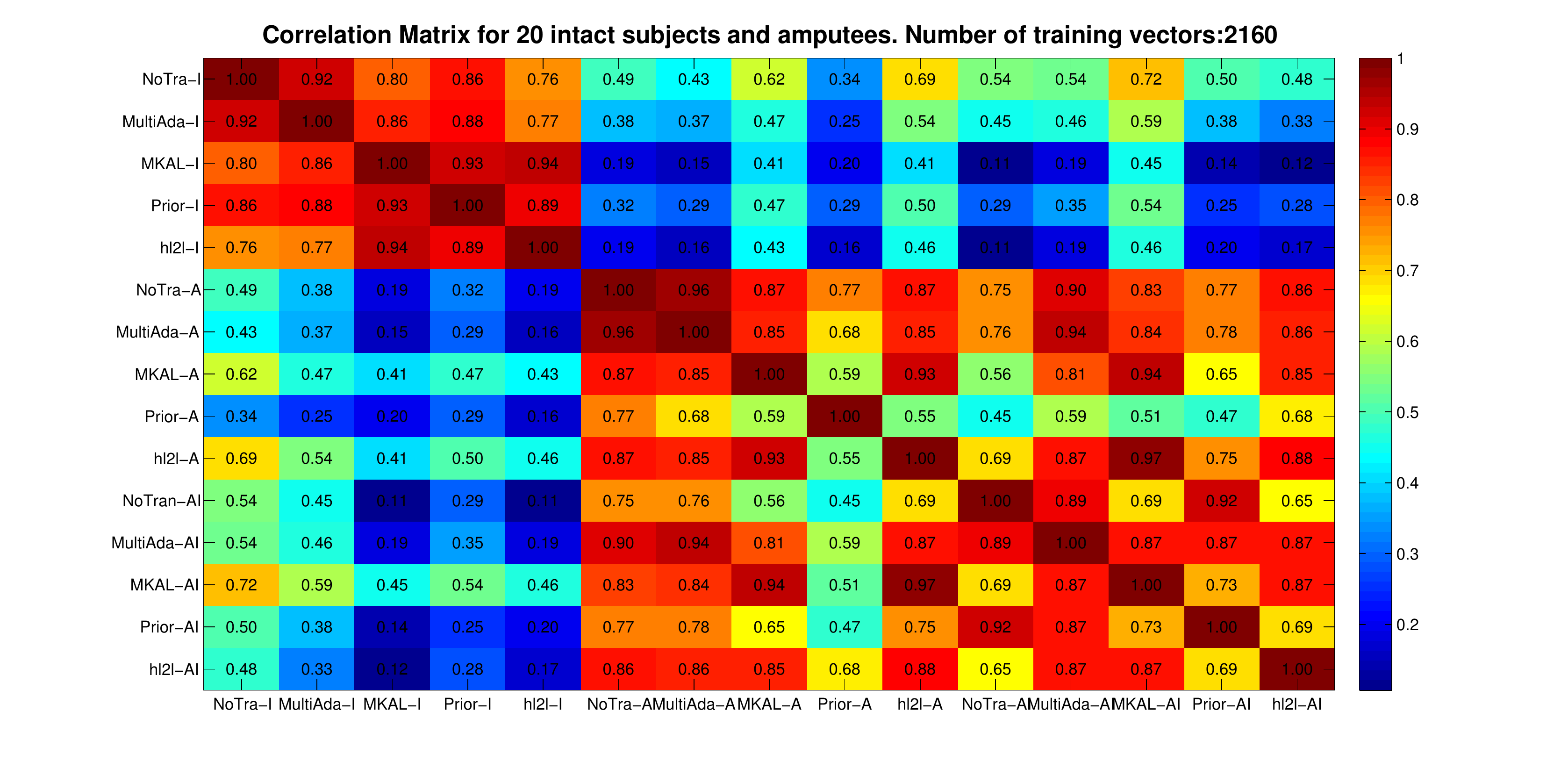}
\caption{Correlation matrix for each algorithm and experimental setting. I, A and AI stand respectively for: intact (i.e. first experiment), amputees (i.e. second experiment) and amputees-intact (i.e. third experiment).}
\label{fig:CorrMatrix_TrainVec2160}
\end{figure}

\paragraph{Comparison: Correlation Matrix.} We study now if there are movements learned with ease or difficulty by intact subjects and vice versa by amputees. In order to achieve this goal we consider the recognition percentage of each movement for a given number of training vectors. We normalize each recognition with respect to the maximum in order to compare results from different settings and algorithms without considering the absolute performance. For these values (one for each algorithm and experiment) the correlation matrix is calculated and reported in Figure \ref{fig:CorrMatrix_TrainVec2160}.\\

In previous matrix the warm colors and cool colors represent respectively high and low correlation between algorithms. As Figure \ref{fig:CorrMatrix_TrainVec2160} shows, we obtain a block matrix. The two diagonal blocks contain respectively the correlations between intact and amputees. The block outside the diagonal holds the correlation between intact and amputees.\\
The correlations between different methods of the same experiment is high. It means that there aren't movements learned simply with a method and hardly with another.\\ 
The correlation between second and third experiment is high, thus an amputee learns a movement with the same simplicity or difficulty from intact and amputated subjects. Instead the correlation between amputees and intact is low. It means that for amputees and intact simple and difficult movements are different.\\

\clearpage{\pagestyle{empty}\cleardoublepage}
\chapter*{Conclusion}
\addcontentsline{toc}{chapter}{Conclusions}%

In this work we have dealt with the control of non-invasive myoelectric prosthesis in order to overcome the most common problems that an amputee tackles: fatigue and pain due to the long learning process and the discouragement due to the mismatch between desired and performed movements. In particular we have investigated the variation of the learning curve when a form of prior knowledge is used and if the type of knowledge (from amputees or intact) is significant with respect to the final learning performance for an amputee.\\
Our approach has been computational. We worked with public data from database NinaPro to make the experiment reproducible and results statistically relevant (considering the large number of subjects in the database). The algorithms used for data representation, and to solve the movements classification problem represent, respectively, the state of the art in the field of features extraction and domain adaptation.\\
We have done three experiments in order to answer the previous questions. In the first we have considered intact that exploit the knowledge from other intact subjects. In the second we have studied amputated subjects that use other amputees as prior. In the last experiment we looked at amputees as target and intact subjects as their sources.\\

Our findings are very interesting and can contribute to improve this field of research. Our results show that the prior knowledge positively affects the trend of recognition curve. With respect to the case in which no kind of knowledge is transferred, with adaptive algorithms the asymptotic performance is reached before and with an higher value. The performance achieved without transfer considering the maximum number of training vectors (i.e. 2160) is typically reached by adaptive methods exploiting an order of magnitude less of training vectors. These results allow us to reduce the training time, speeding up the learning.\\
The last two experiments have been done in order to answer the question about previous knowledge for an amputee target: our findings show that there is not any change in the trend and in the asymptotic performance of each algorithms when amputees or intact subjects are used as source. This means that an amputee, in order to boost the control learning of her prosthetic hand, can exploit the knowledge of intact subjects that are simpler to recruit than amputees. Furthermore, for an intact user to complete the experiment requires less time and effort with respect to an amputee, thus the source models are created with ease.\\

\paragraph{Future work.}
The asymptotic performance reached by amputees is not very high (50 $\%$ at the best). A possible direction for future works could be to try to improve this result focusing on deep learning theory that has had very interesting results in the field of machine learning.\\

From analysis of confusion matrices we have seen that, given different algorithms and number of training vectors, a posture (i.e. a class) is, in most cases, misclassified with the same wrong classes. In order to solve the problem a possible direction for future research could be to insert a sort of classes selector that, a priori, cuts out on classification for some postures.\\ 

Lastly, we have made in this work the hypothesis that all source subjects had performed the same exercises and are therefore able to perform the same postures, both among them and with respect to the target subject. This in general will not be the case. Different people might require different functionalities from their prostheses. As of today, there are no domain adaptation algorithms able to deal with this scenario. Future work will focus on these research threads.\\

\clearpage{\pagestyle{empty}\cleardoublepage}
\part*{Allegati} 
\appendix
\chapter{Histogram analysis}

This analysis is done in order to evaluate if there is an adaptive method that recognizes a class better than others and if this statistic changes increasing the number of training vectors. For each real class we consider the four classes with highest predictions, in order to evaluate the differences in misclassification. Generally the fifth predicted class has a recognition lower than 5 $\%$, for this reason it and the following classes are not taken into account.\\
In the following we analyse the results obtained for each experiment.\\
We can find the complete results in\\ \url{https://sites.google.com/site/noninvasiveprosthetichand/}.\\

\section{Histogram: first experiment}\label{sec:HistI}
\begin{figure} [H]
\centering
\subfigure
   {\includegraphics[scale=0.43]{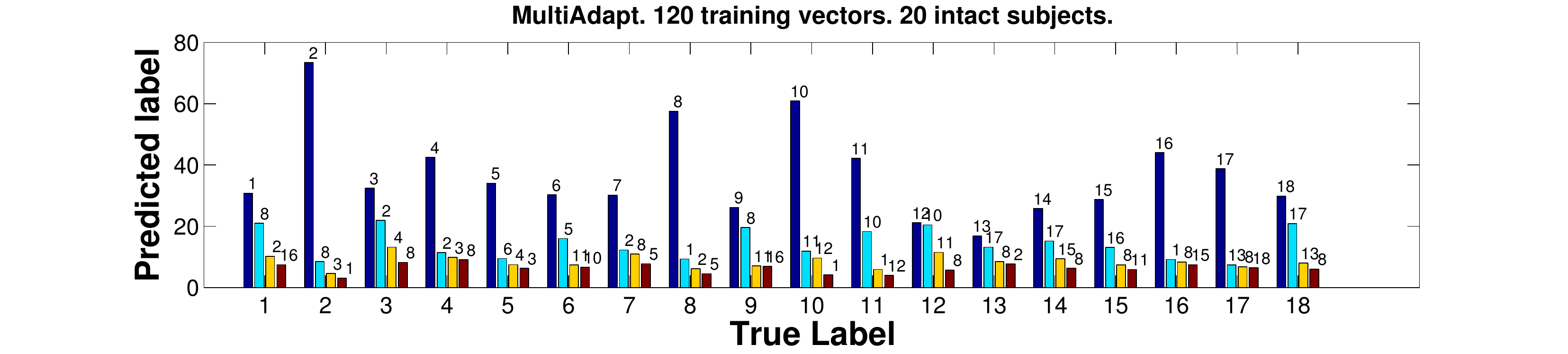}}
 \hspace{5mm}
 \subfigure
   {\includegraphics[scale=0.43]{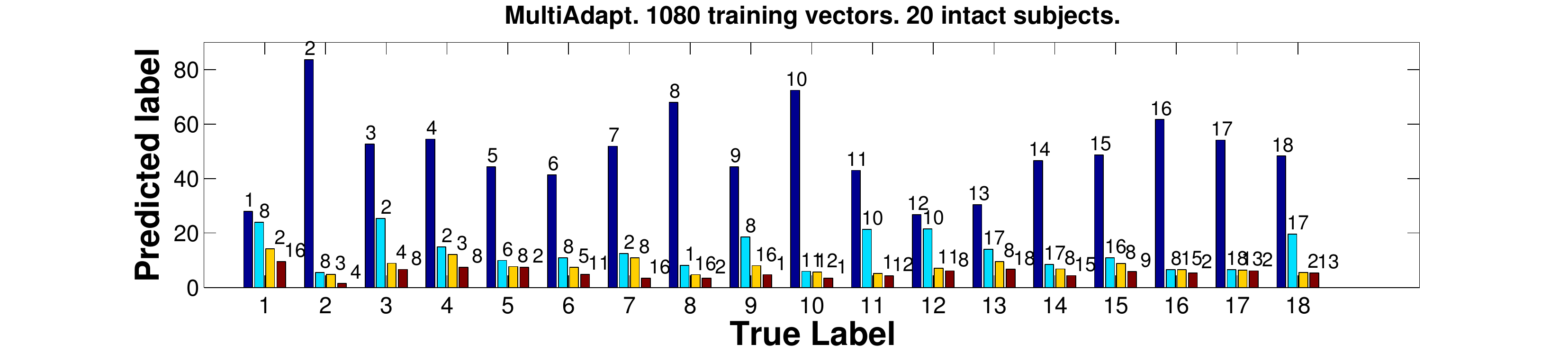}}
   \hspace{5mm}
 \subfigure
   {\includegraphics[scale=0.43]{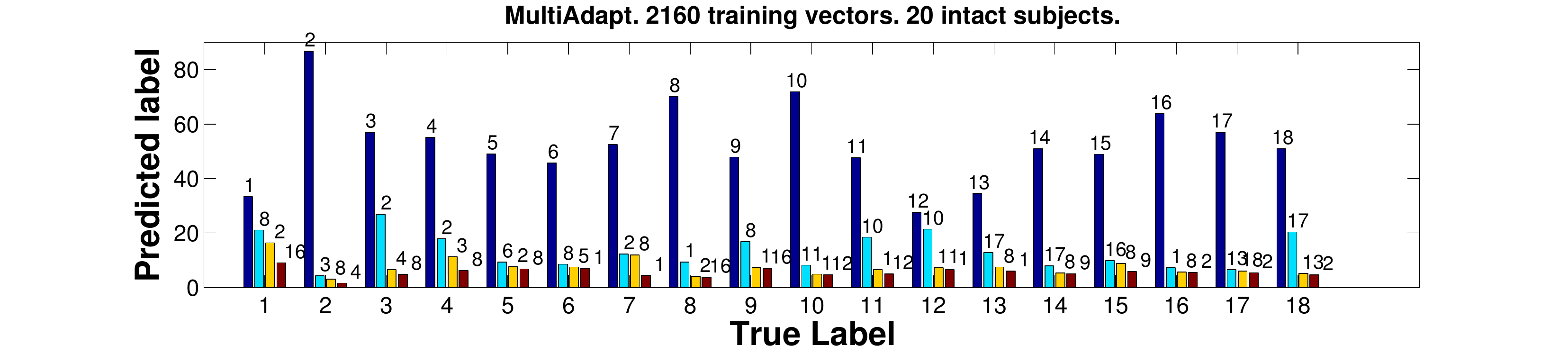}}
  \caption{The first four predicted classes for each true class for Multi Adapt method 120, 1080 and 2160 training vectors.}\label{fig:MAhist} 
\end{figure}

\begin{figure} [H]
\centering
\subfigure
   {\includegraphics[scale=0.43]{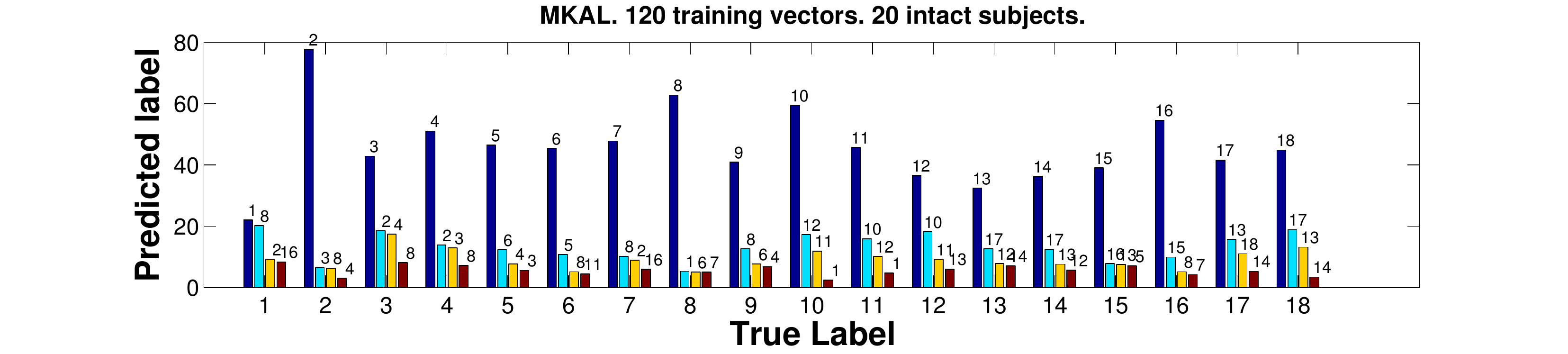}}
 \hspace{5mm}
 \subfigure
   {\includegraphics[scale=0.43]{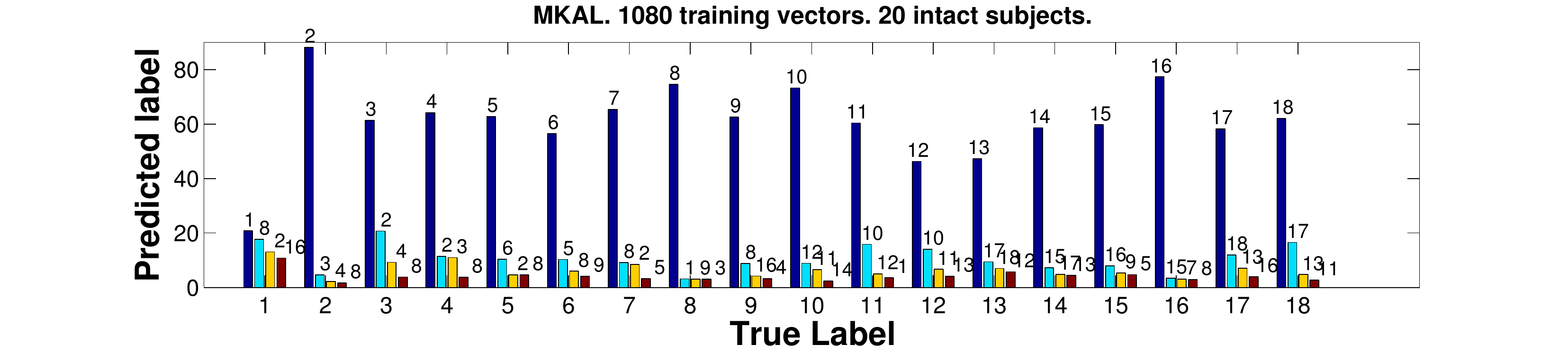}}
   \hspace{5mm}
 \subfigure
   {\includegraphics[scale=0.43]{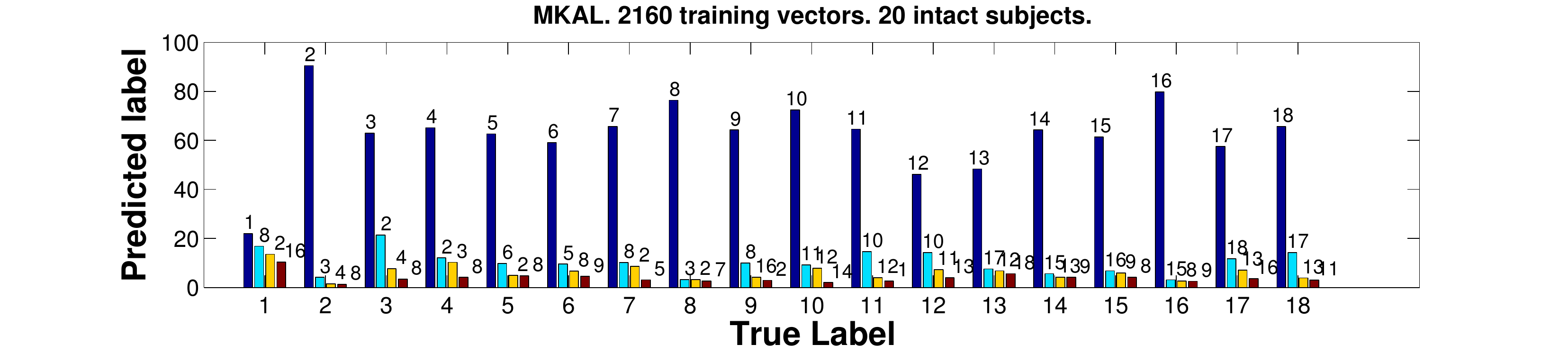}}
  \caption{The first four predicted classes for each true class for MKAL method 120, 1080 and 2160 training vectors.}\label{fig:MKALhist} 
\end{figure}

\begin{figure} [H]
\centering
\subfigure
   {\includegraphics[scale=0.43]{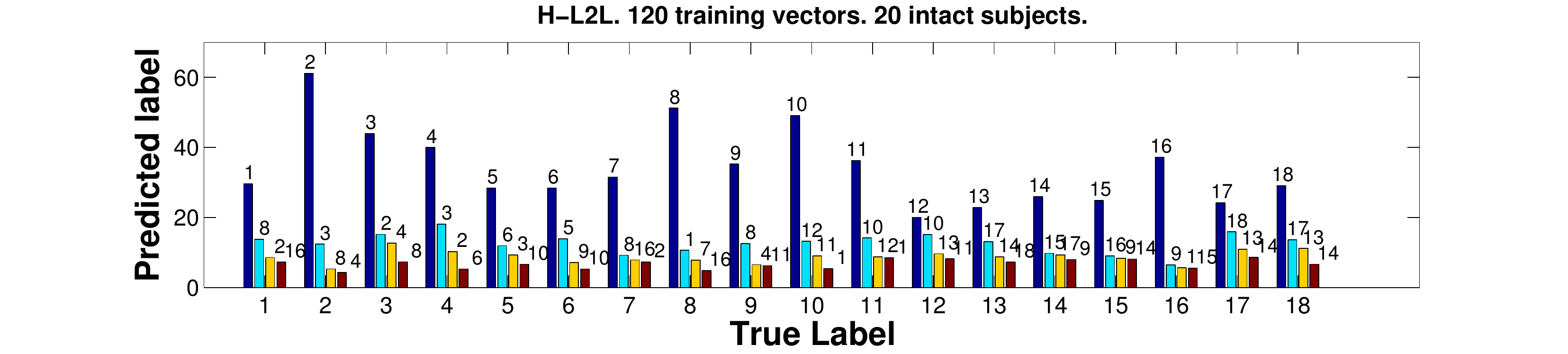}}
 \hspace{5mm}
 \subfigure
   {\includegraphics[scale=0.43]{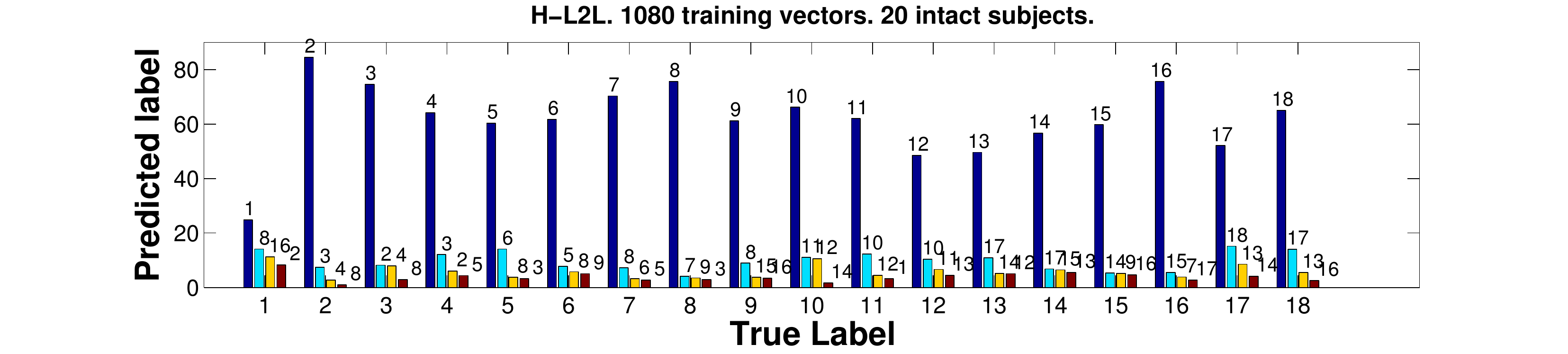}}
   \hspace{5mm}
 \subfigure
   {\includegraphics[scale=0.43]{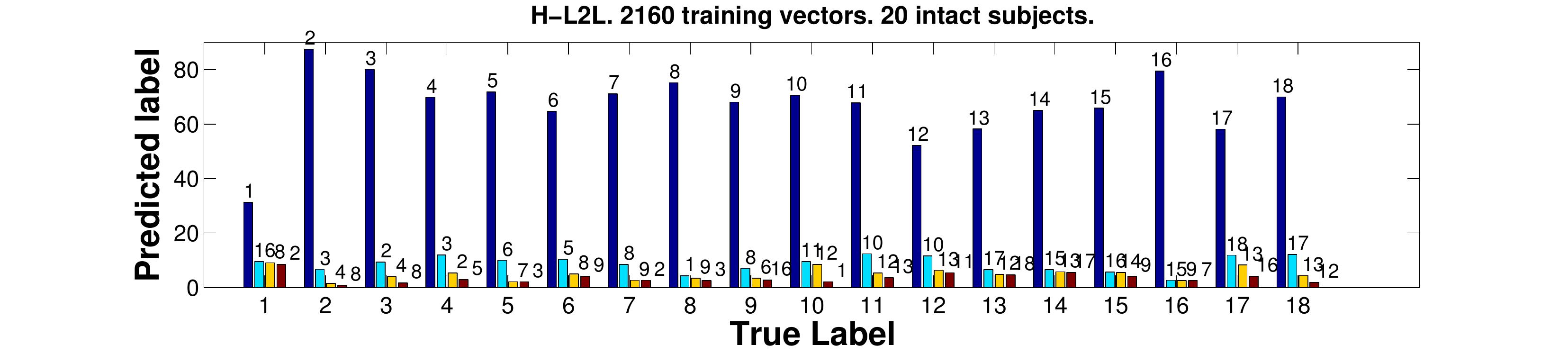}}
  \caption{The first four predicted classes for each true class for H-L2L method 120, 1080 and 2160 training vectors.}\label{fig:H-L2Lhist} 
\end{figure}

We report the histogram of the first four classes predicted for each true label. Methods involved in this analysis are Multi Adapt (Figure \ref{fig:MAhist}), MKAL (Figure \ref{fig:MKALhist}) and H-L2L (Figure \ref{fig:H-L2Lhist}) for 120 (i.e. initial step), 1080 (i.e. middle step) and 2160 (i.e. final step) training vectors.

We study the change in misclassification of each class when the number of training vectors increases. For each algorithm we calculate the percentage of classes that have at least the $\frac{3}{4}$ of predicted labels equal, changing the number of training vectors. The results are reported in Table \ref{tab:PercII}.

\begin{table}[H]
	
	\resizebox{12.8cm}{!}	{
	
	\begin{tabular} {|| p{7cm} | p{5cm} ||}
	
	\hline
	\hline
	\textbf{Multi Adapt}  & \textbf{Percentage} \\ 		\hline
	\hline
	\textbf{120-1080} & 94 $\%$  ($17/18$) \\    \hline
	\textbf{1080-2160} &   100 $\%$ ($18/18$) \\  \hline
	\hline
	\hline
	\textbf{MKAL}  & \textbf{Percentage} \\ 		\hline
	\hline
	\textbf{120-1080} & 89 $\%$  ($16/18$) \\    \hline
	\textbf{1080-2160} &  94  $\%$ ($17/18$) \\  \hline
	\hline
	\hline
	\textbf{H-L2L}  & \textbf{Percentage} \\ 		\hline
	\hline
	\textbf{120-1080} & 78 $\%$  ($14/18$) \\    \hline
	\textbf{1080-2160} &  94  $\%$ ($17/18$) \\  \hline
	\hline
	\end{tabular} \\
	}
	\caption{Percentage of similarity in the prediction of classes. We report the percentage of cases in which at least 3 prediction on the 4 considered in previous histograms are the same.}
	\label{tab:PercII}
\end{table}

For adaptive methods, increasing the number of training vectors, a posture is misclassified always with the same wrong classes.\\
The same analysis can be done considering different algorithms with the same number of training vectors. Results of percentage are reported in Table \ref{tab:PercII}.

\begin{table}[H]
	
	\resizebox{13cm}{!}	{
	
	\begin{tabular} {|| p{7cm} | p{5cm} ||}
	
	\hline
	\hline
	\textbf{120 training vectors}  & \textbf{Percentage} \\ 		\hline
	\hline
	\textbf{MultiAdapt - MKAL} & 72 $\%$  ($13/18$) \\    \hline
	\textbf{MKAL - H-L2L} &  78  $\%$ ($14/18$) \\  \hline
	\textbf{Multi Adapt - H-L2L} &  83  $\%$ ($15/18$) \\  \hline
	\hline
	\hline
	\textbf{1080 training vectors}  & \textbf{Percentage} \\ \hline
	\hline
	\textbf{MultiAdapt - MKAL} & 94 $\%$  ($17/18$) \\    \hline
	\textbf{MKAL - H-L2L} & 100 $\%$ ($18/18$) \\  \hline
	\textbf{Multi Adapt - H-L2L} &  78  $\%$ ($14/18$) \\  \hline
	\hline
	\hline
	\textbf{2160 training vectors}  & \textbf{Percentage} \\ 		\hline
	\hline
	\textbf{MultiAdapt - MKAL} & 78 $\%$  ($14/18$) \\    \hline
	\textbf{MKAL - H-L2L} &  83  $\%$ ($15/18$) \\  \hline
	\textbf{Multi Adapt - H-L2L} &  72  $\%$ ($13/18$) \\  \hline
	\hline
	\end{tabular} \\
	}
	\caption{Percentage of similarity in the prediction of classes. We report the percentage of cases in which at least 3 prediction on the 4 considered in previous histograms are the same.}
	\label{tab:PercII}
\end{table}

From this analysis we can conclude that, given a class, different methods, generally, misclassify it with the same wrong classes. Thus the misclassification is due to the similarity between the classes and it is independent from the algorithm.\\
We must remember that the problem that we tackle is the recognition and classification of 8 fingers movements and 9 wrist postures (see Figure \ref{fig:AllMovements}). These movements are very similar, thus it is reasonable to expect an high degree of confusion between different postures.\\

\section{Histogram: second experiment}\label{sec:HistA}
\begin{figure} [H]
\centering
\subfigure
   {\includegraphics[scale=0.43]{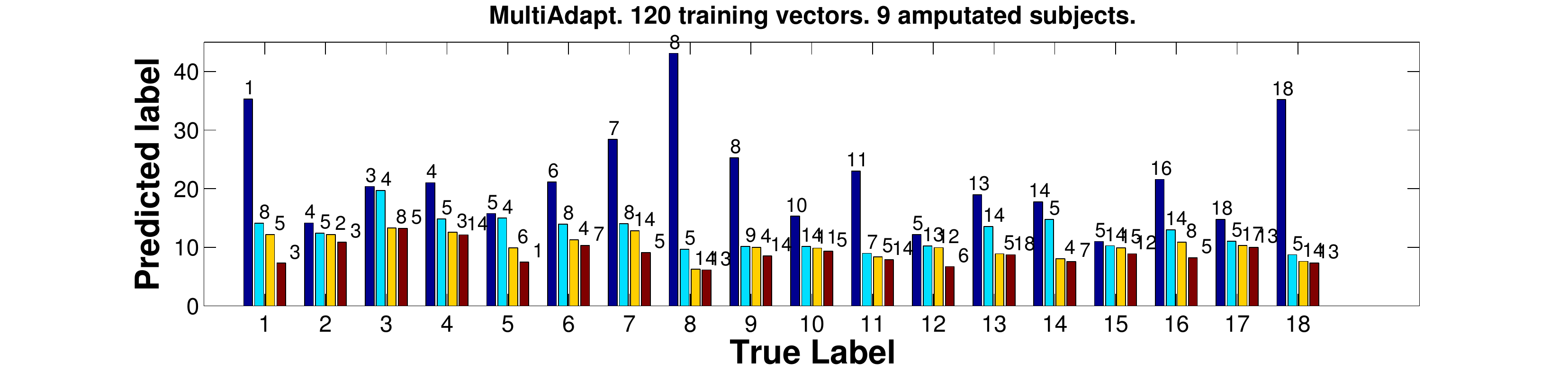}}
 \hspace{5mm}
 \subfigure
   {\includegraphics[scale=0.43]{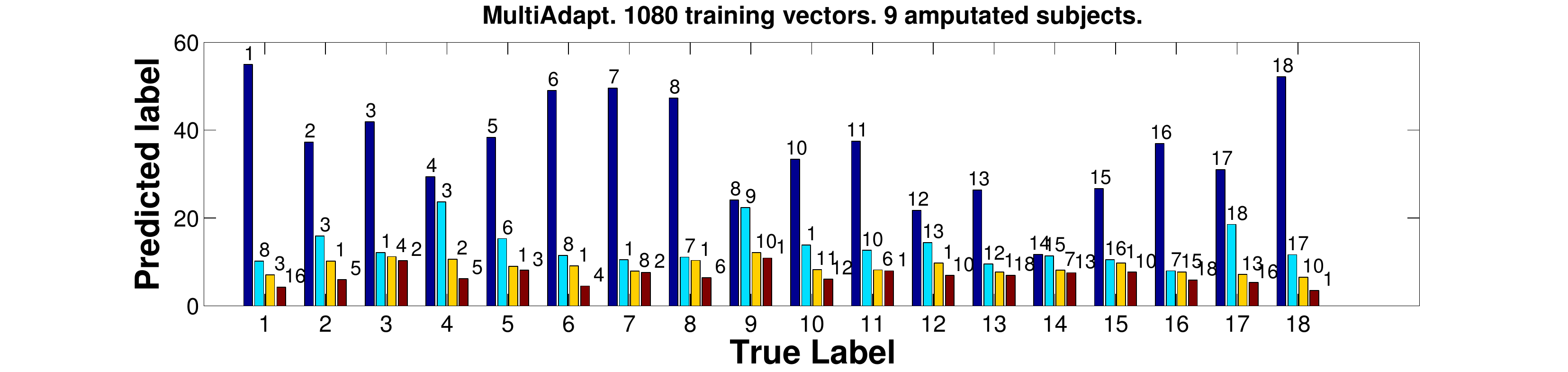}}
   \hspace{5mm}
 \subfigure
   {\includegraphics[scale=0.43]{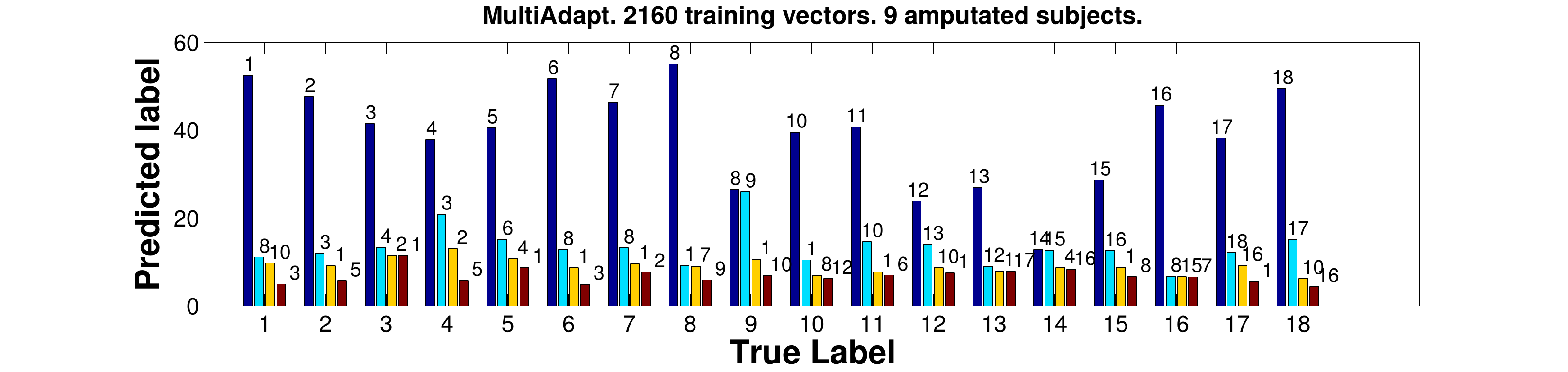}}
  \caption{The first four predicted classes for each true class for Multi Adapt method 120, 1080 and 2160 training vectors.}\label{fig:MAhistA} 
\end{figure}

\begin{figure} [H]
\centering
\subfigure
   {\includegraphics[scale=0.43]{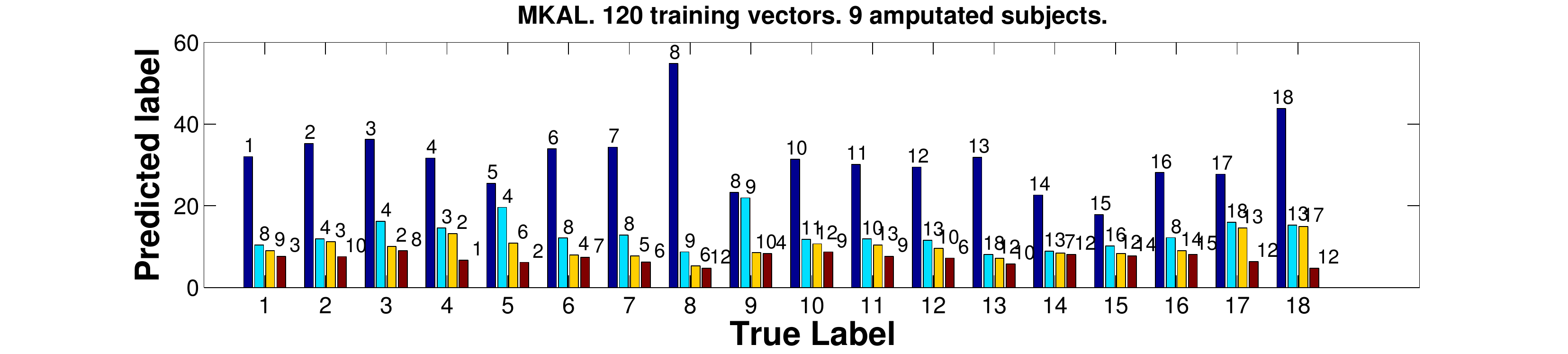}}
 \hspace{5mm}
 \subfigure
   {\includegraphics[scale=0.43]{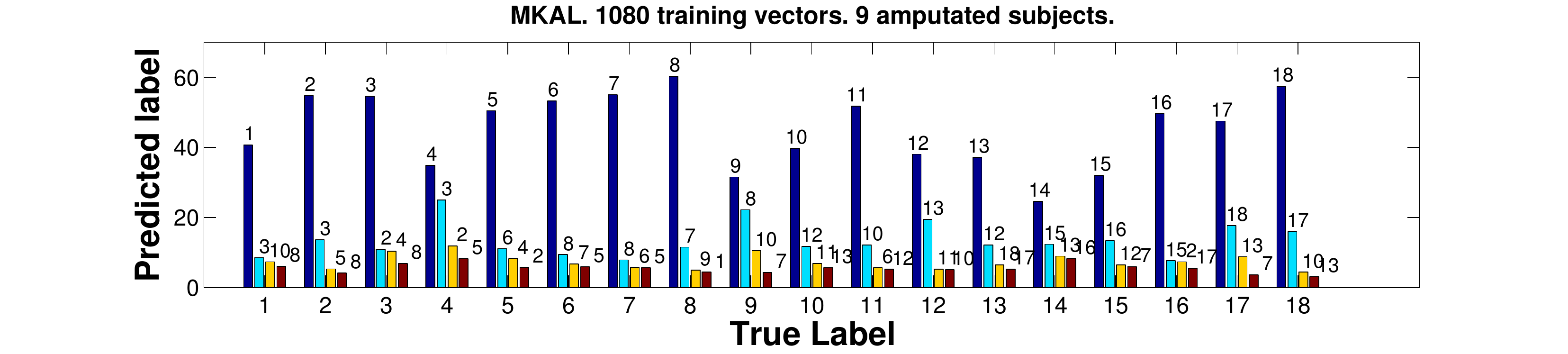}}
   \hspace{5mm}
 \subfigure
   {\includegraphics[scale=0.43]{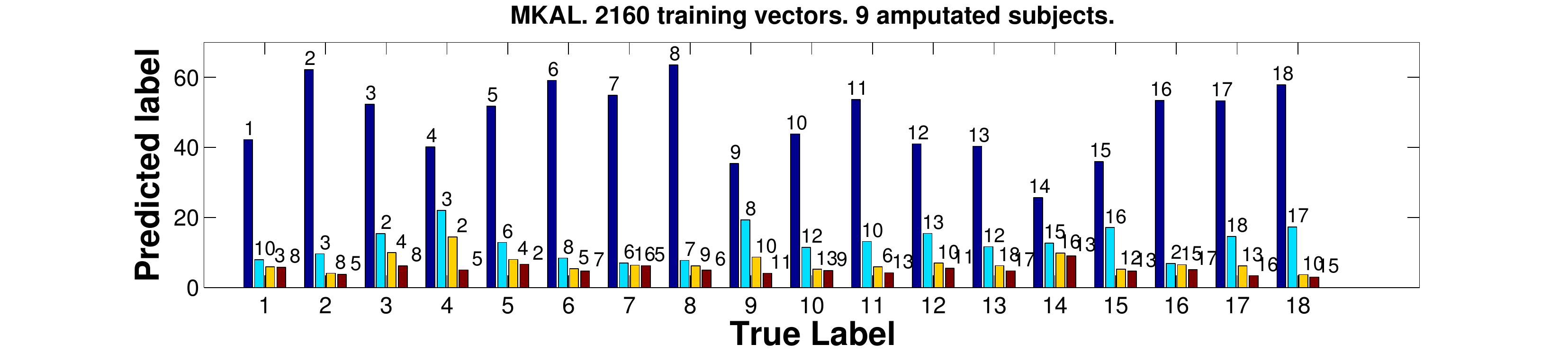}}
  \caption{The first four predicted classes for each true class for MKAL method 120, 1080 and 2160 training vectors.}\label{fig:MKALhistA} 
\end{figure}

\begin{figure} [H]
\centering
\subfigure
   {\includegraphics[scale=0.43]{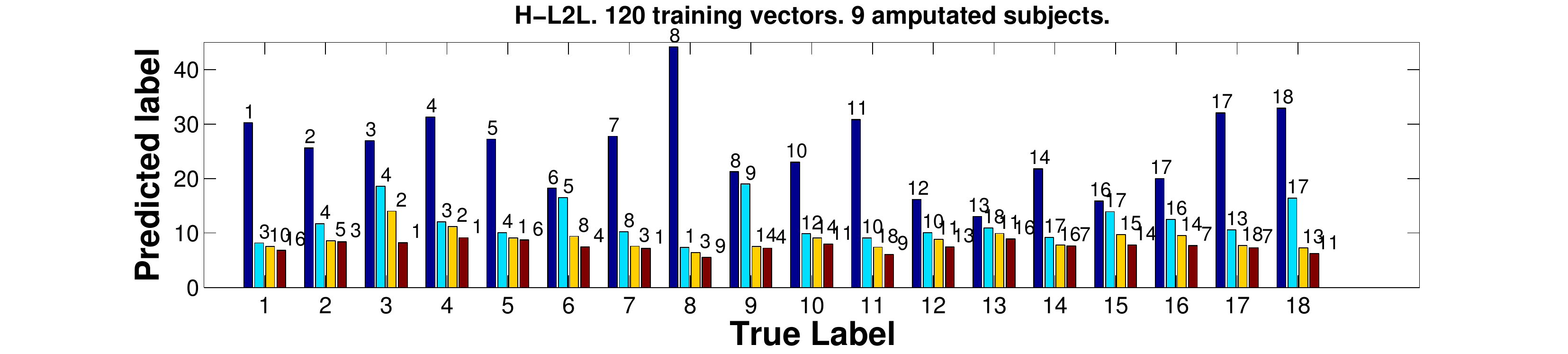}}
 \hspace{5mm}
 \subfigure
   {\includegraphics[scale=0.43]{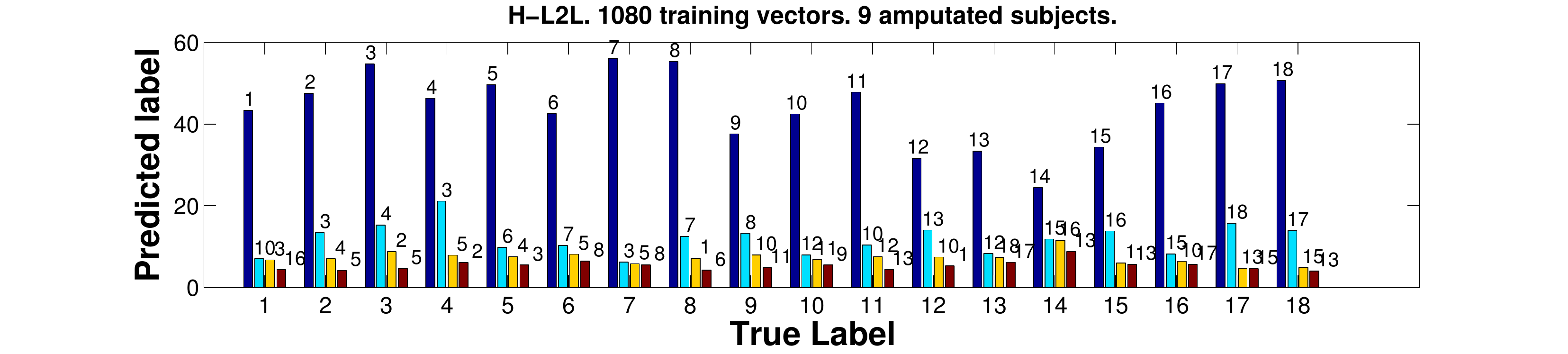}}
   \hspace{5mm}
 \subfigure
   {\includegraphics[scale=0.43]{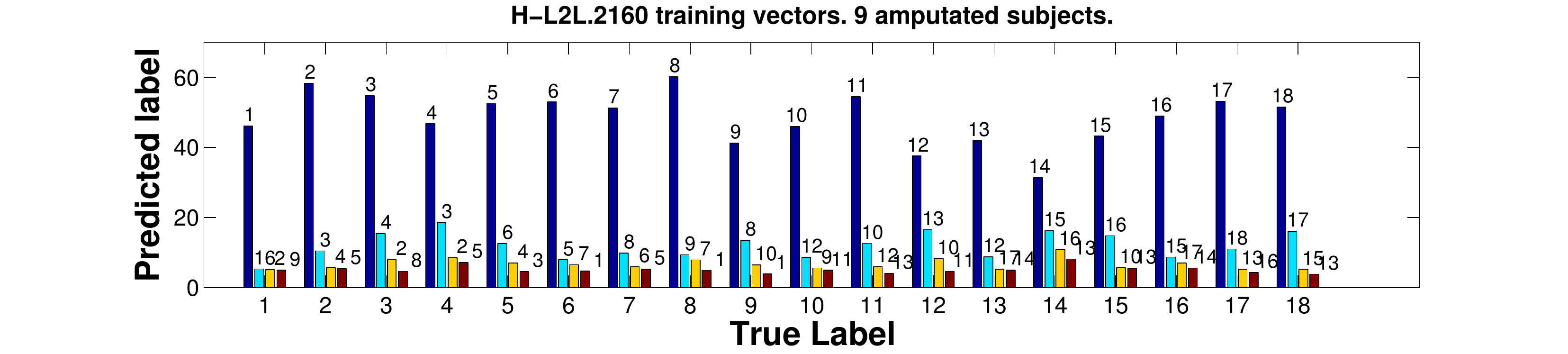}}
  \caption{The first four predicted classes for each true class for H-L2L method 120, 1080 and 2160 training vectors.}\label{fig:H-L2LhistA} 
\end{figure}

We report the histogram of the first four classes predicted for each true label. Methods involved in this analysis are Multi Adapt (Figure \ref{fig:MAhistA}), MKAL (Figure \ref{fig:MKALhistA}) and H-L2L (Figure \ref{fig:H-L2LhistA}) for 120 (i.e. initial step), 1080 (i.e. middle step) and 2160 (i.e. final step) training vectors.\\

We study the change in misclassification of each class when the number of training vectors increases. For each algorithm we calculate the percentage of classes that have at least the $\frac{3}{4}$ of predicted labels equal, changing the number of training vectors. The results are reported in Table \ref{tab:PercAA}.

\begin{table}[H]
	
	\resizebox{12.8cm}{!}	{
	
	\begin{tabular} {|| p{7cm} | p{5cm} ||}
	
	\hline
	\hline
	\textbf{Multi Adapt}  & \textbf{Percentage} \\ 		\hline
	\hline
	\textbf{120-1080} & 33 $\%$  ($6/18$) \\    \hline
	\textbf{1080-2160} &  94  $\%$ ($17/18$) \\  \hline
	\hline
	\hline
	\textbf{MKAL}  & \textbf{Percentage} \\ 		\hline
	\hline
	\textbf{120-1080} & 72 $\%$  ($13/18$) \\    \hline
	\textbf{1080-2160} &   100 $\%$ ($18/18$) \\  \hline
	\hline
	\hline
	\textbf{H-L2L}  & \textbf{Percentage} \\ 		\hline
	\hline
	\textbf{120-1080} & 61 $\%$  ($11/18$) \\    \hline
	\textbf{1080-2160} &  94  $\%$ ($17/18$) \\  \hline
	\hline
	\end{tabular} \\
	}
	\caption{Percentage of similarity in the prediction of classes. We report the percentage of cases in which at least 3 prediction on the 4 considered in previous histograms are the same.}
	\label{tab:PercAA}
\end{table}

For adaptive methods, increasing the number of training vectors, a posture is misclassified always with the same wrong classes. Only Multi Adapt shows a low percentage passing from 120 to 1080 training vectors. Probably this model is not very reliable at first step, in fact from first image of Figure \ref{fig:MAhistA} we can see that the bins with highest probability have not the label equal to the right one.\\
The same analysis can be done considering different algorithms with the same number of training vectors. Results of percentage are reported in Table \ref{tab:PercAA}.

\begin{table}[H]
	
	\resizebox{13cm}{!}	{
	
	\begin{tabular} {|| p{7cm} | p{5cm} ||}
	
	\hline
	\hline
	\textbf{120 training vectors}  & \textbf{Percentage} \\ 		\hline
	\hline
	\textbf{MultiAdapt - MKAL} & 61 $\%$  ($11/18$) \\    \hline
	\textbf{MKAL - H-L2L} &  67  $\%$ ($12/18$) \\  \hline
	\textbf{Multi Adapt - H-L2L} &  33  $\%$ ($6/18$) \\  \hline
	\hline
	\hline
	\textbf{1080 training vectors}  & \textbf{Percentage} \\ \hline
	\hline
	\textbf{MultiAdapt - MKAL} & 72 $\%$  ($13/18$) \\    \hline
	\textbf{MKAL - H-L2L} & 94 $\%$ ($17/18$) \\  \hline
	\textbf{Multi Adapt - H-L2L} & 72 $\%$ ($13/18$) \\  \hline
	\hline
	\hline
	\textbf{2160 training vectors}  & \textbf{Percentage} \\ 		\hline
	\hline
	\textbf{MultiAdapt - MKAL} &  72 $\%$  ($13/18$) \\    \hline
	\textbf{MKAL - H-L2L} &  94  $\%$ ($17/18$) \\  \hline
	\textbf{Multi Adapt - H-L2L} &  56  $\%$ ($10/18$) \\  \hline
	\hline
	\end{tabular} \\
	}
	\caption{Percentage of similarity in the prediction of classes. We report the percentage of cases in which at least 3 prediction on the 4 considered in previous histograms are the same.}
	\label{tab:PercAA}
\end{table}

From values of previous Table we conclude that, given a class, different methods, generally, misclassify it with the same wrong classes.\\ 
We remember that the problem that we tackle is the recognition and classification of 8 fingers movements and 9 wrist postures (see Figure \ref{fig:AllMovements}). These movements are very similar, thus it is reasonable to expect an high degree of confusion between different postures.\\

\section{Histogram: third experiment}\label{sec:HistAI}
\begin{figure} [H]
\centering
\subfigure
   {\includegraphics[scale=0.43]{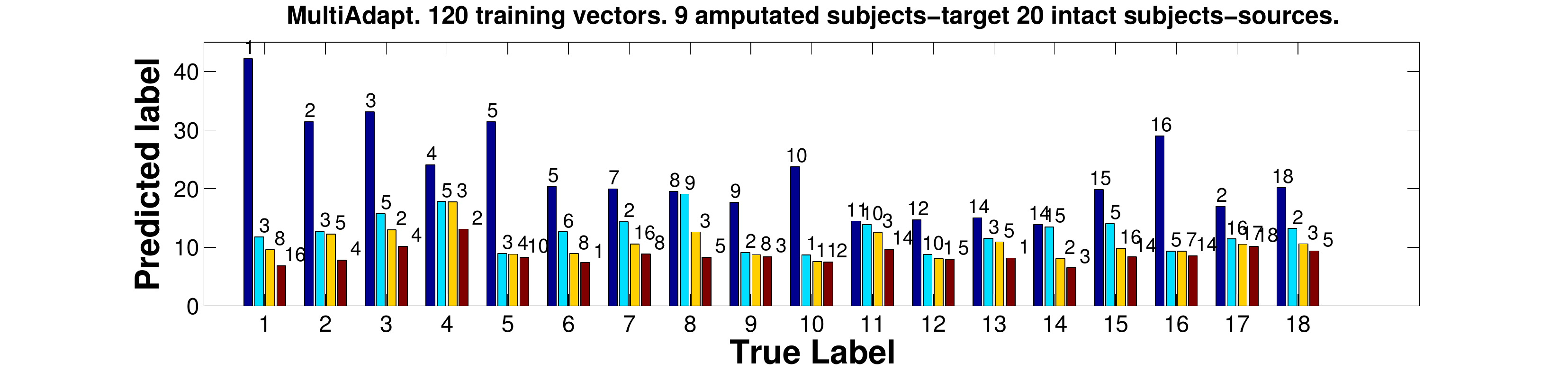}}
 \hspace{5mm}
 \subfigure
   {\includegraphics[scale=0.43]{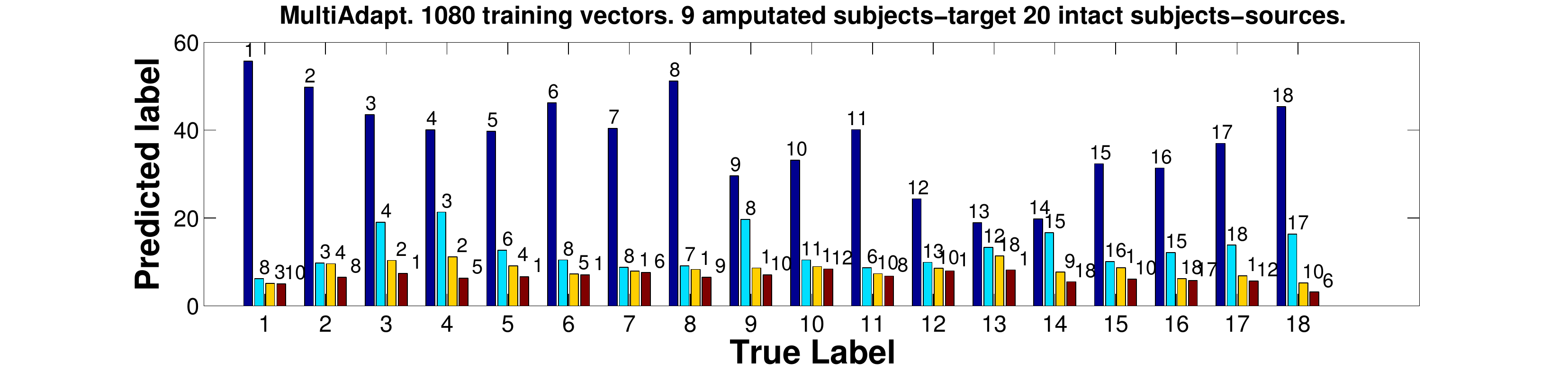}}
   \hspace{5mm}
 \subfigure
   {\includegraphics[scale=0.43]{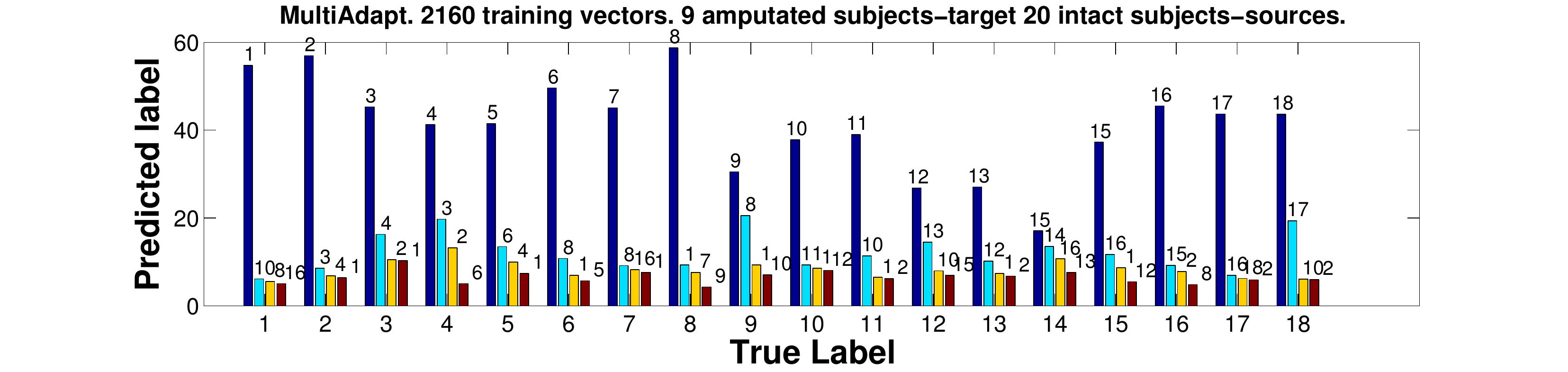}}
  \caption{The first four predicted classes for each true class for Multi Adapt method 120, 1080 and 2160 training vectors.}\label{fig:MAhistAI} 
\end{figure}

\begin{figure} [H]
\centering
\subfigure
   {\includegraphics[scale=0.43]{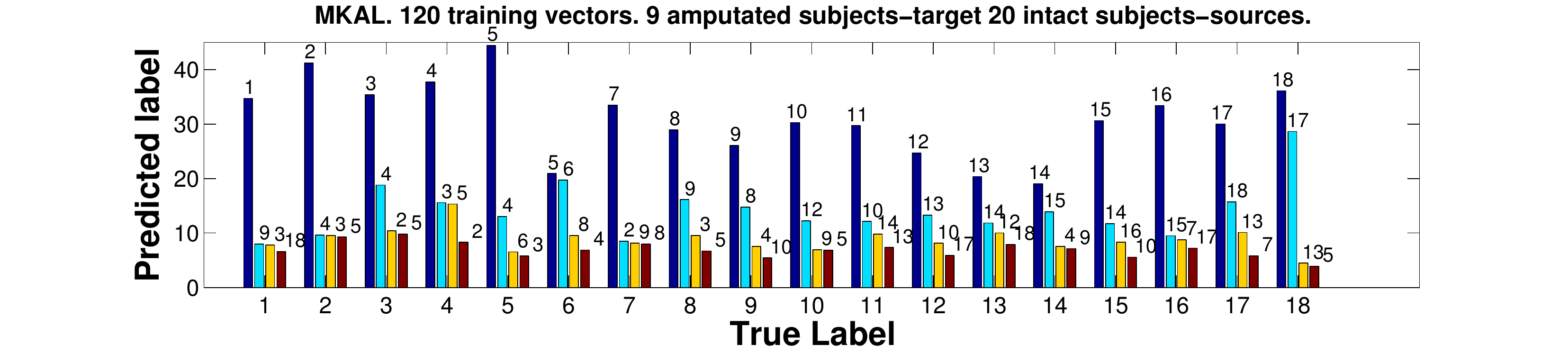}}
 \hspace{5mm}
 \subfigure
   {\includegraphics[scale=0.43]{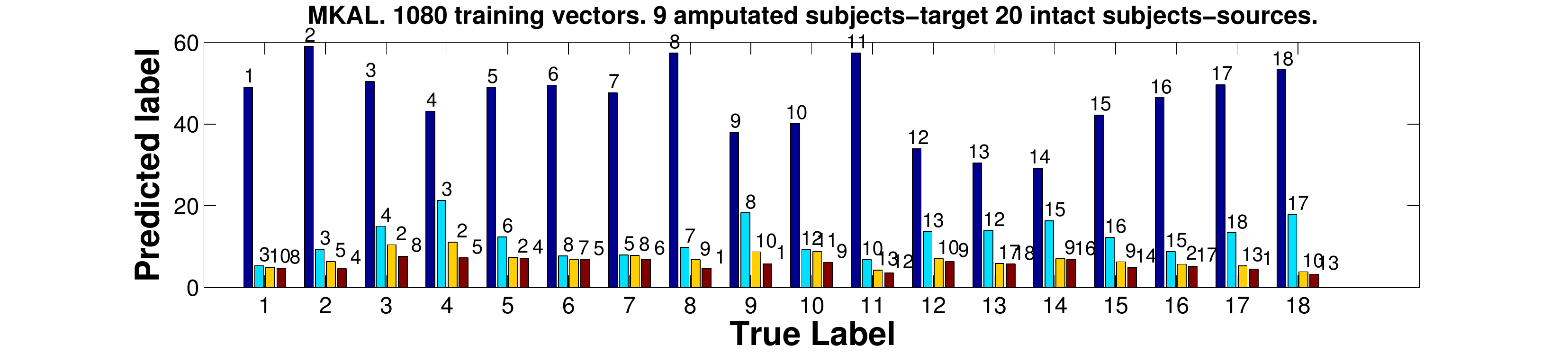}}
   \hspace{5mm}
 \subfigure
   {\includegraphics[scale=0.43]{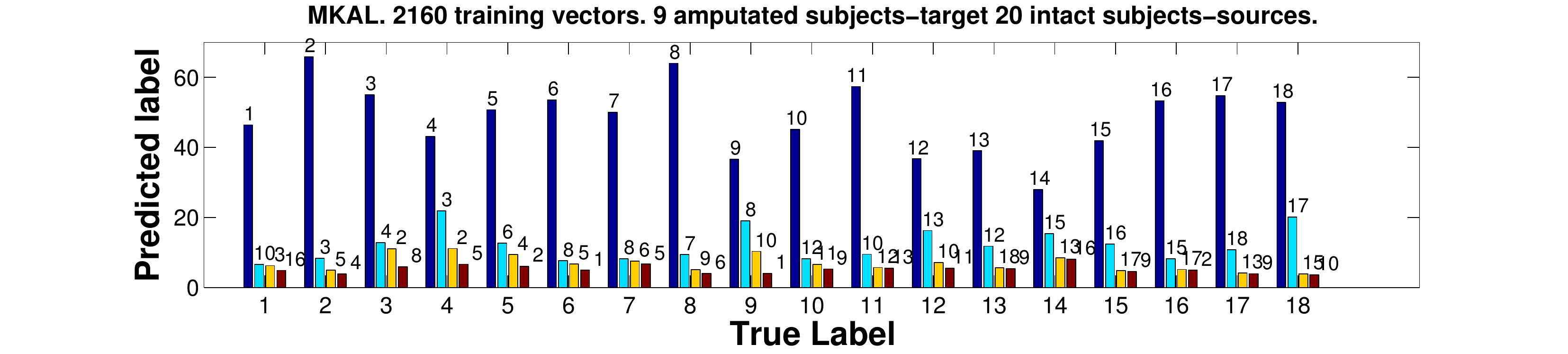}}
  \caption{The first four predicted classes for each true class for MKAL method 120, 1080 and 2160 training vectors.}\label{fig:MKALhistAI} 
\end{figure}

\begin{figure} [H]
\centering
\subfigure
   {\includegraphics[scale=0.43]{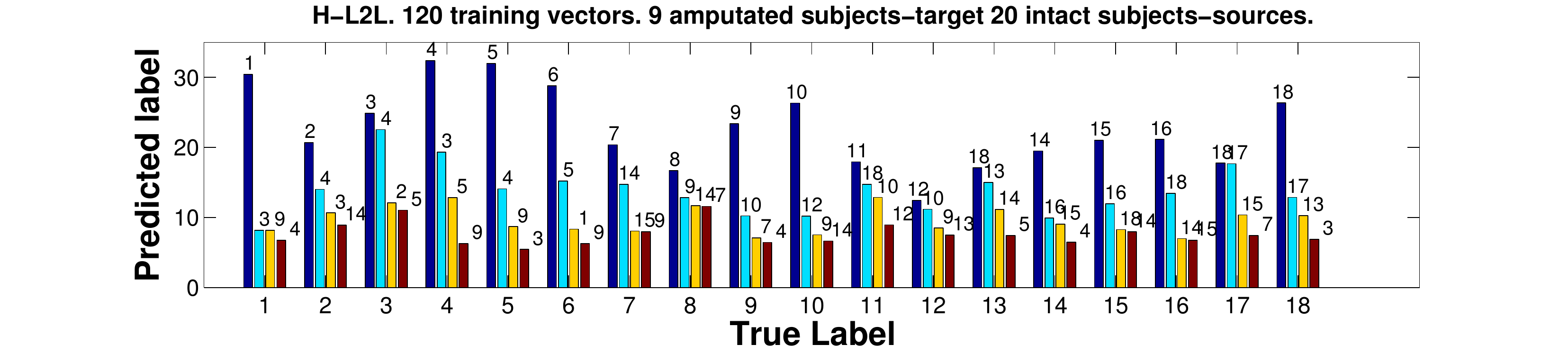}}
 \hspace{5mm}
 \subfigure
   {\includegraphics[scale=0.43]{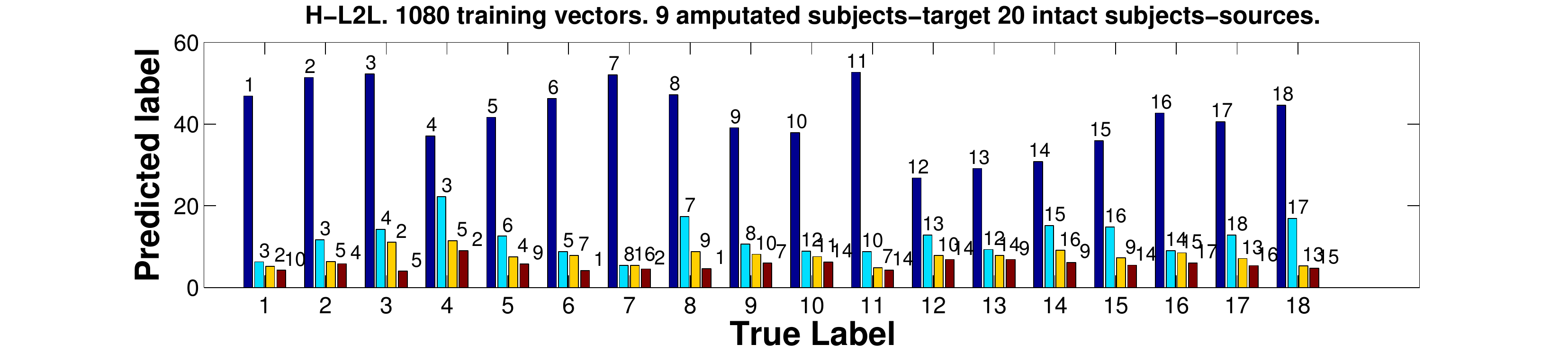}}
   \hspace{5mm}
 \subfigure
   {\includegraphics[scale=0.43]{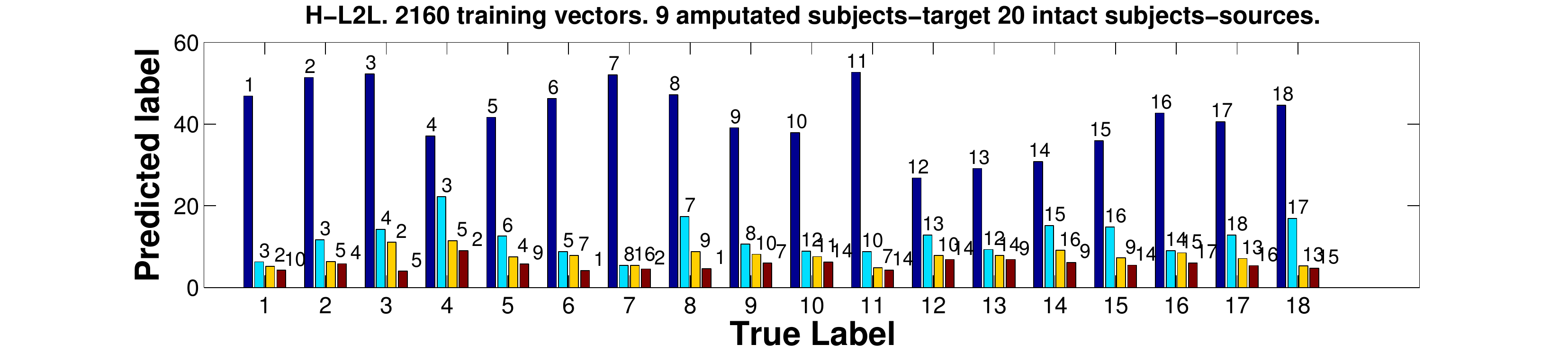}}
  \caption{The first four predicted classes for each true class for H-L2L method 120, 1080 and 2160 training vectors.}\label{fig:H-L2LhistAI} 
\end{figure}

We report the histogram of the first four classes predicted for each true label. Methods involved in this analysis are Multi Adapt (Figure \ref{fig:MAhistAI}), MKAL (Figure \ref{fig:MKALhistAI}) and H-L2L (Figure \ref{fig:H-L2LhistAI}) for 120 (i.e. initial step), 1080 (i.e. middle step) and 2160 (i.e. final step) training vectors.\\

We study the change in misclassification of each class when the number of training vectors increases. As done in previous sections, for each algorithm we calculate the percentage of classes that have at least the $\frac{3}{4}$ of predicted labels equal, changing the number of training vectors. The results are reported in Table \ref{tab:PercAI}.

\begin{table}[H]
	
	\resizebox{12.8cm}{!}	{
	
	\begin{tabular} {|| p{7cm} | p{5cm} ||}
	
	\hline
	\hline
	\textbf{Multi Adapt}  & \textbf{Percentage} \\ 		\hline
	\hline
	\textbf{120-1080} & 39 $\%$  ($7/18$) \\    \hline
	\textbf{1080-2160} &  78  $\%$ ($14/18$) \\  \hline
	\hline
	\hline
	\textbf{MKAL}  & \textbf{Percentage} \\ 		\hline
	\hline
	\textbf{120-1080} & 83 $\%$  ($15/18$) \\    \hline
	\textbf{1080-2160} &  100  $\%$ ($18/18$) \\  \hline
	\hline
	\hline
	\textbf{H-L2L}  & \textbf{Percentage} \\ 		\hline
	\hline
	\textbf{120-1080} & 72 $\%$  ($13/18$) \\    \hline
	\textbf{1080-2160} &  94  $\%$ ($17/18$) \\  \hline
	\hline
	\end{tabular} \\
	}
	\caption{Percentage of similarity in the prediction of classes. We report the percentage of cases in which at least 3 prediction on the 4 considered in previous histograms are the same.}
	\label{tab:PercAI}
\end{table}

For adaptive methods, increasing the number of training vectors, a posture is misclassified always with the same wrong classes. It means that the increment of training samples helps in term of average recognition, but the classifier mistakes in the same way different postures.\\
The same analysis can be done considering different algorithms with the same number of training vectors. Results of percentage are reported in Table \ref{tab:PercAI}.

\begin{table}[H]
	
	\resizebox{13cm}{!}	{
	
	\begin{tabular} {|| p{7cm} | p{5cm} ||}
	
	\hline
	\hline
	\textbf{120 training vectors}  & \textbf{Percentage} \\ 		\hline
	\hline
	\textbf{MultiAdapt - MKAL} & 50 $\%$  ($9/18$) \\    \hline
	\textbf{MKAL - H-L2L} &  72  $\%$ ($13/18$) \\  \hline
	\textbf{Multi Adapt - H-L2L} &  33  $\%$ ($6/18$) \\  \hline
	\hline
	\hline
	\textbf{1080 training vectors}  & \textbf{Percentage} \\ \hline
	\hline
	\textbf{MultiAdapt - MKAL} & 89 $\%$  ($16/18$) \\    \hline
	\textbf{MKAL - H-L2L} & 83 $\%$ ($15/18$) \\  \hline
	\textbf{Multi Adapt - H-L2L} & 67 $\%$ ($12/18$) \\  \hline
	\hline
	\hline
	\textbf{2160 training vectors}  & \textbf{Percentage} \\ 		\hline
	\hline
	\textbf{MultiAdapt - MKAL} &  72 $\%$  ($13/18$) \\    \hline
	\textbf{MKAL - H-L2L} &  89  $\%$ ($16/18$) \\  \hline
	\textbf{Multi Adapt - H-L2L} &  61  $\%$ ($11/18$) \\  \hline
	\hline
	\end{tabular} \\
	}
	\caption{Percentage of similarity in the prediction of classes. We report the percentage of cases in which at least 3 prediction on the 4 considered in previous histograms are the same.}
	\label{tab:PercAI}
\end{table}

From values of previous Table we conclude that, given a class, different methods, generally, misclassify it with the same wrong classes.\\ 
As done in previously we underline that our classification problem involves very similar classes (8 finger movements and 9 wrist postures, see Figure \ref{fig:AllMovements}), thus it is reasonable to expect an high degree of confusion between similar postures.\\

\clearpage{\pagestyle{empty}\cleardoublepage}

\clearpage{\pagestyle{empty}\cleardoublepage}
\chapter*{Ringraziamenti}
\pagestyle{empty}

La domanda più frequente che mi hanno fatto in questi otto mesi di tesi è stata :« Perché, te che hai studiato Fisica, sei venuta a fare la tesi a Ingegneria Informatica? »\\...in realtà \textit{“ Tutto accadde su un vagone che portava a Frosinone... "} (\textit{cit.} P.F.)\\

...E quindi, per primo, vorrei ringraziare il Professor Giovanni B. Bachelet che mi ha spinto a intraprendere una nuova strada e ad iniziare una nuova avventura. Vorrei poi ringraziare la Professoressa Barbara Caputo, perché non avrei potuto trovare una relatrice che mi seguisse in modo più attento e stimolante. Grazie per avermi dato subito fiducia, per avermi insegnato a non aver paura di tutti quegli argomenti inizialmente incomprensibili e, infine,  per avermi fatto appassionare al mio lavoro.\\

Nonostante questo lavoro sia dedicato a loro, non troverò mai sufficienti parole per ringraziare i miei genitori e mia sorella. Grazie per aver creduto in me durante tutto il viaggio!\\

Un ringraziamento speciale va a Luca, senza il quale il mio percorso universitario e questi ultimi cinque anni di vita non sarebbero stati gli stessi. Grazie per la tua vicinanza in ogni momento, grazie per avermi incoraggiata e avermi anche messo di fronte ai miei errori. Questo mio traguardo è anche un po' tuo.\\

Vorrei poi ringraziare le mie coinquiline, vecchie e nuove, per avermi accompagnata in questi “anni di Roma". Grazie ad Anto e Novi per avermi sopportata in questi mesi di tesi, grazie per aver condiviso dei bei momenti e di essermi state vicine, anche quando il computer era sulle mie gambe dalla mattina alla sera! Le vecchie coinquiline sono troppe per essere nominate tutte, ma vorrei ringraziare in modo particolare Dila e Noemi per non essere solo delle ex-coinquiline, ma delle amiche.\\
  
Inutile dire che in questi otto mesi ho incontrato tantissime persone che non avrei mai conosciuto se non avessi scelto di fare questa tesi. Ringrazio quindi tutti i membri del gruppo VANDAL per avermi aiutata a capire e risolvere anche i dubbi che per voi erano così banali. A particular thanks goes to Ilja, who, despite my english and my poor initial preparation, helped me without discouraging.\\

L'ultimo ringraziamento, ma non per questo meno importante, va a tutti i miei amici: quelli di Fisica, per aver condiviso questo percorso con me, e tutti gli altri, per avermi sempre fatto il tifo da fuori.\\

\end{document}